\documentclass[conference, twocolumn]{IEEEtran}
\usepackage{booktabs}

\usepackage{graphicx}
\usepackage{array}

\usepackage[utf8]{inputenc}
\usepackage[T1]{fontenc}

\usepackage{listings}
\usepackage{subcaption}
\usepackage[table,xcdraw]{xcolor}
\usepackage{amsmath}
\usepackage{bbm}
\usepackage{pifont}
\usepackage{amssymb}
\definecolor{mydarkblue}{rgb}{0,0.08,0.65}
\usepackage[colorlinks=true,
    linkcolor=mydarkblue,
    citecolor=mydarkblue,
    filecolor=mydarkblue,
    urlcolor=mydarkblue]{hyperref}
\usepackage{cleveref}
\usepackage{soul}
\usepackage[shortlabels]{enumitem}
\usepackage{url}
\usepackage{color}
\usepackage{tikz}
\usepackage{balance}
\usepackage{multirow}
\usepackage{comment}
\usepackage{wrapfig,booktabs}

\usepackage[symbol]{footmisc}
\usepackage{tablefootnote}
\usepackage{tabularx}
\usepackage{placeins}
\usepackage{algorithm}
\usepackage{algpseudocode}
\usepackage{ragged2e}
\usepackage{cuted}
\usepackage{float}
\usepackage{caption}

\usepackage{booktabs}
\usepackage{xcolor}

\newcommand{\lossdiff}[1]{\textbf{\textcolor{black}{#1}}}

\usepackage{multicol}

\usepackage{dblfloatfix}
\usepackage{natbib}

\renewcommand{\arraystretch}{1.2}

\definecolor{codegreen}{rgb}{0,0.6,0}
\definecolor{codegray}{rgb}{0.5,0.5,0.5}
\definecolor{codepurple}{rgb}{0.58,0,0.82}
\definecolor{backcolour}{rgb}{0.95,0.95,0.92}

\usepackage{xcolor}
\usepackage{pgfplots}
\pgfplotsset{compat=1.18}
\usepgfplotslibrary{groupplots}

\pgfplotscreateplotcyclelist{nagcolors}{
{blue!100!white, thick},
{blue!80!white, thick},
{blue!60!white, thick},
{blue!40!white, thick},
{blue!20!white, thick},
}

\pgfplotscreateplotcyclelist{baselinecolors}{
{red!100!white, thick},
{red!80!white, thick},
{red!60!white, thick},
{red!40!white, thick},
{red!20!white, thick},
}

\definecolor{nagblue}{HTML}{0072B2}
\definecolor{nagorange}{HTML}{E69F00}
\definecolor{naggreen}{HTML}{009E73}
\definecolor{nagred}{HTML}{D55E00}
\definecolor{nagpurple}{HTML}{CC79A7}

\pgfplotscreateplotcyclelist{multicolor}{%
    {color=nagblue, thick, solid, mark=none}\\%
    {color=nagorange, thick, solid, mark=none}\\%
    {color=naggreen, thick, solid, mark=none}\\%
    {color=nagred, thick, solid, mark=none}\\%
    {color=nagpurple, thick, solid, mark=none}\\%
}

\makeatletter
\def\blfootnote{\xdef\@thefnmark{}\@footnotetext}
\makeatother

\lstdefinestyle{mystyle}{
  backgroundcolor=\color{backcolour},   commentstyle=\color{codegreen},
  keywordstyle=\color{magenta},
  numberstyle=\tiny\color{codegray},
  stringstyle=\color{codepurple},
  basicstyle=\ttfamily\footnotesize,
  breakatwhitespace=false,
  breaklines=true,
  captionpos=b,
  keepspaces=true,
  numbers=left,
  numbersep=5pt,
  showspaces=false,
  showstringspaces=false,
  showtabs=false,
  tabsize=2,
}

\AtBeginDocument{%
  \providecommand\BibTeX{{%
    \normalfont B\kern-0.5em{\scshape i\kern-0.25em b}\kern-0.8em\TeX}}}

\IEEEoverridecommandlockouts
\begin{document}

\title{Scaling Adaptive Depth with Norm-Agnostic Residual Networks}
\newcommand{\corr}{\textsuperscript{*}}

\author{Tomas Figliolia\corr, Beren Millidge
\\[0.5em]
\textbf{Zyphra}
\\
San Francisco, CA \\
\IEEEauthorblockA{\textsuperscript{*}Corresponding authors: \texttt{tom@zyphra.com}}
}
\maketitle
\setcounter{page}{1}

\begin{abstract}\normalfont\mdseries
Residual architectures are ubiquitous in deep learning, but they suffer from a subtle structural limitation: the norm of the residual stream can grow rapidly with depth. As a result, updates from later layers become small relative to the accumulated residual state. This reduces their impact on the representation and limits the benefits of scaling models in depth. To address this, we introduce NAG, a norm-agnostic residual architecture that separates magnitude from directional information in the residual stream, preserving meaningful layer contributions throughout depth and preventing later updates from being systematically suppressed by residual-norm growth. Importantly, NAG introduces only a negligible number of additional parameters and relies on simple operations that are easily kernel-fusible, preserving training efficiency in practice.
We show that this architecture outperforms baseline Transformers, with gains that increase substantially as depth grows, enabling effective training of much deeper models. The norm-agnostic formulation also leads to an interpretable Mixture-of-Depths (MoD) mechanism that adaptively skips both attention and MLP layers. Beyond serving as a post-training accuracy-compute tradeoff, this mechanism can be used as a pretraining-time scaling strategy: under iso-FLOP training, compute saved by reducing per-token forward-pass cost can be reinvested into training on more tokens while keeping the total parameter count and KV-cache budget fixed. In our experiments, moderate Mixture-of-Depths rates of approximately \(20\%\)--\(25\%\) match full-depth baseline performance under equal training compute while substantially reducing the number of executed layer parameters and forward-pass FLOPs. These results identify sparsity in depth as a new scaling axis for fixed-compute training, enabling very deep yet FLOP-efficient models.
\end{abstract}

\section{Introduction}
\label{sec:introduction}

The residual stream is one of the core innovations enabling the massive scale of deep-learning models \citep{he2016deep,vaswani2017attention,brown2020language}, in part because residual architectures largely solve the vanishing and exploding gradients problem of early models \citep{he2016deep}. By making each layer add an incremental update to a persistent hidden state, rather than directly feeding its output as the input of another layer, the residual stream creates a direct gradient path from input to output through which gradients can flow without vanishing or exploding \citep{elhage2021mathematical}. This enables stable training of deep models which ultimately underpins the massively increasing scale of modern AI systems \citep{brown2020language,hoffmann2022training,achiam2023gpt,liu2024deepseek}.

However, while deep residual networks can be trained stably, recent lines of work have uncovered underlying problems with depth utilization. The outputs from deep layers often appear to have less impact on residual stream representations than early layers, tending to converge to near-identity maps \citep{de2020batch,sun2025curse,csordas2025depth} and resulting in increasingly correlated residual streams with depth \citep{gromov2025unreasonable,jiang2025tracing,men2025shortgpt}. This may reflect a useful refinement process, in which the model moves from token embeddings toward final outputs by iteratively refining intermediate predictions \citep{geva2022transformerfeedforward,belrose2023eliciting,jastrzkebski2017residual,greff2016highway}, as seen in many classical numerical optimization algorithms. However, we argue that this process may also be due to fundamental architectural limitations, resulting in a deadweight loss of model capacity in deeper layers. One key line of evidence supporting this is that these deep layers can often be skipped entirely with minimal performance degradation \citep{veit2016residual,sajjad2022effect,fan2020reducing,csordas2025depth}, and the positions to skip can be predicted accurately using the cosine similarity to the residual stream \citep{gromov2025unreasonable}. The decreasing impact of later layers is also used heavily in the model pruning literature \citep{kim2024shortened,yang2024laco,men2025shortgpt} to often substantially shrink the depth, and hence inference cost of models with only a minor impact on overall quality. 

These facts point to a fundamental inefficiency of current architectures. Deep layers incur roughly the same FLOP and memory costs as early layers during training and inference; therefore, their poor utilization in trained models represents a substantial amount of wasted capacity. While pruning methods recover some of that cost, they are often fragile and do not address the root cause of the poor utilization of deeper layers. Moreover, the inefficient use of deeper layers implicitly weakens scaling in depth relative to scaling in width, leading to larger models tending to adopt wider aspect ratios. This is unfortunate since depth and width offer different expressive potentials, with greater model depth enabling the computation of a substantially larger class of serial algorithms within a single forward pass, potentially reducing reliance on long chain-of-thought traces and improving token efficiency.

Prior work has addressed related depth-utilization failures through normalization placement, residual scaling, and initialization. Pre-norm improves stability relative to post-norm, but can exhibit depth-dependent hidden-state norm growth, representation collapse, and weakened deep-layer contributions \citep{xiong2020layernorm,xie2023residual,sun2026cursedepthlargelanguage}. Methods such as depth-scaled initialization, DeepNorm, ReZero, LayerScale, and LayerNorm Scaling improve trainability by adjusting residual-branch scale or initialization, but they do not directly control a layer's relative effect on the current residual stream \citep{zhang2019improvingdeeptransformerdepthscaled,wang2022deepnet,bachlechner2021rezero,touvron2021going,sun2026cursedepthlargelanguage}. Post-norm variants revisit residual-stream normalization, but typically require modified residual pathways to avoid deep gradient pathologies \citep{srivastava2015training,chen2026postlayernormbackstableexpressive}.

In this paper, we argue that poor depth utilization is a structural consequence of residual-stream norm growth with depth, which itself follows from the additive nature of the residual stream. Since each layer adds an update to the existing representation, the residual norm will tend to grow whenever layer updates are positively correlated with, or even orthogonal to the current residual stream. As this norm grows, later layers must produce increasingly large updates in order to meaningfully rotate or modify the representation, biasing the model toward weak late-layer contributions. The precise mechanism depends on the normalization structure. In pre-norm architectures, weight decay on the layer weights discourages later layers from producing the large output norms required to substantially influence an already high-norm residual stream. In post-norm residual branches, the situation is subtler: normalization does not impose a truly fixed contribution norm once learnable scale and shift parameters are applied. In principle, these parameters can grow and allow later layers to increase their influence. However, this is a limited and inefficient solution. Because the scaling induced by the normalization parameters is comparatively less expressive than an unconstrained residual update, increasing layer influence tends to require injecting a large amount of energy into the residual stream. This accelerates residual norm growth and exacerbates gradient attenuation, producing a different route to poor depth utilization rather than eliminating the problem.

More subtly, in pre-norm architectures, each sublayer receives a normalized version of the residual stream, making the residual branch largely insensitive to the residual-stream norm itself \citep{ba2016layer,brody2023expressivityrole}. Two residual states with similar normalized direction but different magnitudes can therefore induce similar residual-branch outputs. After the update is added back to the residual stream, however, its relative effect depends strongly on the residual norm. The same update has little effect on a large-norm residual stream, but can substantially rotate a smaller-norm one. In this sense, token-wise input normalization removes crucial information about per-token residual-stream norms, which is required to properly calibrate a layer’s output. Pre-norm thus can act as an implicit inhibitor: a layer may produce a nonzero update that is nevertheless functionally weak.

We address this issue with our Norm-AGnostic (NAG) residual stream architecture in which each layer's contribution is independent of the current residual norm. The key idea is to separate scale from direction: each layer output is rescaled so that it primarily induces a controlled rotation of the residual direction, rather than an unconstrained change in magnitude. The norm is then maintained in a separate stream that controls the effective `temperature' of decoding. This makes layer contributions consistent across depth, mitigates the norm-dependent inhibition described above, and reduces the architectural bias toward attenuated gradients in earlier layers. Empirically, this improves training loss across our ablations, with the largest gains observed in the deepest models.

We also empirically study the properties of our norm-agnostic model. We demonstrate that, as expected, the contribution of each layer to the residual stream is equalized across depth. More intriguingly, we show that our norm-agnostic approach ameliorates other pathologies found in existing LLMs such as attention sink \citep{xiao2024streamingllm,gu2025attention_sink}, and heavy-tailed weight distributions leading to outliers in the residual stream \citep{bondarenko2023quantizable,sun2024massive} which are detrimental to low-precision quantization.

Moreover, we show that our formulation naturally extends to an interpretable Mixture-of-Depths (MoD) mechanism, which uses the residual-stream geometry to enable adaptive computation \citep{graves2016adaptive,dehghani2019universal,elbayad2020depthadaptive,raposo2024mixturedepths}. Since each layer's output norm is precisely controlled, a simple geometric argument lets us estimate the ratio between a layer's expected residual contribution and its maximum possible contribution \emph{before actually computing the layer}. This yields a skipping rule based solely on residual-stream geometry, without requiring a separate learned router as in standard router-based MoD. We demonstrate that this mechanism substantially outperforms classical MoD, with the advantage increasing dramatically at higher skipping rates.

Rather than treating MoD only as a post-training compute--accuracy tradeoff, we show that NAG makes MoD feasible as a pretraining-time scaling strategy. To our knowledge, this provides the first justification for using MoD during pretraining. Our iso-FLOP ablations show that forward-pass compute saved by skipping layers can be reinvested into training on more tokens under the same total compute budget. This produces similarly capable models while reducing the per-token forward-pass cost and the number of executed layer parameters, which can also accelerate inference. Our MoD mechanism therefore exposes depth sparsity as a new scaling axis for fixed-compute training, analogous to how Mixture-of-Experts enabled sparse scaling in width. Our results are thus a key step toward adaptive, compute-efficient, extremely deep models.

\section{Background}
\label{sec:background}

The core innovation of a residual network is that rather than layers composing their inputs and outputs directly as \(x_1 = f(x_0)\), we instead write the action of all layers as adding onto a persistent hidden state: \(x_1 = x_0 + f(x_0)\). Following \citet{elhage2021mathematical}, we interpret residual networks as containing a `residual stream', which is this persistent hidden state, to which every layer contributes additively. 

The residual mechanism is fundamental because it preserves an identity path directly from the input to the output, since at initialization, when layer contributions are effectively noise, the direct path from input to output dominates. This prevents vanishing or exploding gradients at initialization, enabling stable training of very deep models compared to what is possible in non-residual networks. Another way to see this intuitively is through deep linear models. Consider a three-layer linear model without a residual stream,
\begin{align}
x_3 = W_3 W_2 W_1 x_0
\end{align}
The output and gradients depend on the product of three independent weight matrices. If the leading singular values of these matrices are consistently greater than one, activations and gradients can grow rapidly with depth; if they are consistently less than one, activations and gradients can shrink rapidly toward zero. Stable training, therefore, requires careful control of these scales. Conversely, consider the equivalent residual network, which can be expressed as a polynomial of combinations of layer products,
\begin{align}
x_3 &= x_0 + W_1 x_0 + W_2 x_0 + W_3 x_0 +\notag \\
&\quad W_2 W_1 x_0 + W_3 W_1 x_0 + W_3 W_2 x_0 + W_3 W_2 W_1 x_0
\end{align}
In this case, activation and gradient shrinkage is far less severe, since the identity and first-order paths exist and can carry information and gradients. Shrinkage mainly affects the higher-order terms in this sum, reducing complex inter-layer interactions but leaving the network as a whole trainable.

However, residual networks still have a subtle flaw. Because the residual stream is updated by successively adding each layer's contribution, its norm tends to grow with depth whenever these updates have a non-negligible component aligned with the current residual stream. Even orthogonal updates increase the squared norm, and in high-dimensional spaces it is unlikely that successive updates consistently have the negative projection required to offset this growth. Thus, unless layer updates are systematically anti-correlated with the current residual stream, addition alone creates a geometric bias toward increasing residual norm. This effect can be especially pronounced when layer outputs are positively correlated with the residual stream, as is natural since each layer takes the residual stream itself as input. In early training, when the outputs of these layers point in approximately the same direction as the original embedding, the residual-stream norm can grow approximately linearly with the number of layers. Later, we will compare norm profiles with and without our proposed norm-agnostic architecture, showing that in our ablations, standard residual-stream models can reach norms up to \(120\times\) larger than their norm-agnostic counterparts.

This increase in norm makes it harder for later layers to produce large updates to the residual stream representation, in effect reducing their impact on the model and, ultimately, the efficiency of the FLOPs and parameters used to compute their outputs. To understand this, we consider three scenarios.

First, consider the post-normalization scheme used in earlier Transformer architectures \citep{vaswani2017attention,xiong2020layer,devlin2019bert}. The channel-wise parameters applied after normalization are typically exempt from weight decay and other forms of regularization. As a result, they can allow later layers to amplify their influence, but without providing precise control over how much information is injected into the residual stream. Consequently, each token-level update to the residual stream can produce a substantial increase in the residual stream norm. This rapid increase in norm also creates difficulties for gradient propagation: it imposes a strong implicit prior that causes gradients to decay quickly in earlier layers, thereby reducing the trainability of deeper models.

Secondly, let us consider the more common pre-normalization architecture, where normalization is applied to the input of the layer and the output is free to vary. Here, unlike in post-norm, each layer can theoretically increase the norm of its output to keep up with the growing residual stream norm. However, it faces two practical challenges. First, since the input is normalized, increasing the output norm requires increasing the weight norm. This is directly penalized by weight decay, which is ubiquitous in LLM training. Second, and more subtly, while the residual stream norm can differ across tokens, this information is discarded by token-wise input normalization, which reduces all inputs to the layer to normalized embeddings. The layer therefore has no way of knowing or adapting to the input residual norm of each token, or to the likely output residual norm it must match. The layer can only match the norm on average across tokens, so a layer will undershoot or overshoot the norm when the residual stream norm for a particular token is unusually small or large.

Finally, in the case of sandwich norm \citep{ding2021cogview,team2024gemma}, which uses both pre-norm and post-norm, we observe similar pathologies as post-norm. While these issues do not prevent deep models from training stably empirically, since even deep layers receive gradients albeit under a strong decay prior, we argue that this reduces the expressivity and ultimate efficiency of the model. Each layer incurs comparable compute and memory cost, yet the structure of the residual stream means that layers do not have an equal opportunity to influence the representation. Our intuition is that an optimal architecture would place every parameter on an equal footing, so that each unit of compute has a comparable ability to modify the residual stream.

\section{Architecture}
\label{sec:architecture}

\subsection{The Norm-Agnostic Residual Stream Network}
\label{subsec:the_norm_agnostic_residual_stream_network}

To address residual-stream norm growth and the resulting depth-dependent imbalance in layer influence, we introduce the \emph{Norm-AGnostic} residual network (NAG). NAG decouples the magnitude of the residual stream from its representational content by keeping the high-dimensional residual direction on a fixed-radius hypersphere.

Let \(R_l \in \mathbb{R}^d\) denote the residual stream at layer \(l\), where \(d\) is the hidden size. We write the normalized residual stream as
\begin{equation}
\bar R_l = N_{\mathrm{in}}(R_l)
\end{equation}
In the setting considered here, \(N_{\mathrm{in}}\) is a scale-only normalization that preserves direction, so that the residual stream can be decomposed as
\begin{equation}
R_l = \rho_l \bar R_l
\end{equation}
where \(\rho_l\) represents the scalar norm and \(\bar R_l\) represents the angular direction of the normalized residual stream vector. For example, if \(N_{\mathrm{in}}\) normalizes vectors to have \(\ell_2\)-norm \(\sqrt d\), then
\begin{equation}
\rho_l = \frac{\|R_l\|_2}{\sqrt d}
\end{equation}
The standard residual update is
\begin{equation}
R_{l+1} = R_l + f_l\!\left(N_{\mathrm{in}}(R_l)\right) = R_l + f_l(\bar R_l)
\end{equation}
A natural first attempt at making the layer norm-agnostic is to scale the layer output by the current residual scale
\begin{equation}
R_{l+1} = R_l + \rho_l f_l(\bar R_l) = \rho_l\left(\bar R_l + f_l(\bar R_l)\right)
\end{equation}
Under this update, the direction of \(R_{l+1}\) depends only on \(\bar R_l\), not on the magnitude \(\rho_l\). Thus, the layer computation is theoretically invariant to the norm of the residual stream, while the residual scale is transported multiplicatively through depth.

Despite its intuitive appeal, this update proved highly unstable in practice. Layer outputs are often strongly correlated with their inputs, especially early in training. Consequently, \(f_l(\bar R_l)\) can contain a large component in the direction of \(\bar R_l\), causing the update
\begin{equation}
\bar R_{l+1} = \bar R_l + f_l(\bar R_l)
\end{equation}
to behave locally like a multiplicative gain on the residual stream
\begin{equation}
\bar R_{l+1} \approx g_l \bar R_l
\end{equation}
with \(g_l\) usually larger than one at the beginning of training. Since each subsequent layer can impose its own gain, these effects compound multiplicatively across depth, leading to exponential growth of the residual-stream norm, creating massive differences in the scale of gradient norms across the model. In practice, this can make gradients in later layers many orders of magnitude larger than those in earlier layers, producing an effective vanishing-gradient problem for the early layers and recreating the low depth utilization problem.

To address the problem of norm growth by correlated writes to the residual stream, we orthogonalize the output of each layer with respect to its input. In addition, we center each residual-branch output by subtracting its feature-wise mean, which we found improved numerical stability and downstream performance. This centering is implemented as the final operation inside \(f_l\), so throughout the following notation \(f_l(\bar R_l)\) denotes the centered layer output. The embedding-table rows are also centered at zero. If \(R_l\) is centered and the update added at layer \(l\) is centered, then \(R_{l+1}\) remains centered as well. Let
\begin{equation}
f_l^\perp(\bar R_l) = f_l(\bar R_l) - \frac{\left\langle f_l(\bar R_l), \bar R_l \right\rangle}{\|\bar R_l\|_2^2} \bar R_l
\label{eq:orthogonalization}
\end{equation}
denote the component of the layer output orthogonal to the normalized residual stream. The residual update then becomes
\begin{equation}
R_{l+1} = \rho_l\left(\bar R_l + f_l^\perp(\bar R_l)\right)
\end{equation}
This orthogonalization is motivated by ideas similar to those in \citet{zhai2026exclusiveselfattention}, where values in the KV cache are orthogonalized with respect to the current value. Intuitively, if layers are no longer making decisions based on the residual norm, then adding a component parallel to the residual stream does not rotate the residual direction; it primarily increases the transported norm. This is undesirable because parallel writes cause the residual norm to grow faster than orthogonal writes, strengthening the norm-induced decay prior on gradients to earlier layers. Since the norm is already tracked separately, there is little reason to inject a forward-pass update aligned with the residual stream itself. We therefore remove this component through orthogonalization, forcing each layer contribution to act primarily as a controlled rotation of the residual representation.

However, because the residual-branch output is often highly correlated with the residual direction early in training, the orthogonal component \(f_l^\perp(\bar R_l)\) can initially be very small. This reduces the effective size of the layer update and can hurt sample efficiency. To avoid this, we define the normalization function such that its output has norm \(\sqrt d\):
\begin{equation}
N_{\mathrm{out}}(x) = \sqrt d \frac{x}{\|x\|_2}
\end{equation}
To provide finer control over the norm of the residual-stream update, we introduce two additional mechanisms: a trainable scalar \(\alpha_l\) and an input-dependent norm modulator \(m_l\). Prior work on latent MoEs \citep{elango2026latentmoe} and CCA \citep{figliolia2025compressed} suggests that model computations can often be carried out in spaces that are substantially more rank-constrained than the full residual-stream dimensionality. Motivated by this observation, we hypothesize that each layer is associated with a set of preferred directions in representation space. When the residual stream aligns with these directions, the layer should be allowed to inject a larger update; otherwise, its contribution should be suppressed.
Accordingly, we constrain the norm modulator to output values in \([0,1]\), while the trainable scalar \(\alpha_l\) determines the maximum contribution that \(m_l\) can provide. We parametrize the norm modulator as
\begin{equation}
m_l = \left(\sum_{i=0}^{C-1} p_{li} \sigma({\bar R_l}^\top w_{li} + b_{li}) \right)^{\beta_{l}}
\end{equation}
Here, \(C\) is a hyperparameter specifying the number of preferred directions, \(\sigma\) denotes the sigmoid function, \(\beta_l > 0\) controls the sharpness of the modulation, and \(w_{li}\) are learned preferred directions with their biases \(b_{li}\). The coefficients \(p_{li}\) are obtained by applying a softmax over learnable parameters, ensuring that \(p_{li} \geq 0\) and \(\sum_i p_{li} = 1\). Thus, before exponentiation, the modulator is a convex combination of sigmoid gates and therefore lies in \([0,1]\). Raising this value to the positive power \(\beta_l\) preserves the same range while changing the curvature of the response: values \(\beta_l > 1\) make intermediate modulator values more suppressive, whereas values \(0 < \beta_l < 1\) make them closer to one. The role of \(\beta_l\) becomes especially important when we introduce our MoD method in a later section.

This parametrization has a natural geometric interpretation. Each of the \(C\) components defines a learned soft preference over the normalized residual space through the projection \(\bar R_l^{\top} w_{li}\). When the residual stream aligns with directions favored by the layer, the corresponding gates activate and the modulator outputs a larger value, allowing the layer update to contribute more strongly to the residual stream. Conversely, when the residual stream lies outside these preferred regions, the modulator outputs a smaller value and suppresses the update. In this sense, the modulator acts as a lightweight input-dependent gate, allowing each layer to specialize its updates to particular regions or directions of representation space.

This construction is also analogous to the router in an MoE layer: just as a linear classifier learns directions that score and select among experts, the norm modulator learns directions that determine when a layer's residual update should be emphasized or suppressed. Since we apply this mechanism to MoE layers in our models, we choose \(C\) in a layer to be at least the number of experts in that layer, ensuring that the modulator has sufficient directional capacity to represent expert-specific preferences. In our ablations, we use \(C=32\); empirically, we find that \(C\) typically needs to be at least \(8\)--\(16\) to provide sufficient flexibility.

Putting these pieces together, the full norm-agnostic update is
\begin{equation}
R_{l+1} = \rho_l \bar R_l + \rho_l \alpha_l m_l(\bar R_l) N_{\mathrm{out}}\left(f_l^\perp(\bar R_l) \right)
\label{eq:full_update}
\end{equation}
Our update has a simple geometric interpretation. The layer rotates the residual stream in a direction orthogonal to its current state by an amount controlled by the norm modulator. Since both \(\bar R_l\) and \(N_{\mathrm{out}}(f_l^\perp(\bar R_l))\) have norm \(\sqrt d\), and since they are orthogonal, the residual norm grows by
\begin{equation}
\frac{\|R_{l+1}\|_2}{\|R_l\|_2} = \sqrt{1 + \alpha_l^2 m_l^2(\bar R_l)}
\label{eq:gain_update}
\end{equation}
Thus, for fixed \(\alpha_l\), the norm-modulator controls both the maximum rotation applied by the layer and the maximum per-layer increase in the residual norm.

For improved numerical properties, we store the norm scalar \(\rho_l\) in log space. Let
\begin{equation}
s_l = \log \rho_l
\end{equation}
Since each layer operates only on the normalized residual stream \(\bar R_l\), while \(s_l\) tracks the accumulated norm through the forward pass, after applying a layer update, we renormalize the resulting direction and transfer the corresponding gain to the log-scale variable. Concretely, the full norm-agnostic residual stream update becomes,
\begin{equation}
\begin{aligned}
\bar R_{l+1} &= \frac{\bar R_l + \alpha_l m_l(\bar R_l) N_{\mathrm{out}}\left(f_l^\perp(\bar R_l)\right)}{\sqrt{1 + \alpha_l^2 m_l^2(\bar R_l)}}\\
s_{l+1} &= s_l + \frac{1}{2} \log{(1 + \alpha_l^2 m_l^2(\bar R_l))}
\end{aligned}
\end{equation}
By storing the residual stream and its norm separately, we keep the high-dimensional residual stream bounded on the normalized hypersphere while preserving the accumulated norm in a high-precision scalar channel.

\begin{figure}[t]
\centering
\includegraphics[width=1.0\linewidth]{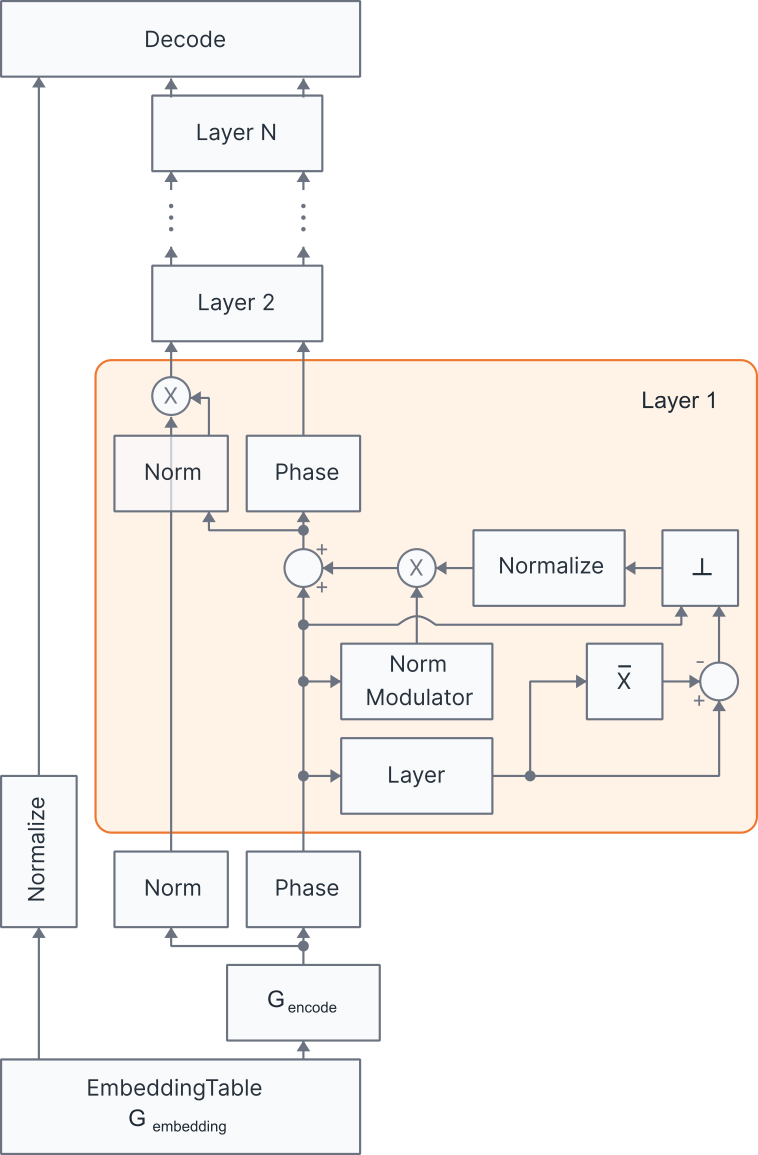}
\caption{Schematic of the norm-agnostic architecture. The residual stream is decomposed into separately propagated norm and direction components. Each layer operates only on the direction component: it computes an update, removes its mean, orthogonalizes it with respect to the incoming residual-stream direction, re-normalizes it, applies norm modulation, and adds the resulting direction update back into the direction-only residual stream. The induced norm increase is then transported to the norm lane.}
\label{fig:schematic}
\end{figure}

Figure~\ref{fig:schematic} illustrates the norm-agnostic architecture. An embedding is first retrieved, scaled by the gain factor \(G_{\mathrm{encode}}\), and decomposed into a norm lane and a phase lane. All subsequent layers operate only on the phase component of the residual stream. After each layer computes an update, the result is normalized before being added back into the phase lane; the corresponding local gain is accumulated in the norm lane, where it is stored in high precision. Thus, unlike conventional normalization, the procedure does not discard norm information but rather transports it explicitly through the network.

At decoding time, the final residual-stream direction \(\bar R_L\) is compared with the normalized rows of the embedding table, producing cosine similarities. These logits are scaled by the accumulated confidence \(\rho_L=\exp(s_L)\), which contains the product of the initial embedding norm \(G_{\mathrm{embedding}}\), the learnable post-embedding gain \(G_{\mathrm{encode}}\), and the transport gain accumulated through the network, \(G_{\mathrm{travel}}\). Because the decoder uses normalized embedding vectors, this carried norm acts as an inverse temperature. We therefore define the effective temperature as its reciprocal:
\begin{equation}
T_{\mathrm{eff}} = \frac{1}{G_{\mathrm{travel}} \cdot G_{\mathrm{embedding}} \cdot G_{\mathrm{encode}}}
\end{equation}
Thus, output probabilities are determined by angular similarity to the embedding table, while the sharpness of the distribution is controlled by the norm accumulated during the forward pass. This gives the model a mechanism for learning its effective decoding temperature: higher temperatures can support uncertain predictions early in training, while lower temperatures allow sharper predictions as the model becomes more confident.

We apply the norm-agnostic formulation to all layer types in the model, both MoE layers and attention layers. In standard architectures, output normalization can degrade attention-layer performance, since the norm of the attention output may encode information retrieved from the KV cache. However, in our norm-agnostic architecture, we did not observe any detrimental effect from applying the same normalization procedure to attention layers.

The additional operations introduced by the norm-agnostic formulation are lightweight and add only a negligible number of parameters to the model. The only added parameters come primarily from the norm modulator, which contributes approximately \(Cd\) parameters per layer, a negligible increase relative to the full model size which scales as \(d^2\). Orthogonalization, output normalization, norm modulation, and norm tracking are all scalar per-token operations, making them well suited to efficient implementation using kernel fusion.

\subsection{Depth-Scaled Initialization}
\label{subsec:depth_scaled_initialization}

A natural question is how to initialize the trainable parameters \(\alpha_l\). If \(\alpha_l\) were initialized uniformly across layers, then every layer would initially be allowed to contribute the same maximum amount of rotation, regardless of depth. As a starting point, we analyze an idealized non-norm-agnostic setting in which the initial residual stream begins with unit norm and each layer adds an orthogonal update with unit norm. Equivalently, all norms in this discussion are measured relative to the normalization scale \(\sqrt d\). Since each update is orthogonal to the current residual stream, the squared norm grows additively:
\begin{equation}
\begin{aligned}
\|R_1\| &= \sqrt{1^2 + 1^2} \\
\|R_2\| &= \sqrt{1^2 + 1^2 + 1^2}
\end{aligned}
\end{equation}
and more generally, after \(l\) layers,
\begin{equation}
\|R_l\| = \sqrt{1 + l}
\end{equation}
The relative norm gain between two consecutive layers is therefore
\begin{equation}
\frac{\|R_{l}\|}{\|R_{l - 1}\|} = \frac{\sqrt{1 + l}}{\sqrt{1 + l - 1}} = \sqrt{1 + \frac{1}{l}}
\end{equation}
In the norm-agnostic formulation, the layer update is scaled relative to the current residual norm. Ignoring the norm modulator at initialization, the per-layer norm gain is
\begin{equation}
\frac{\|R_{l}\|}{\|R_{l - 1}\|} = \sqrt{1 + \alpha_l^2}
\end{equation}
To match the additive orthogonal-growth behavior described above, we initialize
\begin{equation}
\label{eq:init_gain}
\alpha_l = \frac{1}{\sqrt{l}}
\end{equation}
where \(l\) is the layer index, starting from \(l=1\).

The norm-agnostic formulation does not require this specific initialization of \(\alpha_l\). In principle, one could use a larger initialization, such as \(\alpha_l = 1\) for all layers, allowing every layer to potentially contribute up to the same maximum magnitude regardless of its depth. We therefore explored a more general family of initializations
\begin{equation}
\alpha_l = \frac{1}{l^p}
\end{equation}
where \(p\) controls how quickly the initial update scale decays with depth. The choice \(p = 0.5\) recovers Eq.~\eqref{eq:init_gain}. Smaller values, such as \(p = 0.25\), make the initial contributions of different layers more uniform, while larger values, such as \(p = 0.75\), make deeper layers contribute less at initialization. Although we find empirically that the best initial values of \(\alpha\) are obtained with \(p = 0.5\), this does not preclude different layers from inducing comparable average rotations of the residual stream, as we show in the results section.

\section{Results}
\label{sec:results}

\subsection{NAG Loss and Evaluations}
\label{subsec:nag_loss_and_evaluations}

After training with input normalization, each layer implicitly becomes tuned to an expected residual-stream scale at which its contribution is meaningful. However, residual embeddings with different norms can induce similar layer outputs. In one case, the output may substantially rotate the residual stream; in another, it may be diluted by a much larger residual norm. We refer to this norm-dependent weakening of a layer's relative contribution as an inhibitory effect. NAG is designed to remove this inhibitory effect by making each layer's contribution independent of the current residual-stream norm. This property should become increasingly important in deeper models, recurrent or looped architectures \citep{raposo2024mixture,csordas2025depth,csordas2024moeut,geiping2026scaling}, and mixture-of-depth settings where tokens may undergo different amounts of prior computation \citep{wang2022deepnet,dehghani2018universal,raposo2024mixture,bae2025mixture}. In such settings, the residual-stream norm can vary substantially across tokens, making norm-dependent inhibition especially severe.

To test whether mitigating this inhibitory effect improves depth utilization, we conducted an ablation study on MoE Transformer models with and without NAG. We considered width-to-depth ratios of 80, 60, 40, 30, and 20, where smaller ratios correspond to deeper models. Across all settings, both the total parameter count and the forward-pass parameter count were kept fixed. Table~\ref{tab:ablation-config} summarizes the configuration of each experiment.

\begin{table*}[t]
\centering
{\small
\begin{tabular*}{0.75\textwidth}{@{\extracolsep{\fill}}lrrrrr}
\toprule
 & \multicolumn{5}{c}{Model Ratio} \\
\cmidrule(lr){2-6}
 & 80 & 60 & 40 & 30 & 20\\
\midrule
Number of layers  & 18 & 22 & 28 & 34 & 40\\
Hidden size & 1408 & 1280 & 1120 & 1024 & 928\\
Average number of experts & 8 & 7.91 & 7.82 & 7.82 & 7.78\\
MLP hidden size & 2816 & 2560 & 2328 & 2096 & 1984\\
\bottomrule
\end{tabular*}}
\caption{Model configurations used in the ablation study across different model ratios. All configurations maintain approximately 150M forward-pass parameters and 900M total parameters. Non-integer expert counts are handled by interleaving layers with more and fewer experts to achieve the desired ratio.}
\label{tab:ablation-config}
\end{table*}

All ablation models are MoE Transformer architectures with 150M forward-pass parameters and 900M total parameters. Each Transformer layer consists of an attention block followed by an MoE feed-forward block. Models were trained with a sequence length of 2048 using our CCA attention layer with CCGQA \citep{figliolia2025compressed}, with 8 query heads and 2 key/value heads. To realize effective non-integer average expert counts, we interleaved layers using two different numbers of experts.

All models were trained for 50 billion tokens on the Zyda2 dataset \citep{tokpanov2024zyda,tokpanov2024zyda2}. We used the Muon optimizer \citep{jordan2024muon}, a batch size of approximately 3 million tokens, and a cosine learning-rate schedule initialized at \(7.5 \times 10^{-4}\) and decayed to \(5 \times 10^{-5}\).

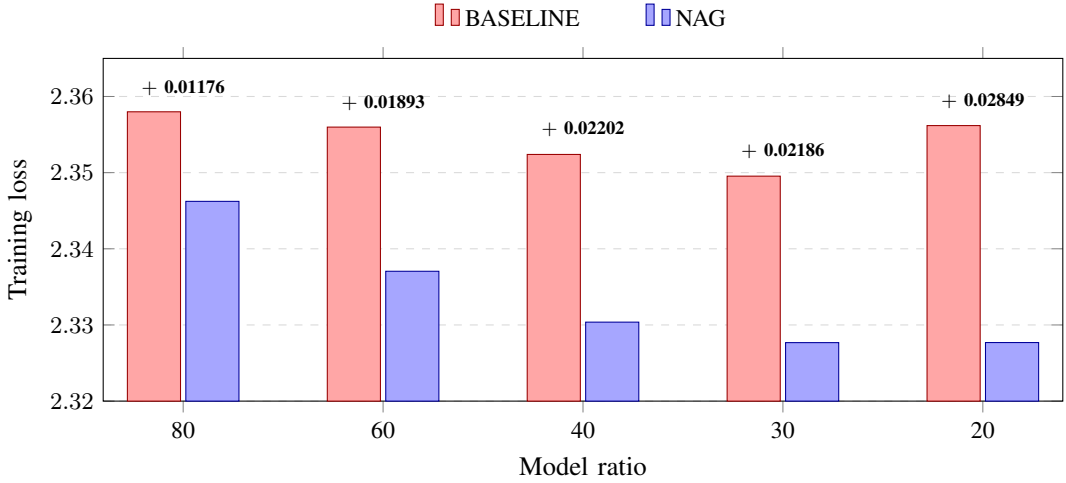
\begin{figure*}[t]
\centering
\begin{tikzpicture}[trim axis left, trim axis right]
\begin{axis}[
    ybar,
    scale only axis=true,
    width=0.70\textwidth,
    height=0.25\textwidth,
    bar width=20pt,
    enlarge x limits=0.10,
    ymin=2.32,
    ymax=2.365,
    ylabel={Training loss},
    ytick distance=0.01,
    xlabel={Model ratio},
    symbolic x coords={80,60,40,30,20},
    xtick=data,
    legend style={
        at={(0.5,1.05)},
        anchor=south,
        legend columns=2,
        draw=none,
        font=\small,
        /tikz/every even column/.append style={column sep=0.8cm}
    },
    tick label style={font=\small},
    ymajorgrids=true,
    grid style={dashed,gray!30},
    clip=false,
]
\addplot+[ybar, fill=red!35!white, draw=red!60!black] coordinates {
    (80,2.35799)
    (60,2.35598)
    (40,2.35239)
    (30,2.34954)
    (20,2.35618)
};

\addplot+[ybar, fill=blue!35!white, draw=blue!60!black] coordinates {
    (80,2.34623)
    (60,2.33705)
    (40,2.33037)
    (30,2.32768)
    (20,2.32769)
};

\legend{BASELINE, NAG}

\node[font=\scriptsize, anchor=south] at (axis cs:80,2.3590) {$+$ \lossdiff{0.01176}};
\node[font=\scriptsize, anchor=south] at (axis cs:60,2.3570) {$+$ \lossdiff{0.01893}};
\node[font=\scriptsize, anchor=south] at (axis cs:40,2.3535) {$+$ \lossdiff{0.02202}};
\node[font=\scriptsize, anchor=south] at (axis cs:30,2.3506) {$+$ \lossdiff{0.02186}};
\node[font=\scriptsize, anchor=south] at (axis cs:20,2.3572) {$+$ \lossdiff{0.02849}};

\end{axis}
\end{tikzpicture}

\caption{Training losses for the baseline and norm-agnostic MoE models across model ratios. Labels above each pair report the loss difference, computed as baseline loss minus NAG loss. NAG improves over the baseline at every model ratio, with the largest gain observed in the deepest configurations.}
\label{fig:ablation-loss}
\end{figure*}

Figure~\ref{fig:ablation-loss} reports the final training losses and loss differences after training. We also report downstream prefill evaluations, i.e. evaluations computed from model likelihoods rather than generated completions, in Table~\ref{tab:eval-results}. NAG consistently improves HellaSwag accuracy across the evaluated configurations. Since HellaSwag is known to correlate well with language-modeling loss, these results are consistent with the loss improvements observed in Figure~\ref{fig:ablation-loss}.

Across all measured model ratios, NAG achieves lower training loss than the corresponding baseline. Loss generally improves as models become deeper, although the gains begin to saturate at the deepest configurations. We believe this saturation is due, at this model scale, to the increasingly constrained residual-stream width required to keep the ablation models parameter-matched. As shown in Table~\ref{tab:ablation-config}, increasing depth while maintaining approximately equal parameter count reduces the hidden size from 1408 to 928, which may make the residual stream a stronger bottleneck in the deepest models. Despite this saturation, the improvement from NAG grows substantially with depth, indicating that controlling the residual-stream norm becomes increasingly important in deeper networks, as predicted by our analysis.

\begin{table*}[ht]
\centering
{\small
\begin{tabular*}{\textwidth}{@{\extracolsep{\fill}}lccccccc}
\toprule
& openbookqa & boolq & hellaswag & arc\_challenge & arc\_easy & winogrande & piqa \\
\midrule
BASELINE-80  & 0.356000 & 0.478593 & 0.507568 & 0.308020 & 0.589646 & 0.520126 & 0.723069 \\
NAG-80  & 0.348000 & 0.582875 & {\bf 0.516730} & 0.310580 & 0.577862 & 0.559590 & 0.719804 \\
\midrule
BASELINE-60  & 0.322000 & 0.578593 & 0.512547 & 0.308020 & 0.584596 & 0.530387 & 0.715452 \\
NAG-60 & 0.352000 & 0.582263 & {\bf 0.525991} & 0.315700 & 0.583754 & 0.518548 & 0.719804 \\
\midrule
BASELINE-40  & 0.352000 & 0.531804 & 0.513742 & 0.313993 & 0.585438 & 0.538279 & 0.717084 \\
NAG-40  & 0.366000 & 0.529358 & {\bf 0.533260} & 0.324232 & 0.607744 & 0.531176 & 0.719804 \\
\midrule
BASELINE-30  & 0.368000 & 0.552294 & 0.516033 & 0.311433 & 0.589646 & 0.533544 & 0.731230 \\
NAG-30  & 0.360000 & 0.535474 & {\bf 0.529576} & 0.335324 & 0.603535 & 0.552486 & 0.716540 \\
\midrule
BASELINE-20  & 0.358000 & 0.587462 & 0.515435 & 0.300341 & 0.573653 & 0.533544 & 0.720892 \\
NAG-20  & 0.358000 & 0.571254 & {\bf 0.534256} & 0.309727 & 0.581650 & 0.543015 & 0.728509 \\
\bottomrule
\end{tabular*}}
\caption{Evaluation accuracy across benchmarks for BASELINE and NAG models trained for 50B tokens. We emphasize HellaSwag because, in this training regime, it is the least noisy and tracks loss trends more closely than the other benchmarks. NAG outperforms the corresponding BASELINE model on HellaSwag at every evaluated depth.}
\label{tab:eval-results}
\end{table*}

\subsection{Residual Stream Norm}
\label{subsec:residual_stream_norm}

\begin{figure*}[t]
\centering
\begin{subfigure}[t]{0.33\textwidth}
\centering
\begin{tikzpicture}[trim axis left, trim axis right]
\begin{axis}[
    width=\linewidth,
    height=0.85\linewidth,
    xlabel={Layer},
    ylabel={Norm},
    xmin=0,
    xmax=80,
    ymin=0,
    ymax=400,
    grid=both,
    legend pos=north west,
    legend style={font=\scriptsize},
    tick label style={font=\scriptsize},
    label style={font=\small},
    cycle list name=baselinecolors,
    ytick distance=40,
    xtick distance=10,
]
\addplot table[col sep=comma, x=Layer, y={Baseline-80}] {Norm_BASELINE.csv};
\addlegendentry{80}

\addplot table[col sep=comma, x=Layer, y={Baseline-60}] {Norm_BASELINE.csv};
\addlegendentry{60}

\addplot table[col sep=comma, x=Layer, y={Baseline-40}] {Norm_BASELINE.csv};
\addlegendentry{40}

\addplot table[col sep=comma, x=Layer, y={Baseline-30}] {Norm_BASELINE.csv};
\addlegendentry{30}

\addplot table[col sep=comma, x=Layer, y={Baseline-20}] {Norm_BASELINE.csv};
\addlegendentry{20}
\end{axis}
\end{tikzpicture}
\caption{Baseline norm.}
\label{fig:BASELINE-norm}
\end{subfigure}
\hspace{-0.02\textwidth}
\begin{subfigure}[t]{0.33\textwidth}
\centering
\begin{tikzpicture}[trim axis left, trim axis right]
\begin{axis}[
    width=\linewidth,
    height=0.85\linewidth,
    xlabel={Layer},
    ylabel={Norm},
    xmin=0,
    xmax=80,
    ymin=0,
    ymax=16,
    grid=both,
    legend pos=north west,
    legend style={font=\scriptsize},
    tick label style={font=\scriptsize},
    label style={font=\small},
    cycle list name=nagcolors,
    ytick distance=2,
    xtick distance=10,
]
\addplot table[col sep=comma, x=Layer, y={NAG-80}] {Norm-NAG.csv};
\addlegendentry{80}

\addplot table[col sep=comma, x=Layer, y={NAG-60}] {Norm-NAG.csv};
\addlegendentry{60}

\addplot table[col sep=comma, x=Layer, y={NAG-40}] {Norm-NAG.csv};
\addlegendentry{40}

\addplot table[col sep=comma, x=Layer, y={NAG-30}] {Norm-NAG.csv};
\addlegendentry{30}

\addplot table[col sep=comma, x=Layer, y={NAG-20}] {Norm-NAG.csv};
\addlegendentry{20}
\end{axis}
\end{tikzpicture}
\caption{NAG norm.}
\label{fig:NAG-norm}
\end{subfigure}
\hspace{-0.02\textwidth}
\begin{subfigure}[t]{0.33\textwidth}
\centering
\begin{tikzpicture}[trim axis left, trim axis right]
\begin{axis}[
    width=\linewidth,
    height=0.85\linewidth,
    ymin=500,
    ymax=1100,
    xtick distance=1,
    ytick distance=100,
    xmin=34,
    xmax=82,
    ylabel={Cumulative rotation ($^\circ$)},
    xlabel={Model depth (layers)},
    xtick={36,44,56,68,80},
    legend pos=north west,
    legend style={font=\scriptsize},
    tick label style={font=\scriptsize},
    label style={font=\small},
    ymajorgrids=true,
    grid style={dashed,gray!30},
    clip=false,
]
\addplot+[thick, no markers, red!90!black] coordinates {
(36,623)
(44,660)
(56,729)
(68,801)
(80,822)
};

\addplot+[thick, no markers, blue!90!black] coordinates {
(36,551)
(44,634)
(56,748)
(68,870)
(80,1011)
};

\legend{Baseline, NAG}
\end{axis}
\end{tikzpicture}
\caption{Cumulative rotation.}
\label{fig:angular-change}
\end{subfigure}
\caption{Layer-wise norm values and cumulative embedding rotation across models of different depths. Left: norm values for baseline models with model ratios 80, 60, 40, 30, and 20. Middle: norm values for NAG models with model ratios 80, 60, 40, 30, and 20. Note the different y-axis scales for the baseline and NAG norm plots. Right: average cumulative embedding rotation across tested model depths, comparing models trained with and without NAG. Rotation values are measured in angular degrees.}
\label{fig:norms-and-angular-change}
\end{figure*}
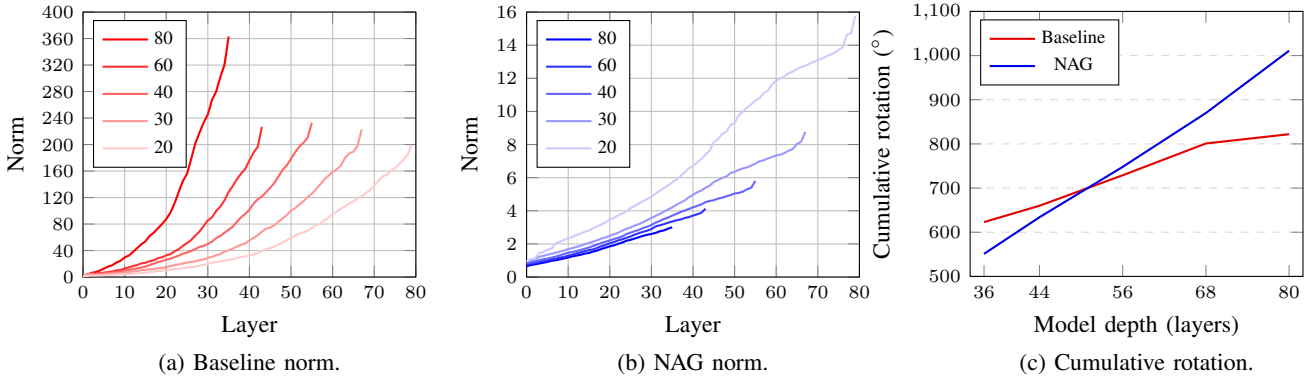

We now examine how the NAG architecture changes the behavior of the residual stream. The first two panels in Figure~\ref{fig:norms-and-angular-change} show the growth of residual-stream norm across layers in the baseline and norm-agnostic settings. NAG substantially reduces norm accumulation by inserting updates into the residual stream in a controlled, orthogonal manner, thereby limiting uncontrolled growth in magnitude. For the model with aspect ratio 80, this corresponds to an approximately \(120\times\) reduction in residual-stream norm relative to the baseline.

Interestingly, within the NAG models, the residual-stream norm increases with deeper models. Under our decoding parameterization, where the final residual norm acts as an inverse temperature, this corresponds to a lower effective temperature for deeper models. This suggests that deeper NAG models may learn to make sharper, more confident next-token predictions. Importantly, this trend is not solely a mechanical consequence of using more layers: the embedding-table norm, learnable post-embedding gain, layer scales, and norm modulators can all reduce the effective norm used at decoding if doing so is beneficial. The observed increase therefore suggests that deeper NAG models preserve, and in some cases amplify, this confidence signal as depth increases.

We aggregate the results from the model-depth ratio ablations to estimate the average cumulative angular change experienced by the token representations with and without NAG. The rightmost panel in Figure~\ref{fig:norms-and-angular-change} reports this quantity across models of varying depth. The two settings exhibit distinct scaling behavior. With NAG, cumulative rotation grows approximately linearly as depth increases, suggesting that additional layers continue to produce meaningful angular updates along the representation trajectory. In contrast, the non-NAG baseline shows weaker, sublinear growth with depth, suggesting a saturation effect in which later layers contribute progressively less to the overall representation change. This pattern is consistent with recent analyses of the curse of depth in pre-norm language models, which argue that deeper Transformer blocks can become increasingly ineffective as their derivatives approach the identity, thereby limiting their contribution during training \citep{sun2025curse}.

\subsection{Residual Stream Geometry}
\label{sec:residual_stream_geometry}

In a standard pre-norm model, each layer receives a normalized residual representation. Consequently, the layer has limited direct access to the current norm of the residual stream for each token. Its weights must therefore learn not only which direction to write into the residual stream, but also what update magnitude is useful on average.

This magnitude is implicitly tuned during training to the typical residual-stream norm encountered at that layer. When the actual residual norm for a token deviates from this typical value, the same update can induce a different effective rotation of the residual stream than intended. Equivalently, the layer output can be viewed as a directionally structured update plus a norm-dependent deviation whose effect is not directly controlled on a per-token basis. Across many layers, such deviations may accumulate and introduce additional variability into the residual stream, mixing useful directions with norm-induced perturbations and broadening the effective covariance spectrum.

Norm-agnostic updates are designed to remove this source of variation by controlling the resulting residual-stream rotation more directly for each token. Under NAG, we therefore hypothesize that the normalized residual-stream geometry becomes more concentrated: representations become less affected by norm-induced noise, allowing the model to reduce directional dispersion without necessarily collapsing onto fewer active directions.

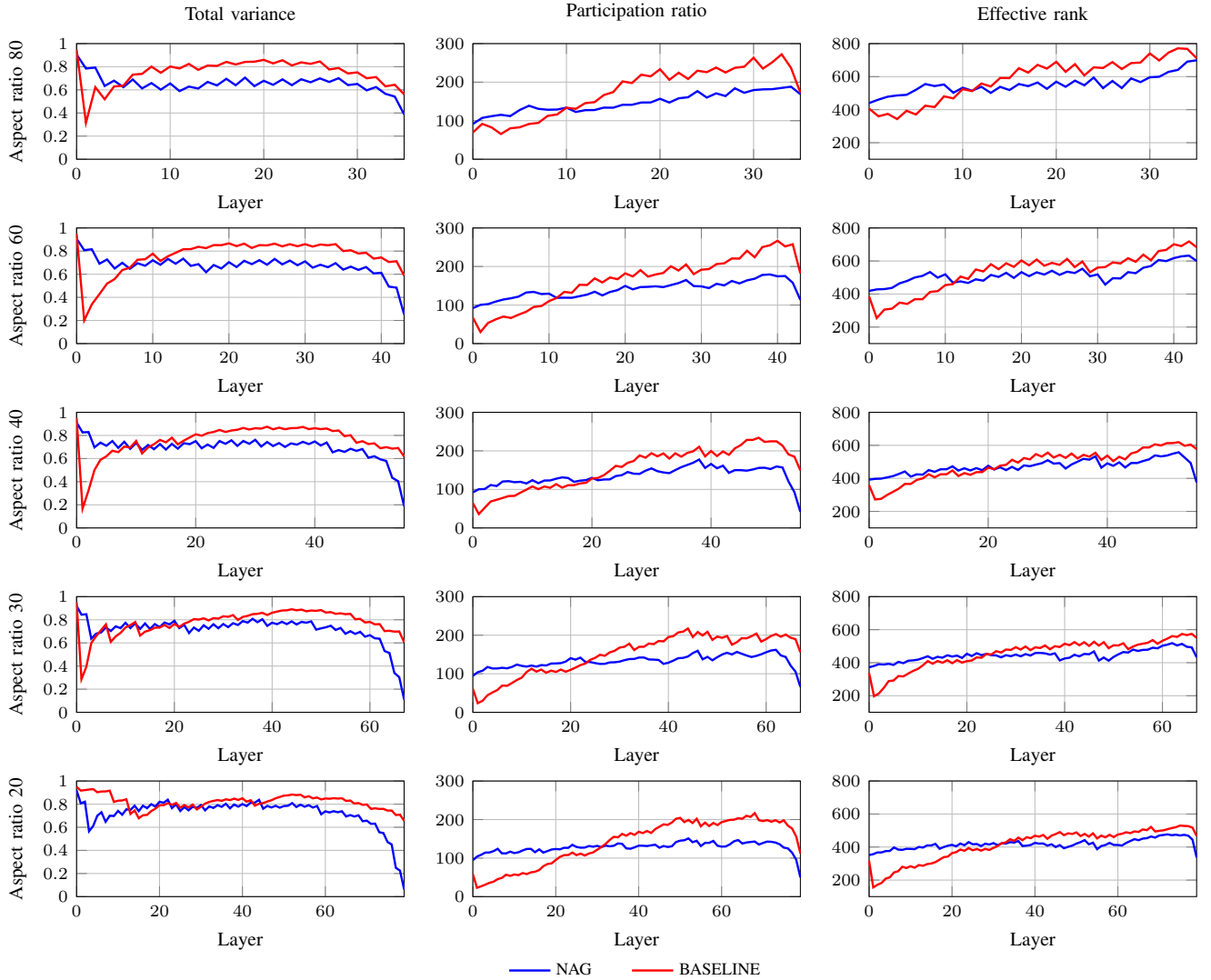
\begin{figure*}[t]
\centering
\begin{tikzpicture}
\pgfplotsset{
    nagstyle/.style={
        mark=none,
        solid,
        line width=0.9pt
    },
    baselinestyle/.style={
        mark=none,
        solid,
        line width=0.9pt
    },
    layerlimits80/.style={
        xmin=0,
        xmax=35
    },
    layerlimits60/.style={
        xmin=0,
        xmax=43
    },
    layerlimits40/.style={
        xmin=0,
        xmax=55
    },
    layerlimits30/.style={
        xmin=0,
        xmax=67
    },
    layerlimits20/.style={
        xmin=0,
        xmax=79
    },
    totalvariancelimits80/.style={
        layerlimits80,
        ymin=0,
        ymax=1
    },
    totalvariancelimits60/.style={
        layerlimits60,
        ymin=0,
        ymax=1
    },
    totalvariancelimits40/.style={
        layerlimits40,
        ymin=0,
        ymax=1
    },
    totalvariancelimits30/.style={
        layerlimits30,
        ymin=0,
        ymax=1
    },
    totalvariancelimits20/.style={
        layerlimits20,
        ymin=0,
        ymax=1
    },
    participationratiolimits80/.style={
        layerlimits80,
        ymin=0,
        ymax=300
    },
    participationratiolimits60/.style={
        layerlimits60,
        ymin=0,
        ymax=300
    },
    participationratiolimits40/.style={
        layerlimits40,
        ymin=0,
        ymax=300
    },
    participationratiolimits30/.style={
        layerlimits30,
        ymin=0,
        ymax=300
    },
    participationratiolimits20/.style={
        layerlimits20,
        ymin=0,
        ymax=300
    },
    effectiveranklimits80/.style={
        layerlimits80,
        ymin=100,
        ymax=800
    },
    effectiveranklimits60/.style={
        layerlimits60,
        ymin=100,
        ymax=800
    },
    effectiveranklimits40/.style={
        layerlimits40,
        ymin=100,
        ymax=800
    },
    effectiveranklimits30/.style={
        layerlimits30,
        ymin=100,
        ymax=800
    },
    effectiveranklimits20/.style={
        layerlimits20,
        ymin=100,
        ymax=800
    }
}

\begin{groupplot}[
    group style={
        group size=3 by 5,
        horizontal sep=1.0cm,
        vertical sep=1.0cm
    },
    width=0.35\textwidth,
    height=0.18\textwidth,
    xlabel={Layer},
    grid=both,
    tick label style={font=\scriptsize},
    label style={font=\footnotesize},
    title style={font=\footnotesize}
]
\nextgroupplot[
    title={Total variance},
    ylabel={Aspect ratio 80},
    totalvariancelimits80,
    legend to name=sharedlegend,
    legend columns=2,
    legend style={
        font=\scriptsize,
        draw=none,
        cells={anchor=west},
        /tikz/every even column/.append style={column sep=0.5cm}
    }
]
\addplot+[nagstyle] table[x=layer,y=NAG,col sep=comma] {total_variance/total-variance-80.csv};
\addlegendentry{NAG}
\addplot+[baselinestyle] table[x=layer,y=BASELINE,col sep=comma] {total_variance/total-variance-80.csv};
\addlegendentry{BASELINE}

\nextgroupplot[
    title={Participation ratio},
    participationratiolimits80
]
\addplot+[nagstyle] table[x=layer,y=NAG,col sep=comma] {participation_ratio/participation-ratio-80.csv};
\addplot+[baselinestyle] table[x=layer,y=BASELINE,col sep=comma] {participation_ratio/participation-ratio-80.csv};

\nextgroupplot[
    title={Effective rank},
    effectiveranklimits80
]
\addplot+[nagstyle] table[x=layer,y=NAG,col sep=comma] {effective_rank/effective-rank-80.csv};
\addplot+[baselinestyle] table[x=layer,y=BASELINE,col sep=comma] {effective_rank/effective-rank-80.csv};

\nextgroupplot[
    ylabel={Aspect ratio 60},
    totalvariancelimits60
]
\addplot+[nagstyle] table[x=layer,y=NAG,col sep=comma] {total_variance/total-variance-60.csv};
\addplot+[baselinestyle] table[x=layer,y=BASELINE,col sep=comma] {total_variance/total-variance-60.csv};

\nextgroupplot[
    participationratiolimits60
]
\addplot+[nagstyle] table[x=layer,y=NAG,col sep=comma] {participation_ratio/participation-ratio-60.csv};
\addplot+[baselinestyle] table[x=layer,y=BASELINE,col sep=comma] {participation_ratio/participation-ratio-60.csv};

\nextgroupplot[
    effectiveranklimits60
]
\addplot+[nagstyle] table[x=layer,y=NAG,col sep=comma] {effective_rank/effective-rank-60.csv};
\addplot+[baselinestyle] table[x=layer,y=BASELINE,col sep=comma] {effective_rank/effective-rank-60.csv};

\nextgroupplot[
    ylabel={Aspect ratio 40},
    totalvariancelimits40
]
\addplot+[nagstyle] table[x=layer,y=NAG,col sep=comma] {total_variance/total-variance-40.csv};
\addplot+[baselinestyle] table[x=layer,y=BASELINE,col sep=comma] {total_variance/total-variance-40.csv};

\nextgroupplot[
    participationratiolimits40
]
\addplot+[nagstyle] table[x=layer,y=NAG,col sep=comma] {participation_ratio/participation-ratio-40.csv};
\addplot+[baselinestyle] table[x=layer,y=BASELINE,col sep=comma] {participation_ratio/participation-ratio-40.csv};

\nextgroupplot[
    effectiveranklimits40
]
\addplot+[nagstyle] table[x=layer,y=NAG,col sep=comma] {effective_rank/effective-rank-40.csv};
\addplot+[baselinestyle] table[x=layer,y=BASELINE,col sep=comma] {effective_rank/effective-rank-40.csv};

\nextgroupplot[
    ylabel={Aspect ratio 30},
    totalvariancelimits30
]
\addplot+[nagstyle] table[x=layer,y=NAG,col sep=comma] {total_variance/total-variance-30.csv};
\addplot+[baselinestyle] table[x=layer,y=BASELINE,col sep=comma] {total_variance/total-variance-30.csv};

\nextgroupplot[
    participationratiolimits30
]
\addplot+[nagstyle] table[x=layer,y=NAG,col sep=comma] {participation_ratio/participation-ratio-30.csv};
\addplot+[baselinestyle] table[x=layer,y=BASELINE,col sep=comma] {participation_ratio/participation-ratio-30.csv};

\nextgroupplot[
    effectiveranklimits30
]
\addplot+[nagstyle] table[x=layer,y=NAG,col sep=comma] {effective_rank/effective-rank-30.csv};
\addplot+[baselinestyle] table[x=layer,y=BASELINE,col sep=comma] {effective_rank/effective-rank-30.csv};

\nextgroupplot[
    ylabel={Aspect ratio 20},
    totalvariancelimits20
]
\addplot+[nagstyle] table[x=layer,y=NAG,col sep=comma] {total_variance/total-variance-20.csv};
\addplot+[baselinestyle] table[x=layer,y=BASELINE,col sep=comma] {total_variance/total-variance-20.csv};

\nextgroupplot[
    participationratiolimits20
]
\addplot+[nagstyle] table[x=layer,y=NAG,col sep=comma] {participation_ratio/participation-ratio-20.csv};
\addplot+[baselinestyle] table[x=layer,y=BASELINE,col sep=comma] {participation_ratio/participation-ratio-20.csv};

\nextgroupplot[
    effectiveranklimits20
]
\addplot+[nagstyle] table[x=layer,y=NAG,col sep=comma] {effective_rank/effective-rank-20.csv};
\addplot+[baselinestyle] table[x=layer,y=BASELINE,col sep=comma] {effective_rank/effective-rank-20.csv};
\end{groupplot}

\node[below=0.7cm] at (group c2r5.south) {\pgfplotslegendfromname{sharedlegend}};

\end{tikzpicture}
\caption{Comparison of NAG and baseline across layers for total variance, participation ratio, and effective rank across five aspect-ratio settings. Blue denotes NAG and red denotes the baseline.}
\label{fig:three-metrics-by-aspect-ratio}
\end{figure*}

We first ask whether controlling residual-stream rotations changes the global geometry of the normalized residual representations. To test this hypothesis, Figure~\ref{fig:three-metrics-by-aspect-ratio} analyzes the covariance spectrum of the normalized residual stream \(\bar{R}_l\) entering each layer. Let \(\bar{R}_{l,i} \in \mathbb{R}^d\) denote the normalized residual vector for sample \(i\) at layer \(l\). We compute the centered covariance
\begin{align}
\Sigma_l &= \frac{1}{n} \sum_{i=1}^n \left(\bar{R}_{l,i} - \mu_l\right)\left(\bar{R}_{l,i} - \mu_l\right)^\top\\
\mu_l &= \frac{1}{n} \sum_{i=1}^{n} \bar{R}_{l,i}
\end{align}
We denote the eigenvalues of \(\Sigma_l\) by \({\lambda_{l,j}}_{j=1}^d\).
We summarize this spectrum using three complementary quantities. First, the total variance is
\begin{equation}
\mathrm{TV}_l = \mathrm{Tr}(\Sigma_l) = \sum_{j=1}^d \lambda_{l,j} = \frac{1}{n} \sum_{i=1}^n \left\| \bar{R}_{l,i} - \mu_l \right\|_2^2
\end{equation}
This measures the overall directional dispersion of the normalized residual vectors around their mean direction. Since normalization fixes the norm of each residual vector, \(\|\bar{R}_{l,i}\|_2^2 = d\), we have
\begin{equation}
\mathrm{TV}_l = \frac{1}{n} \sum_{i=1}^n \left\| \bar{R}_{l,i} \right\|_2^2 - \left\| \mu_l \right\|_2^2 = d - \|\mu_l\|_2^2
\end{equation}
Thus, changes in total variance cannot be attributed to changes in representation scale. Instead, lower total variance is exactly equivalent to a larger mean-vector norm, meaning that the normalized residual vectors are more aligned with a shared direction. A reduction in \(\mathrm{TV}_l\) therefore indicates increased directional concentration, rather than collapse or expansion in scale.
Second, we measure the effective rank,
\begin{align}
\mathrm{erank}_l &= \exp\bigg( -\sum_{j=1}^d p_{l,j} \log p_{l,j} \bigg) \\
p_{l,j} &= \frac{ \lambda_{l,j}}{ \sum_{k=1}^d \lambda_{l,k}}
\end{align}
The effective rank estimates the entropy-based number of principal directions carrying substantial variance.
Third, we compute the participation ratio,
\begin{equation}
\mathrm{PR}_l = \frac{ \left( \sum_{j=1}^d \lambda_{l,j} \right)^2 }{ \sum_{j=1}^d \lambda_{l,j}^2 } = \frac{1}{\sum_{j=1}^d p_{l,j}^2}
\end{equation}
Like effective rank, the participation ratio measures how broadly variance is distributed across principal directions, but it is based on the second moment of the normalized eigenvalue distribution.

In the norm-agnostic model, the total variance of \(\bar{R}_l\) decreases with depth, whereas the baseline does not exhibit such a pronounced trend. This suggests that norm-agnostic updates progressively reduce directional dispersion in the residual stream. At the same time, effective rank and participation ratio remain comparatively stable across layers, indicating that this increased concentration does not come from a substantial reduction in the number of active principal directions.

By contrast, the baseline begins with lower effective dimensionality but shows a marked increase in effective rank and participation ratio in later layers. This late-layer expansion indicates that variance becomes spread across an increasingly large set of directions. Such broadening is consistent with our hypothesis that standard pre-norm updates accumulate norm-dependent deviations across layers.

Together, these results support the view that norm-agnostic updates induce a more stable and controlled residual-stream geometry. Under NAG, the normalized residual stream becomes more directionally concentrated, while avoiding the late-layer broadening of the principal spectrum observed in the baseline.

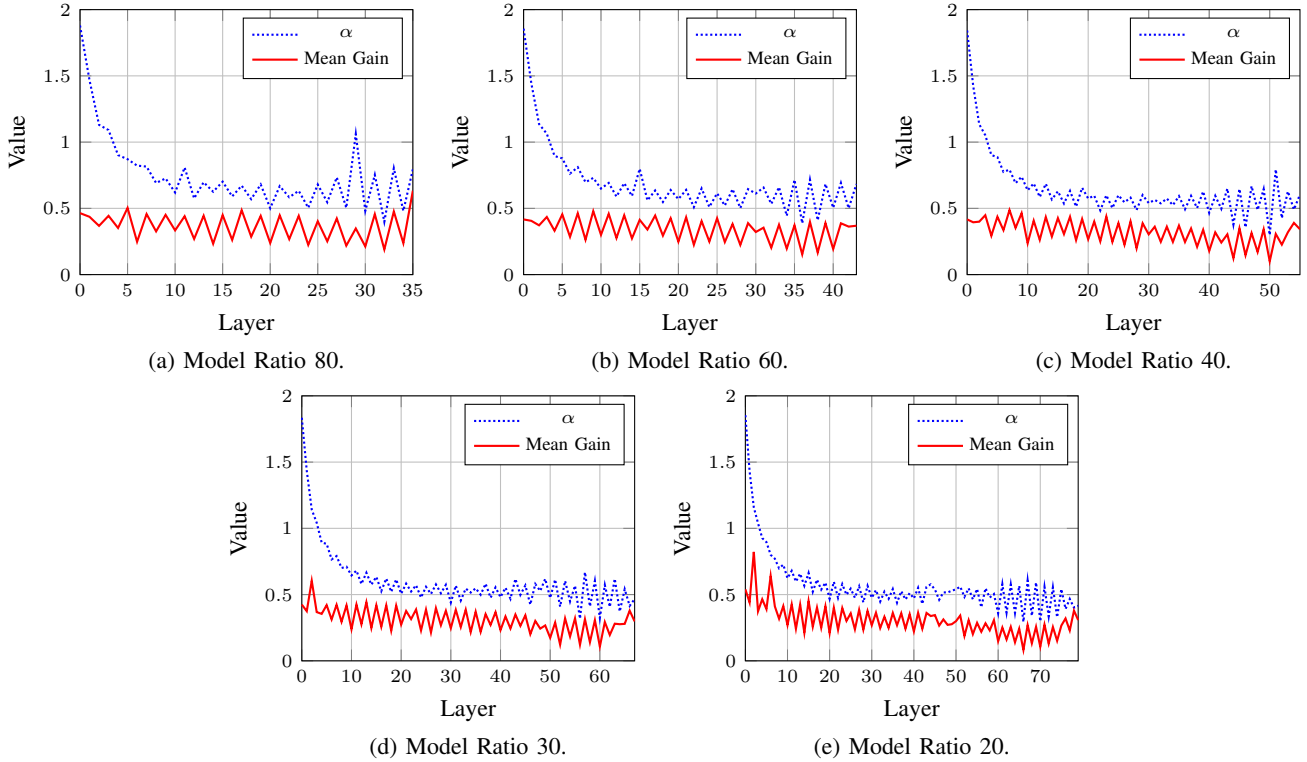
\begin{figure*}[t]
\centering
\begin{subfigure}[t]{0.33\textwidth}
\centering
\begin{tikzpicture}[trim axis left, trim axis right]
\begin{axis}[
    width=\linewidth,
    height=0.85\linewidth,
    xlabel={Layer},
    ylabel={Value},
    xmin=0,
    xmax=35,
    ymin=0,
    ymax=2,
    grid=both,
    legend pos=north east,
    legend style={font=\scriptsize},
    tick label style={font=\scriptsize},
    label style={font=\small},
    xtick distance=5,
]
\addplot+[thick, densely dotted, mark=none]
table[col sep=comma, x={layer_number}, y={nag_out_scale}]
{alpha_and_avg_gain/alpha_and_avg_gain_80.csv};
\addlegendentry{$\alpha$}
\addplot+[thick, solid, mark=none]
table[col sep=comma, x={layer_number}, y={avg_norm_att_score}]
{alpha_and_avg_gain/alpha_and_avg_gain_80.csv};
\addlegendentry{Mean Gain}
\end{axis}
\end{tikzpicture}
\caption{Model Ratio 80.}
\label{fig:alpha-gain-80}
\end{subfigure}
\hspace{-0.02\textwidth}
\begin{subfigure}[t]{0.33\textwidth}
\centering
\begin{tikzpicture}[trim axis left, trim axis right]
\begin{axis}[
    width=\linewidth,
    height=0.85\linewidth,
    xlabel={Layer},
    ylabel={Value},
    xmin=0,
    xmax=43,
    ymin=0,
    ymax=2,
    grid=both,
    legend pos=north east,
    legend style={font=\scriptsize},
    tick label style={font=\scriptsize},
    label style={font=\small},
    xtick distance=5,
]
\addplot+[thick, densely dotted, mark=none]
table[col sep=comma, x={layer_number}, y={nag_out_scale}]
{alpha_and_avg_gain/alpha_and_avg_gain_60.csv};
\addlegendentry{$\alpha$}
\addplot+[thick, solid, mark=none]
table[col sep=comma, x={layer_number}, y={avg_norm_att_score}]
{alpha_and_avg_gain/alpha_and_avg_gain_60.csv};
\addlegendentry{Mean Gain}
\end{axis}
\end{tikzpicture}
\caption{Model Ratio 60.}
\label{fig:alpha-gain-60}
\end{subfigure}
\hspace{-0.02\textwidth}
\begin{subfigure}[t]{0.33\textwidth}
\centering
\begin{tikzpicture}[trim axis left, trim axis right]
\begin{axis}[
    width=\linewidth,
    height=0.85\linewidth,
    xlabel={Layer},
    ylabel={Value},
    xmin=0,
    xmax=55,
    ymin=0,
    ymax=2,
    grid=both,
    legend pos=north east,
    legend style={font=\scriptsize},
    tick label style={font=\scriptsize},
    label style={font=\small},
    xtick distance=10,
]
\addplot+[thick, densely dotted, mark=none]
table[col sep=comma, x={layer_number}, y={nag_out_scale}]
{alpha_and_avg_gain/alpha_and_avg_gain_40.csv};
\addlegendentry{$\alpha$}

\addplot+[thick, solid, mark=none]
table[col sep=comma, x={layer_number}, y={avg_norm_att_score}]
{alpha_and_avg_gain/alpha_and_avg_gain_40.csv};
\addlegendentry{Mean Gain}
\end{axis}
\end{tikzpicture}
\caption{Model Ratio 40.}
\label{fig:alpha-gain-40}
\end{subfigure}
\makebox[\textwidth][c]{
\begin{subfigure}[t]{0.33\textwidth}
\centering
\begin{tikzpicture}[trim axis left, trim axis right]
\begin{axis}[
    width=\linewidth,
    height=0.85\linewidth,
    xlabel={Layer},
    ylabel={Value},
    xmin=0,
    xmax=67,
    ymin=0,
    ymax=2,
    grid=both,
    legend pos=north east,
    legend style={font=\scriptsize},
    tick label style={font=\scriptsize},
    label style={font=\small},
    xtick distance=10,
]
\addplot+[thick, densely dotted, mark=none]
table[col sep=comma, x={layer_number}, y={nag_out_scale}]
{alpha_and_avg_gain/alpha_and_avg_gain_30.csv};
\addlegendentry{$\alpha$}
\addplot+[thick, solid, mark=none]
table[col sep=comma, x={layer_number}, y={avg_norm_att_score}]
{alpha_and_avg_gain/alpha_and_avg_gain_30.csv};
\addlegendentry{Mean Gain}
\end{axis}
\end{tikzpicture}
\caption{Model Ratio 30.}
\label{fig:alpha-gain-30}
\end{subfigure}
\hspace{-0.02\textwidth}
\begin{subfigure}[t]{0.33\textwidth}
\centering
\begin{tikzpicture}[trim axis left, trim axis right]
\begin{axis}[
    width=\linewidth,
    height=0.85\linewidth,
    xlabel={Layer},
    ylabel={Value},
    xmin=0,
    xmax=79,
    ymin=0,
    ymax=2,
    grid=both,
    legend pos=north east,
    legend style={font=\scriptsize},
    tick label style={font=\scriptsize},
    label style={font=\small},
    xtick distance=10,
]
\addplot+[thick, densely dotted, mark=none]
table[col sep=comma, x={layer_number}, y={nag_out_scale}]
{alpha_and_avg_gain/alpha_and_avg_gain_20.csv};
\addlegendentry{$\alpha$}
\addplot+[thick, solid, mark=none]
table[col sep=comma, x={layer_number}, y={avg_norm_att_score}]
{alpha_and_avg_gain/alpha_and_avg_gain_20.csv};
\addlegendentry{Mean Gain}
\end{axis}
\end{tikzpicture}
\caption{Model Ratio 20.}
\label{fig:alpha-gain-20}
\end{subfigure}
}
\caption{Layer-wise values of \(\alpha\) and Mean Gain across model depths. The dotted curve shows \(\alpha\), while the solid curve shows Mean Gain. We observe that the mean gain is effectively flat across layers, implying that, on average, under our NAG residual stream, each layer contributes equally to rotating the residual stream.}
\label{fig:alpha-and-avg-gain}
\end{figure*}

Having examined the geometry of the residual stream itself, we next ask whether the learned NAG parameters produce comparable layer-wise contributions across depth. Although we initialize \(\alpha_l\) according to Eq.~\eqref{eq:init_gain}, we leave them learnable during training. Figure~\ref{fig:alpha-and-avg-gain} shows the final learned values of \(\alpha_l\) across layers for all ablated models. We report both the maximum possible gain, \(\alpha_l\), and the average realized gain, \(\alpha_l m_l(\bar R_l)\), at each layer. The average realized gain remains fairly constant across depth. Since this gain directly determines the angular update imposed on the residual stream, this suggests that NAG encourages layers to contribute comparable amounts of residual-stream rotation throughout the network. This is consistent with Figure~\ref{fig:angular-change}, where deeper NAG models exhibit an approximately linear increase in cumulative rotation with depth.

\begin{figure*}[t]
\centering
\begin{subfigure}{0.19\linewidth}
\centering
\includegraphics[width=\linewidth]{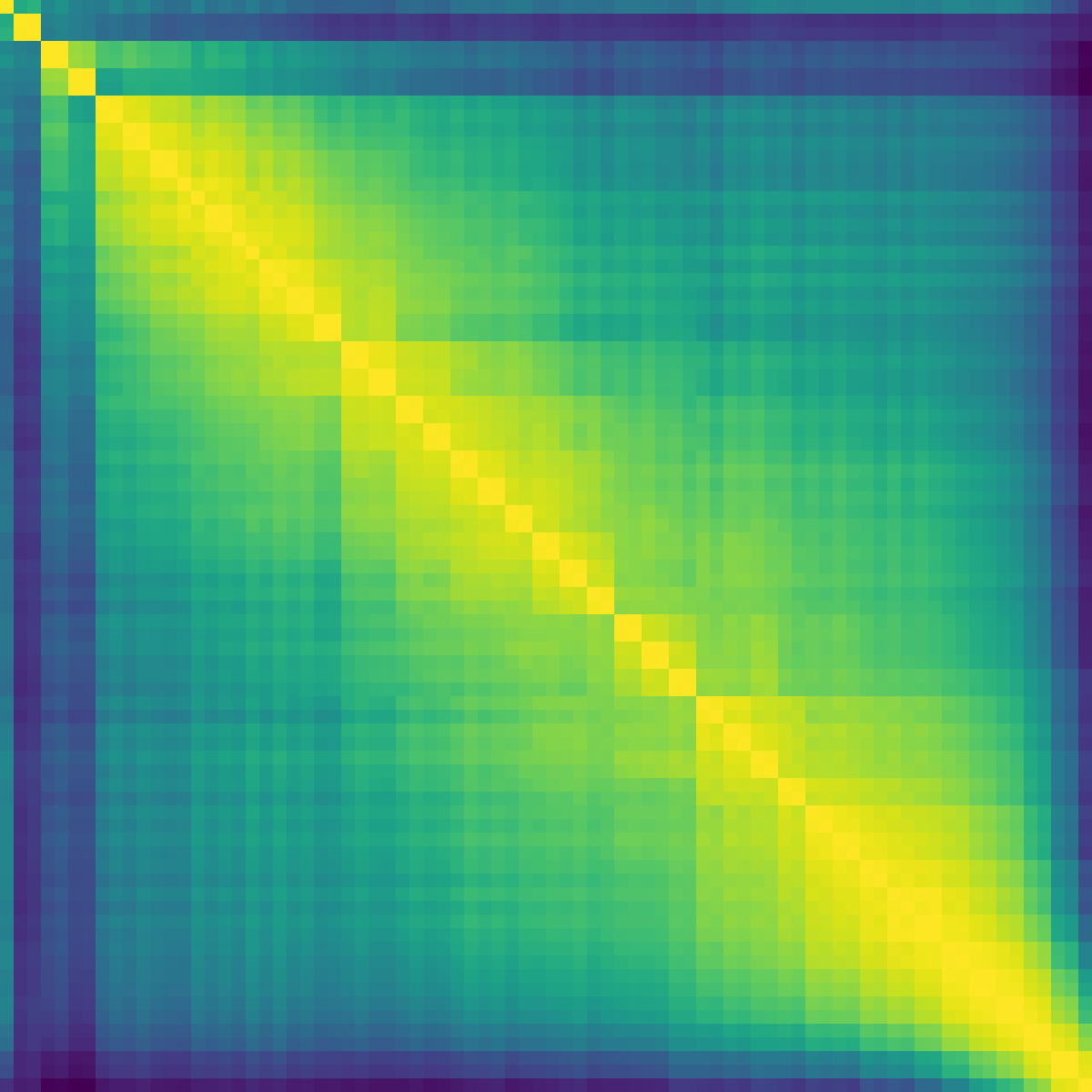}
\caption{Model ratio 20.}
\end{subfigure}
\hfill
\begin{subfigure}{0.19\linewidth}
\centering
\includegraphics[width=\linewidth]{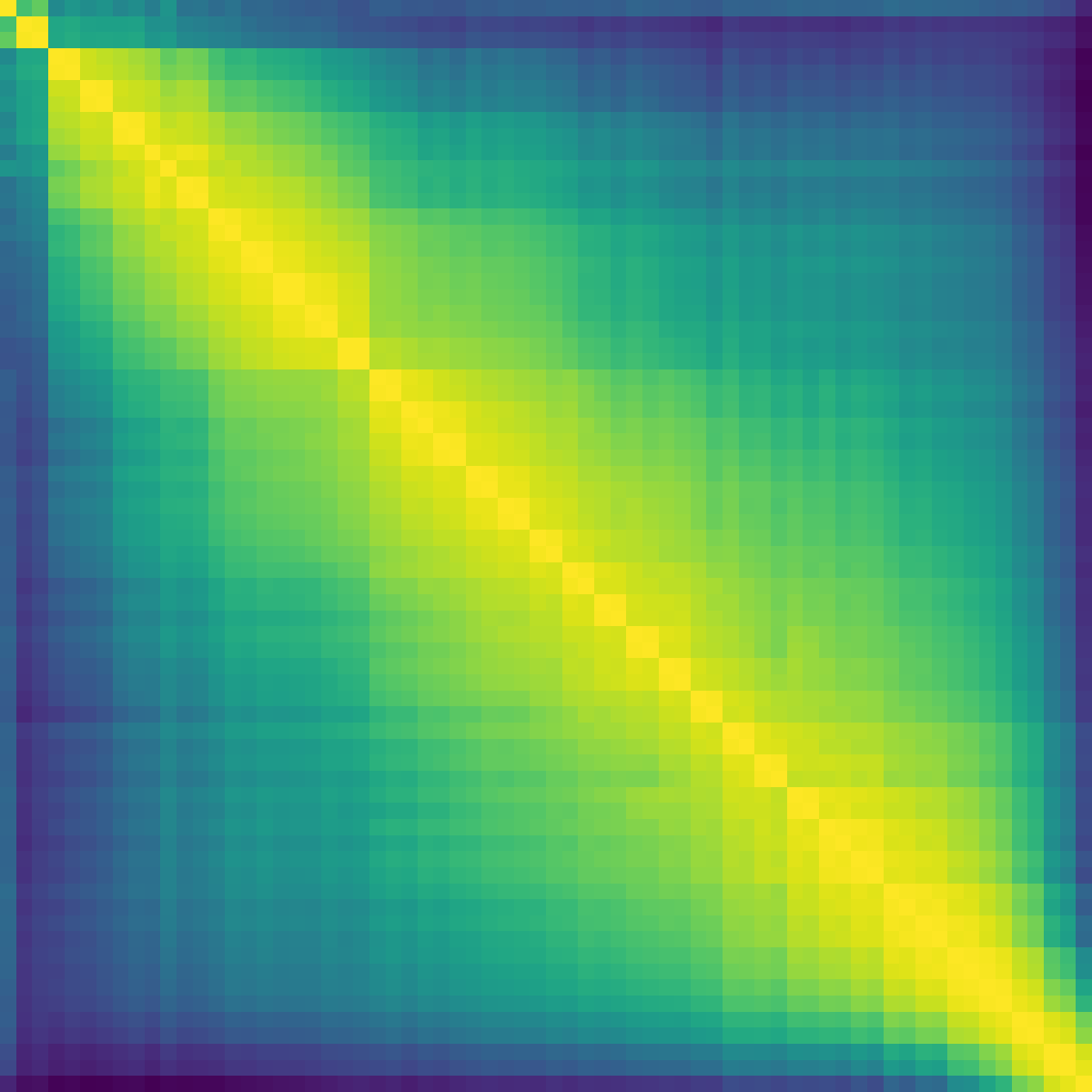}
\caption{Model ratio 30.}
\end{subfigure}
\hfill
\begin{subfigure}{0.19\linewidth}
\centering
\includegraphics[width=\linewidth]{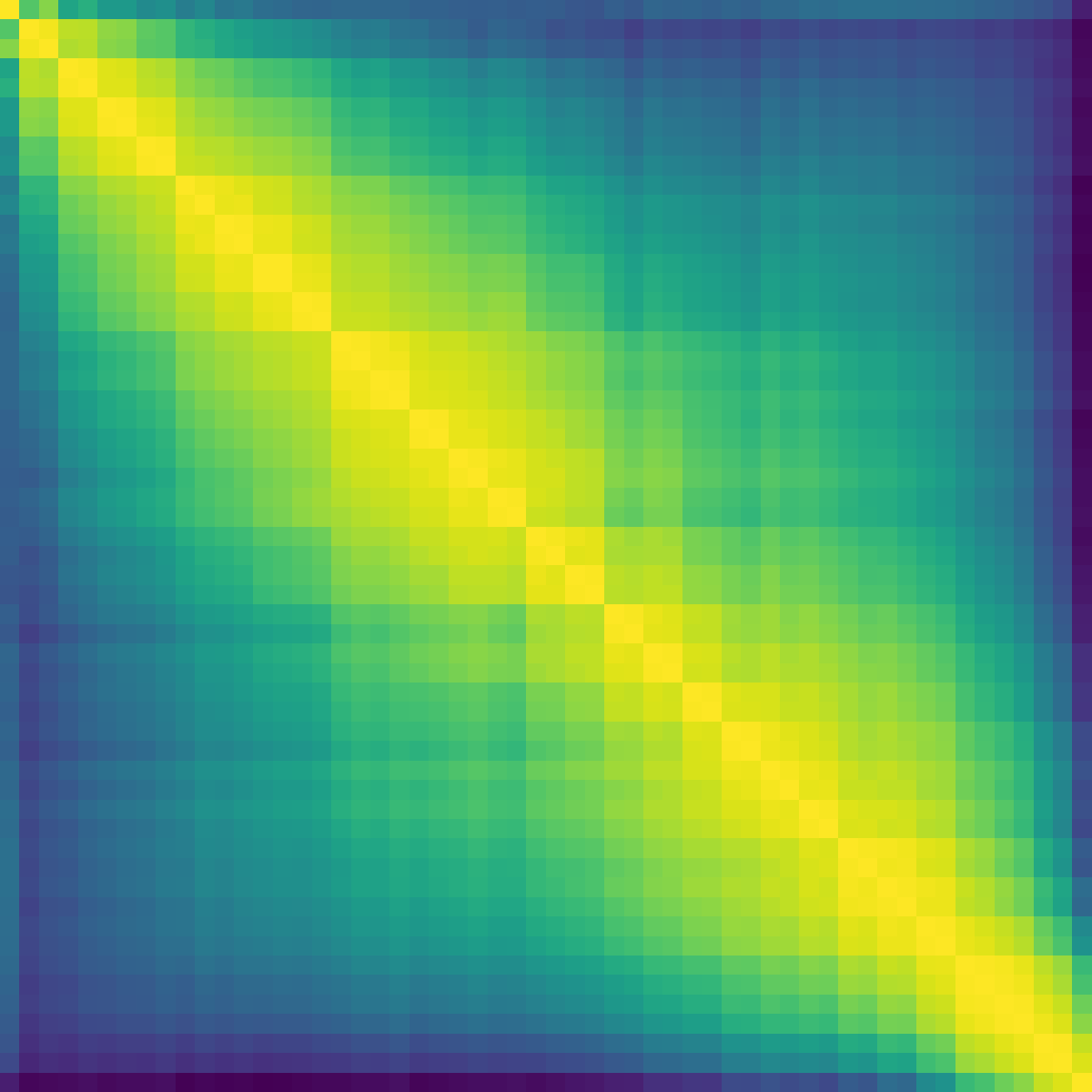}
\caption{Model ratio 40.}
\end{subfigure}
\hfill
\begin{subfigure}{0.19\linewidth}
\centering
\includegraphics[width=\linewidth]{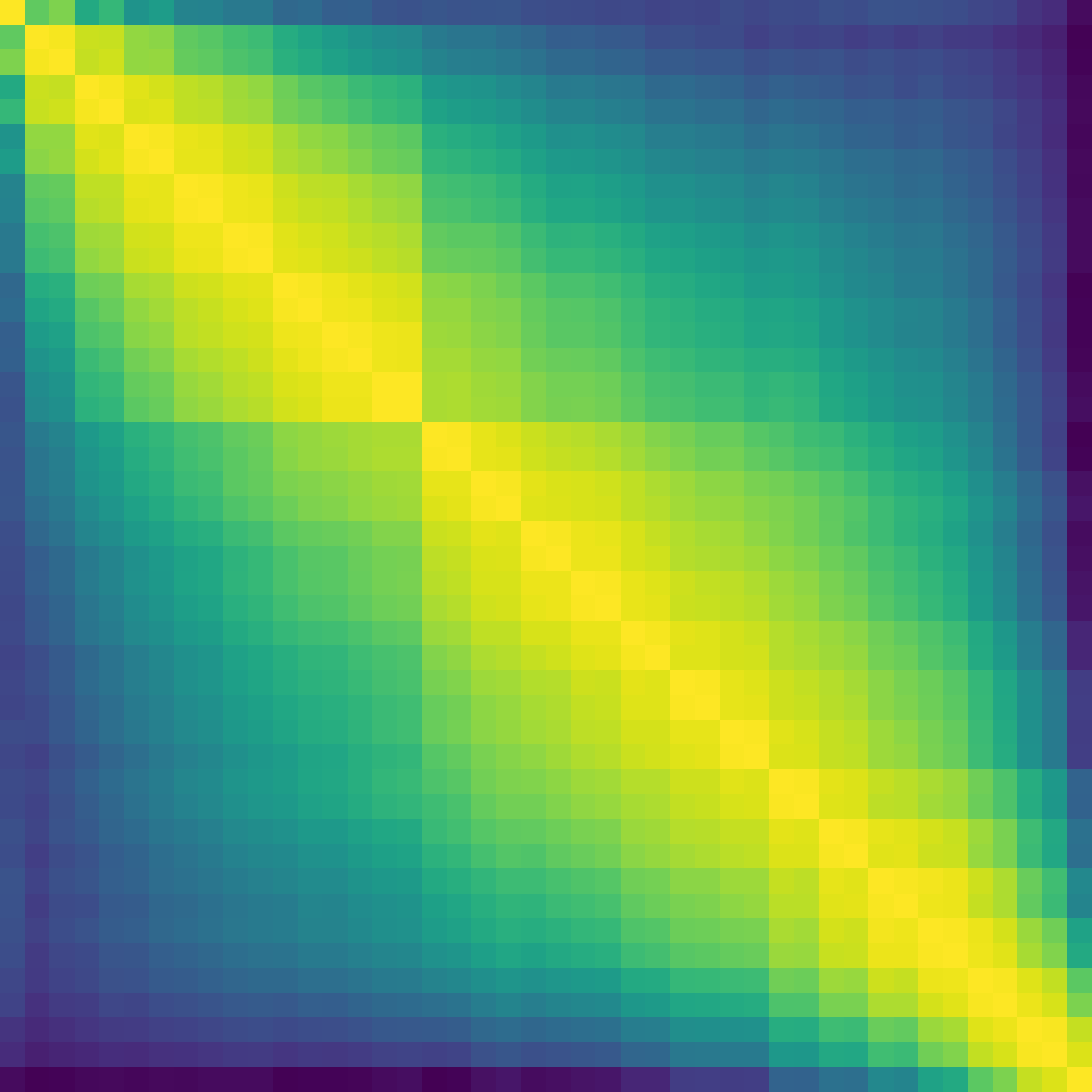}
\caption{Model ratio 60.}
\end{subfigure}
\hfill
\begin{subfigure}{0.19\linewidth}
\centering
\includegraphics[width=\linewidth]{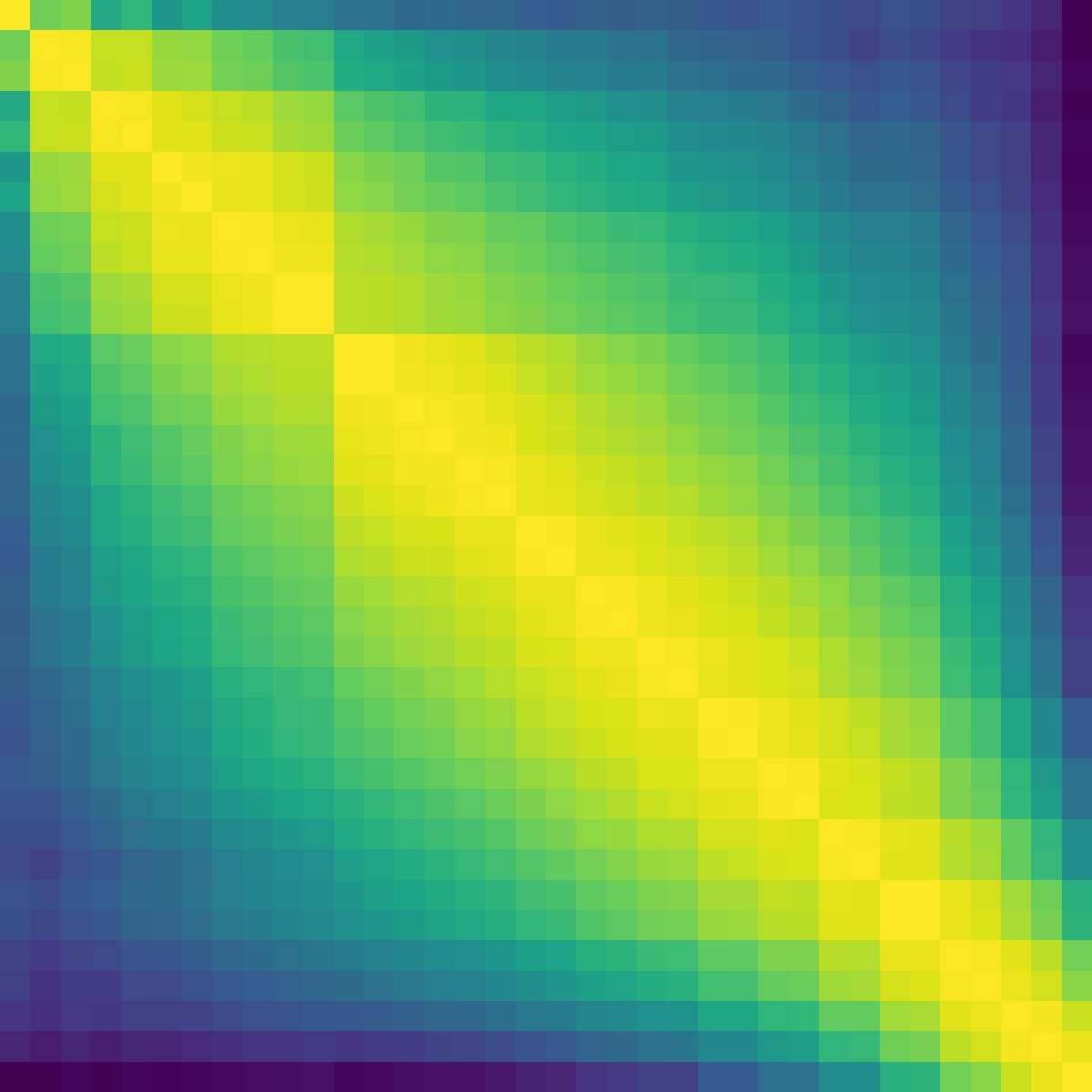}
\caption{Model ratio 80.}
\end{subfigure}
\caption{Cosine similarity between layer-wise preferred directions for each ablated model. Preferred directions are computed as weighted averages of normalized residual stream activations, with norm modulator outputs used as weights. In each heatmap, the top-left corner corresponds to the lowest layer and the bottom-right corner corresponds to the final layer. The pattern shows local correlations which are similar to a sliding window in the directions selected by the norm-modulator, showing that the norm-modulator tends to choose to operate in similar subspaces serially through depth.}
\label{fig:preferred-directions}

\end{figure*}

The previous analysis shows that the realized update magnitudes are balanced across depth. We next examined which regions each layer is tuned to, and whether these regions differ across layers. To do this, we used the normalized residual stream \(\bar{R}\) and computed a weighted average direction over an entire training token batch \(B\), using the norm modulator outputs as weights. Specifically, for each layer \(l\), we computed the preferred direction \(\vec{p_{rl}}\)
\begin{equation}
\vec{p}_{rl} = \frac{\sum_{i=0}^{B-1} m_l(\bar{R}_{il})\bar{R}_{il}}{\sum_{i=0}^{B-1} m_l(\bar{R}_{il})}
\end{equation}
We then measured the cosine similarity between these vectors across all pairs of layers. The resulting heatmaps in Figure~\ref{fig:preferred-directions} show a clear local structure across depth: nearby layers tend to have more similar preferred directions, suggesting that adjacent layers operate on related regions of representation space. Outside this local window, the preferred directions become less similar, indicating that later layers do not simply reuse the same regions throughout the network. This suggests a possible state-machine interpretation of the NAG forward pass. Each layer acts as a conditional transition operator: when a token lies near the layer's preferred region, the layer applies a meaningful rotation; otherwise, its contribution is suppressed. In this view, a token follows a trajectory through representation space, and local decisions to modify or preserve the residual direction influence which subsequent layers become active for that token.

Finally, we return to the inhibitory-effect hypothesis directly by measuring whether angularly similar residual representations can nevertheless have substantially different norms in the baseline model. For each layer, we group residual-stream representations by angular similarity and then measure how much their norms vary within each group. Equivalently, after approximately fixing the angular component of the residual stream, we measure the remaining variation in its radial component.

We estimate norm variation within angularly similar residual representations by clustering the unit-normalized residual embeddings using von Mises--Fisher (vMF) clustering. This choice is natural because vMF distributions are defined on the unit hypersphere and therefore cluster points according to angular similarity rather than Euclidean distance. For each layer, we use a number of clusters equal to the original vocabulary size, namely 32,000. After assigning embeddings to clusters, we normalize each residual norm by the mean norm of its corresponding cluster. We then compute the 10th and 90th percentiles of these normalized norms within each cluster, average the resulting percentile values across all clusters in a layer, and finally average them across layers. The resulting values are reported in Table~\ref{tab:norm-percentiles}. Because the norms are normalized by their cluster mean, values below 1 indicate representations with smaller-than-average norm within an angularly similar group, while values above 1 indicate larger-than-average norm. Across model aspect ratios, the average 10th percentile ranges from 0.723 to 0.809, while the average 90th percentile ranges from 1.209 to 1.300. This means that, even among residual embeddings assigned to the same angular cluster, norms can vary by roughly 20--30\% below or above the cluster mean. These results indicate substantial norm variation among angularly similar representations, consistent with the inhibition hypothesis.

\begin{table}[t]
\centering
{\small
\begin{tabular*}{\linewidth}{@{\extracolsep{\fill}}ccc}
\toprule
Model Ratio & 10th Percentile & 90th Percentile \\
\midrule
80 & 0.723064 & 1.299667 \\
60 & 0.750408 & 1.261029 \\
40 & 0.750328 & 1.269054 \\
30 & 0.809038 & 1.210986 \\
20 & 0.799459 & 1.208804 \\
\bottomrule
\end{tabular*}}
\caption{Average 10th and 90th percentiles of residual-stream norms normalized by the corresponding cluster mean norm. Percentiles are computed within each vMF cluster, averaged across clusters within each layer, and then averaged across layers for each model ratio.}
\label{tab:norm-percentiles}
\end{table}

\subsection{Outlier and Quantization Diagnostics}
\label{subsec:outlier_and_quantization_diagnostics}

Because NAG explicitly controls the size of layer updates to the residual stream, we hypothesized that it may reduce residual-stream activation outliers, which are a major obstacle to near-lossless low-precision quantization of large models \citep{dettmers2022llmint8,xiao2023smoothquant,wei2023outliersuppression}. In standard pre-norm models, a layer can produce an unusually large update that substantially increases the residual-stream norm, injecting high-norm activations that later layers must process. By contrast, NAG constrains the effective rotation applied to the residual stream through the parameter \(\alpha\). Since this rotation directly determines the corresponding norm increase, NAG limits the extent to which any individual layer can inject excessively large updates into the residual stream. This suggests that NAG may be a promising architecture for improving low-precision robustness.

As an initial diagnostic, we examine whether NAG also changes simple distributional properties of the learned expert weights. While this is not a full activation-quantization evaluation, weight centering and worst-case normalized weight magnitudes provide useful preliminary indicators of quantization suitability. For each expert weight matrix, we measure zero-centering using \( |\mu|/\sigma \), where \(\mu\) and \(\sigma\) are the mean and standard deviation of the weights. Smaller values indicate that the weights are more tightly centered around zero, which is favorable for symmetric low-precision quantization. We also report the largest normalized weight magnitude, \(\max |W|/\mathrm{RMS}(W)\), as a simple worst-case outlier diagnostic.

The results are shown in Table~\ref{tab:weight-quant-diagnostics}. NAG substantially improves weight centering relative to the baseline. For the first expert linear layer, the mean $|\mu|/\sigma$ decreases from $8.59\times 10^{-4}$ to $1.50\times 10^{-4}$, while the maximum value decreases from $3.51\times 10^{-3}$ to $5.06\times 10^{-4}$. This corresponds to reductions of 82.5\% and 85.6\%, respectively. A similar trend is observed for the second expert linear layer, where the mean $|\mu|/\sigma$ decreases by 79.7\% and the maximum value decreases by 16.7\%.

NAG also does not increase the worst-case normalized weight magnitude. For the first expert linear layer, $\max |W|/\mathrm{RMS}(W)$ decreases from 19.85 to 13.84, while for the second expert linear layer it decreases slightly from 34.25 to 33.44. Overall, these results provide encouraging preliminary evidence that NAG has weight-distribution properties favorable to low-precision training and inference. In particular, the expert weights are substantially more zero-centered than in the baseline, while also avoiding larger worst-case weight outliers. These diagnostics do not constitute a full quantization evaluation, but they suggest that NAG may be more robust under low-precision deployment.

\begin{table}[t]
\centering
{\small
\begin{tabular*}{\columnwidth}{@{}l@{\hspace{0.6em}}l@{\extracolsep{\fill}}rrr@{}}
\toprule
Metric & Linear & NAG & Baseline & Change \\
\midrule
Mean $|\mu|/\sigma$ & fc1 & $1.50{\times}10^{-4}$ & $8.59{\times}10^{-4}$ & $-82.5\%$ \\
Max $|\mu|/\sigma$ & fc1 & $5.06{\times}10^{-4}$ & $3.51{\times}10^{-3}$ & $-85.6\%$ \\
Max $|W|/\mathrm{RMS}(W)$ & fc1 & 13.84 & 19.85 & $-30.3\%$ \\
\midrule
Mean $|\mu|/\sigma$ & fc2 & $1.28{\times}10^{-4}$ & $6.29{\times}10^{-4}$ & $-79.7\%$ \\
Max $|\mu|/\sigma$ & fc2 & $1.68{\times}10^{-3}$ & $2.02{\times}10^{-3}$ & $-16.7\%$ \\
Max $|W|/\mathrm{RMS}(W)$ & fc2 & 33.44 & 34.25 & $-2.4\%$ \\
\bottomrule
\end{tabular*}}
\caption{Simple weight-distribution diagnostics for NAG and the baseline, aggregated over 174 expert tensors for each expert linear projection. The fc1 and fc2 entries correspond to the first and second linear layers in the MLP experts, respectively. Lower $|\mu|/\sigma$ indicates weights that are more tightly centered around zero. The maximum normalized weight magnitude provides a worst-case outlier diagnostic.}
\label{tab:weight-quant-diagnostics}
\end{table}

\subsection{Attention Sink Analysis}
\label{subsec:attention_sink_analysis}

A recurring phenomenon in transformer language models is the emergence of \emph{attention sinks}, in which one or a few early tokens attract a disproportionately large fraction of attention from many subsequent positions~\citep{xiao2024streamingllm,gu2025attentionsink,barbero2025firsttoken}. This behavior can be useful: early tokens can provide a stable attention anchor, and because softmax attention requires each query to allocate all probability mass somewhere, a sink can also serve as a content-independent location for absorbing attention mass that is not needed for the current computation~\citep{gu2025attentionsink,kaul2025fromattention}. However, the same mechanism may be undesirable when a large fraction of attention is repeatedly assigned to fixed initial positions rather than to contextually relevant tokens. In that case, attention sinks can distort the effective use of the context window and make the model overly dependent on a small set of initial tokens.

Attention sinks are also closely related to activation-outlier phenomena that complicate low-precision deployment. Prior work has linked sink tokens and first-token attention dominance to large hidden-state or residual-stream outliers, and has shown that mitigating sink-associated activations can improve activation quantization~\citep{sun2024massiveactivations,son2024prefixing,kaul2025fromattention}. Since NAG explicitly controls the size of layer updates to the residual stream, we hypothesize that it may reduce the formation or persistence of attention-sink patterns.

\begin{figure}[t]
    \centering
    \begin{subfigure}[t]{0.48\linewidth}
        \centering
        \includegraphics[width=\linewidth]{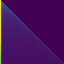}
        \caption{Baseline}
        \label{fig:avg-att-sink-baseline}
    \end{subfigure}
    \hfill
    \begin{subfigure}[t]{0.48\linewidth}
        \centering
        \includegraphics[width=\linewidth]{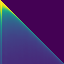}
        \caption{NAG}
        \label{fig:avg-att-sink-nag}
    \end{subfigure}
    \caption{Average post-softmax attention heatmaps across layers for the baseline and NAG models. Rows correspond to query positions, and columns correspond to key positions. Thus, a bright vertical band near the left side indicates that many query positions allocate high attention mass to the same early key tokens. The baseline exhibits a clear attention-sink pattern, whereas NAG substantially reduces this concentration. }
    \label{fig:avg-att-sink}
\end{figure}

To evaluate whether NAG reduces attention-sink behavior, we measure average post-softmax attention weights over the first 64 token positions of each sequence for trained baseline and NAG models with ratio 60. We aggregate attention matrices across evaluation examples and visualize the resulting average attention patterns as heatmaps, normalizing each heatmap by its maximum value. In a model exhibiting an attention sink, these visualizations typically reveal a strong vertical band in the first one or few key-token positions, indicating that many query positions consistently allocate attention to the same initial tokens. In our experiments, the first token is always \texttt{BOS}. This vertical-band pattern is visible across most layers of the baseline model, as shown in Figure~\ref{fig:attention_sink_layerwise_comparison}. Figure~\ref{fig:avg-att-sink} shows the layer-averaged heatmaps from Figure~\ref{fig:attention_sink_layerwise_comparison}, making the reduction in attention-sink behavior under NAG especially apparent.

In contrast, the NAG model does not exhibit a comparable concentrated vertical band. Across layers, attention mass over the first 64 positions remains more broadly distributed rather than collapsing onto the beginning of the sequence. The absence of a dominant vertical stripe in the first-token columns indicates that early tokens do not act as persistent attention attractors to the same extent as in the baseline. These results suggest that, under our evaluation protocol, NAG substantially reduces the prominence of the attention-sink pattern observed in the baseline model.

One possible explanation is that NAG provides an alternative mechanism for suppressing unnecessary attention-layer contributions. In a standard softmax attention layer, attention probability must be allocated across available keys even when the layer has little useful information to contribute. This can encourage the model to allocate probability mass to a stable, low-information sink position. Under NAG, when the norm modulator drives an attention layer's contribution close to zero, the attention output has little influence on the residual stream regardless of where the attention mass is assigned. As a result, the model may have less pressure to use early tokens as persistent ``null'' destinations for attention mass.

\section{A New MoD Approach}
\label{sec:a_new_mod_approach}

Thus far, we have introduced a norm-agnostic residual-stream architecture that improves depth utilization, especially in deeper models. Eq.~\eqref{eq:full_update} reveals an additional consequence of the formulation: the magnitude of a layer's contribution is determined before the expensive layer computation is performed. In the norm-agnostic update, the contribution strength is controlled by the normalized input \(\bar R_l\) through the norm modulator \(m_l(\bar R_l)\). When \(\bar R_l\) lies in a region favored by the modulator, the layer is allowed to contribute strongly to the residual stream; when it lies outside such a region, the layer contribution is suppressed.

This has an important computational implication. Because the modulator depends only on \(\bar R_l\), and not on the layer output \(f_l^\perp(\bar R_l)\), the model can determine the rotation a layer can impose before computing the layer itself. If this rotation is sufficiently small, then the layer can be skipped with limited effect on the residual representation. Unlike standard router-based MoD approaches, this skipping decision is not based on an auxiliary opaque gating signal. It is tied directly to the geometry of the residual stream: the fraction of the layer's maximum possible rotation that would be realized.

Because the layer update is orthogonal to the residual stream, adding the update increases the residual norm according to the Pythagorean theorem. Geometrically, the original residual stream and the layer update form the two legs of a right triangle, while the post-layer residual stream corresponds to the hypotenuse. This geometry allows us to determine the angular displacement of the residual stream from the relative magnitude of the update before computing the full layer output.

Specifically, this angle determines only how far the residual direction moves, not the exact output representation. The precise output still depends on the direction of the orthogonal update. For layer \(l\), the induced rotation can be written as
\begin{equation}
\theta_l = \arctan\left(\alpha_l m_l(\bar R_l)\right)
\end{equation}
where \(\alpha_l\) is the learned layer scale and \(m_l(\bar R_l)\) is the norm modulator output, computed from the normalized residual stream. The maximum possible rotation at layer \(l\) is
\begin{equation}
\theta_l^{\max} = \arctan(\alpha_l)
\end{equation}
Thus, rather than choosing an absolute angular threshold arbitrarily, we normalize the predicted rotation by the maximum rotation possible for that layer. This accounts for the fact that different layers may have different learned values of \(\alpha_l\). We define the normalized rotation score
\begin{equation}
r_l = \frac{\arctan\left(\alpha_l m_l(\bar R_l)\right)}{\arctan(\alpha_l)}
\end{equation}
Given a target skipping-rate threshold \(p_{\mathrm{thres}} \in [0,1]\), the layer-skipping rule is
\begin{equation}
\begin{aligned}
r_l > p_{\mathrm{thres}} &\longrightarrow \text{execute layer } l \\
r_l \le p_{\mathrm{thres}} &\longrightarrow \text{skip layer } l
\end{aligned}
\end{equation}
This criterion compares the rotation induced by a layer to that layer's own maximum possible rotation, rather than applying a fixed angular cutoff uniformly across layers. This accounts for the fact that different layers may learn different values of \(\alpha_l\), and therefore have different maximum rotation capacities. Expressing the decision relative to each layer's own capacity makes the threshold scale-invariant across layers and gives it a direct interpretation: \(r_l\) is the share of the layer's maximum possible rotation that would be applied to the residual stream.

This provides an interpretable mechanism for MoD-style adaptive computation that is integrated directly into the NAG residual update. Before executing a layer, the model can compute how much actual rotation the layer is allowed to contribute relative to its maximum possible rotation, and then decide whether that contribution is large enough to justify the computation. Thus, compute is allocated according to an explicit geometric quantity rather than a separate learned router.

The discontinuity introduced by a hard skip depends on where the threshold lies relative to the layer's maximum rotation. Skipping a layer whose predicted contribution is only \(5\%\) of its maximum rotation is a small perturbation, whereas skipping at \(50\%\) removes a much larger potential update. The latter case creates a sharper transition between applying no update and executing the full layer computation.

To allow for substantial compute reduction while reducing this discontinuity, we introduce a lightweight learned fallback direction. In both the skipped and non-skipped cases, the layer contributes a learned vector \(\vec X_l\), which we interpret as a layer-specific anchor direction. When the layer is executed, the computed update is added around this anchor; when the layer is skipped, the anchor alone provides a cheap substitute for the skipped computation.
\begin{equation}
\hat{f}_l(\bar R_l) =
\begin{cases}
f_l(\bar R_l) + \vec X_l, & r_l > p_{\mathrm{thres}} \\
\vec X_l, & r_l \le p_{\mathrm{thres}}
\end{cases}
\end{equation}
After applying the same centering, orthogonalization, and output normalization as in the base NAG update, the residual update becomes
\begin{equation}
R_{l+1} = R_l + \rho_l \alpha_l m_l(\bar R_l) N_{\mathrm{out}} \left(\hat{f}_l^\perp(\bar R_l)\right)
\end{equation}
Thus, skipping does not remove the layer's contribution entirely. Instead, the skipped layer still contributes a learned, layer-specific direction whose norm is controlled by the modulator, while the expensive computation \(f_l(\bar R_l)\) is evaluated only when the predicted rotation justifies the additional cost.

We apply this skipping mechanism to both MoE and attention blocks. In both cases, we specify a target skip rate, but we enforce this target differently for the two block types.

For MoE layers, we enforce the target skip rate locally at each layer by adjusting a layer-specific threshold. For example, a target skip rate of \(50\%\) means that, within each MoE layer, approximately half of the token positions in a microbatch skip the feed-forward computation during training. This reduces the number of token positions processed by that layer by the same proportion. In practice, the remaining active computation is usually still large enough to avoid entering a bandwidth-limited regime during training.

For attention layers, enforcing the same skip rate independently at every layer caused a substantial increase in loss. We therefore use a global thresholding strategy for attention: rather than requiring each attention layer to skip a fixed fraction of token positions, we enforce the target skip rate only on average across all attention layers through a shared threshold. This allows the model to distribute attention computation unevenly across depth. In our current setup, skipped tokens still produce their corresponding KV elements; that is, skipping affects the attention computation but does not remove those tokens from the KV cache.

In this setting, we found the \(\beta\) parameters in the norm modulator to be particularly useful. Intuitively, each attention layer observes its own distribution of norm-modulator signals, with layer-dependent centers between 0 and 1, which is forced by the norm modulator being constrained to that range. Since the same threshold is applied to all attention layers, a layer whose distribution lies mostly below the threshold will be skipped for most token positions. If the model needs to preserve more computation in that layer, shifting the distribution upward through the sum of sigmoids alone may be difficult or slow. The \(\beta\) parameter provides a more direct mechanism: decreasing \(\beta\) can rapidly raise the modulator outputs, giving the layer a better chance of exceeding the shared threshold and therefore being evaluated.

When training with a target skip rate, the MoD thresholds are updated using the same control principles that we applied in our ZAYA1 series of language models \citet{anthony2025training,washbourne2026zaya1} in balancing experts. Specifically, we use a PID-style controller that compares the desired target skip rate with the empirical skip rate measured on the current batch. Let \(r^\star\) denote the target skip rate and let \(\hat r_t\) denote the observed skip rate at training step \(t\). We define the error as
\begin{equation}
e_t = r^\star - \hat r_t
\end{equation}
This error determines the direction in which the skipping threshold should be adjusted. Since increasing the threshold increases the number of skipped tokens, the threshold is increased when the observed skip rate is below the target and decreased when it is above the target. The resulting PID signal is treated as a pseudo-gradient and applied to the corresponding bias term, allowing the model to track the desired utilization or skipping rate during training. This allows the model to maintain a desired compute budget while still learning how to distribute computation across tokens and layers.

\section{MoD Results}
\label{sec:mod_results}

\subsection{NAG Loss under MoD}
\label{subsec:nag_loss_under_mod}

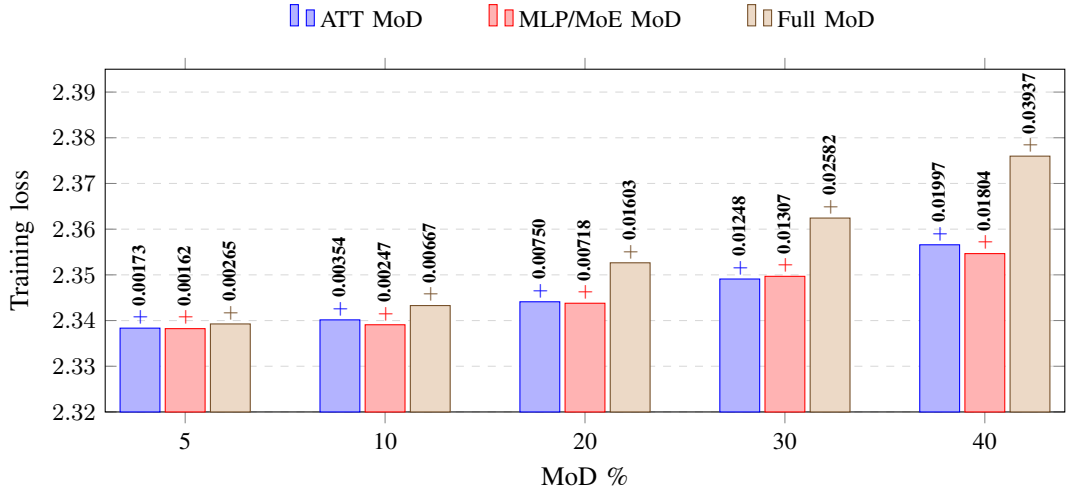
\begin{figure*}[t]
\centering
\begin{tikzpicture}[trim axis left, trim axis right]
\begin{axis}[
    ybar,
    scale only axis=true,
    width=0.7\textwidth,
    height=0.25\textwidth,
    bar width=15pt,
    enlarge x limits=0.10,
    ymin=2.32,
    ymax=2.395,
    ylabel={Training loss},
    ytick distance=0.01,
    xlabel={MoD \%},
    symbolic x coords={5,10,20,30,40},
    xtick=data,
    legend style={
        at={(0.5,1.08)},
        anchor=south,
        legend columns=3,
        draw=none,
        font=\small,
        /tikz/every even column/.append style={column sep=0.8cm}
    },
    tick label style={font=\small},
    ymajorgrids=true,
    grid style={dashed,gray!30},
    clip=false,
    nodes near coords={$+$\,\lossdiff{\pgfplotspointmeta}},
    point meta=explicit symbolic,
    every node near coord/.append style={
        font=\scriptsize,
        rotate=90,
        anchor=west,
        inner sep=1pt
    },
]

\addplot coordinates {
    (5,2.33834)  [0.00173]
    (10,2.34015) [0.00354]
    (20,2.34411) [0.00750]
    (30,2.34909) [0.01248]
    (40,2.35658) [0.01997]
};
\addlegendentry{ATT MoD}

\addplot coordinates {
    (5,2.33823)  [0.00162]
    (10,2.33908) [0.00247]
    (20,2.34379) [0.00718]
    (30,2.34968) [0.01307]
    (40,2.35465) [0.01804]
};
\addlegendentry{MLP/MoE MoD}

\addplot coordinates {
    (5,2.33926)  [0.00265]
    (10,2.34328) [0.00667]
    (20,2.35264) [0.01603]
    (30,2.36243) [0.02582]
    (40,2.37598) [0.03937]
};
\addlegendentry{Full MoD}

\end{axis}
\end{tikzpicture}

\caption{Training losses for attention-only, MLP/MoE-only, and full MoD ablations across different skipping percentages. The no-MoD baseline loss is \(2.33661\). Labels above each bar report the loss increase relative to the baseline. We observe that increasing the MoD skipping rate on a fixed model size (and thus decreasing FLOPs) leads to a small but roughly linear increase in loss. In NAG models, skipping attention leads to roughly the same loss increase as skipping MLP/MoE.}
\label{fig:att-moe-mod-loss}
\end{figure*}

To validate our MoD approach, we first ablate which block types are skipped using the norm-agnostic model with width-to-depth ratio 60. We evaluate attention-only skipping, MLP/MoE-only skipping, and skipping both block types simultaneously, with target skip rates of \(5\%\), \(10\%\), \(20\%\), \(30\%\), and \(40\%\). All runs use the same architecture and training setup described above. Figure~\ref{fig:att-moe-mod-loss} shows that loss increases gradually as the skip rate increases. Attention-only and MLP/MoE-only skipping produce similar loss degradation across the evaluated rates, and the combined setting remains stable even when both block types are skipped. These results indicate that, under NAG, both attention and MLP/MoE blocks can be skipped token-wise without catastrophic degradation.

Next, we compare NAG-MoD against a non-norm-agnostic MoD baseline. In this baseline, a learned linear router produces a gating signal that determines whether the corresponding layer computation should be executed for each token. Specifically, for layer \(l\), the router computes
\begin{equation}
g_l^{\mathrm{MoD}} = \sigma(\bar{R}_l \cdot w_l + b_l)
\end{equation}
As in the NAG setting, a threshold determines which token positions execute the layer and which skip it. The gating signal is also used to modulate the output of the corresponding layer. We use local thresholds for MoE layers and a global threshold for attention layers, matching the NAG-MoD setup.

Figure~\ref{fig:nag-baseline-loss} compares training losses for NAG-MoD and the router-based baseline across all evaluated skip rates. NAG-MoD achieves substantially lower loss at every MoD level. For example, at \(40\%\) MoD, NAG increases loss by only \(0.03937\), whereas the baseline incurs an increase of \(0.10647\), a difference of \(0.06710\). The advantage of NAG remains consistent across the full range of skip rates. These results suggest that tying MoD decisions to residual-stream geometry provides a substantially better training-time skipping criterion than a separate learned router. The annotations below the bar plot report the realized compute savings, along with the theoretical savings expected for longer sequences, where attention costs become increasingly dominant.

\begin{figure*}[t]
\centering
\begin{tikzpicture}[trim axis left, trim axis right]
\begin{axis}[
    ybar,
    scale only axis=true,
    width=0.7\textwidth,
    height=0.25\textwidth,
    bar width=20pt,
    enlarge x limits=0.12,
    ymin=2.32,
    ymax=2.50,
    ylabel={Training loss},
    ytick distance=0.02,
    xlabel={MoD \%},
    symbolic x coords={5,10,20,30,40},
    xtick=data,
    legend style={
        at={(0.5,1.12)},
        anchor=south,
        legend columns=2,
        draw=none,
        /tikz/every even column/.append style={column sep=0.8cm}    
    },
    tick label style={font=\small},
    ymajorgrids=true,
    grid style={dashed,gray!30},
    clip=false,
    nodes near coords={$+$\,\lossdiff{\pgfplotspointmeta}},
    point meta=explicit symbolic,
    every node near coord/.append style={
        font=\scriptsize,
        rotate=90,
        anchor=west,
        inner sep=1pt
    },
]

\addplot+[ybar, fill=blue!35!white, draw=blue!60!black] coordinates {
    (5,2.33926)  [0.00265]
    (10,2.34328) [0.00667]
    (20,2.35264) [0.01603]
    (30,2.36243) [0.02582]
    (40,2.37598) [0.03937]
};
\addlegendentry{NAG}

\addplot+[ybar, fill=red!35!white, draw=red!60!black] coordinates {
    (5,2.37196)  [0.03535]
    (10,2.37881) [0.04220]
    (20,2.39734) [0.06073]
    (30,2.41632) [0.07971]
    (40,2.44308) [0.10647]
};
\addlegendentry{BASELINE}

\end{axis}
\end{tikzpicture}

\vspace{0.75em}

\small
\begin{tabular*}{0.62\textwidth}{@{\extracolsep{\fill}}lccccc}
\toprule
& \multicolumn{5}{c}{MoD \%} \\
\cmidrule(lr){2-6}
Compute savings & 5 & 10 & 20 & 30 & 40 \\
\midrule
Realized savings & 4.47\% & 8.95\% & 17.89\% & 26.84\% & 35.79\% \\
Long-sequence limit & 5.00\% & 10.00\% & 20.00\% & 30.00\% & 40.00\% \\
\bottomrule
\end{tabular*}

\caption{Training losses for NAG and baseline. The no-MoD NAG-60 loss is $2.33661$, and the corresponding BASELINE loss is $2.35539$. Labels above each bar report the loss increase relative to the no-MoD NAG model. The table below reports the corresponding realized compute savings for both training and inference. The long-sequence limit row reports the theoretical savings expected when attention costs dominate total computation, and the generation of KV elements for every token becomes negligible. The realized saving is smaller because when we skip attention we still compute the KV-cache element for the skipped token so not all of the compute in the attention block is avoided.}

\label{fig:nag-baseline-loss}
\end{figure*}

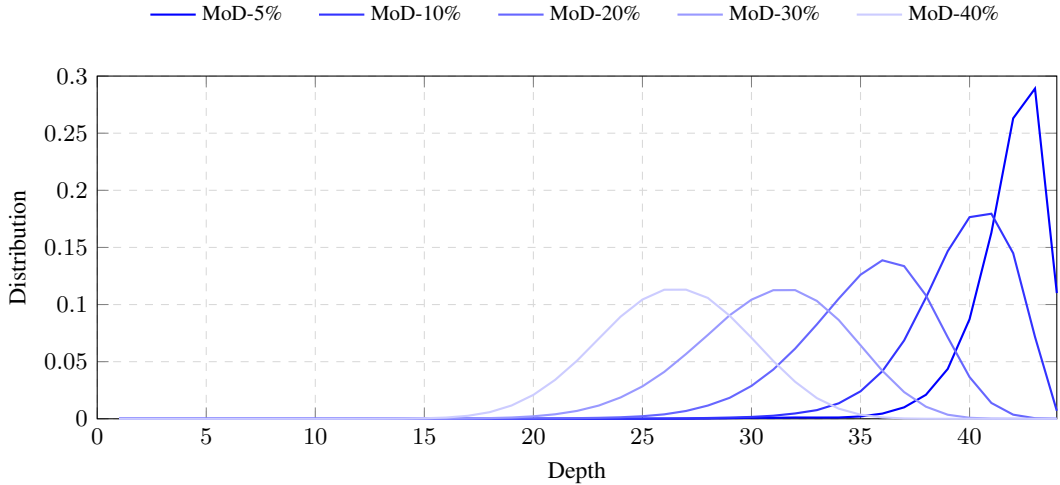
\begin{figure*}[t]
\centering
\begin{tikzpicture}[trim axis left, trim axis right]
\begin{axis}[
    scale only axis=true,
    width=0.7\textwidth,
    height=0.25\textwidth,
    xlabel={Depth},
    ylabel={Distribution},
    xmin=0.0,
    ymin=0.0,
    ymax=0.3,
    xmax=44,
    ytick distance=0.05,
    ymajorgrids=true,
    xmajorgrids=true,
    grid style={dashed,gray!30},
    tick label style={font=\small},
    label style={font=\small},
    cycle list name=nagcolors,
    scaled y ticks=false,
    yticklabel style={
        /pgf/number format/fixed,
        /pgf/number format/precision=2,
    },
    legend style={
        at={(0.5,1.12)},
        anchor=south,
        legend columns=5,
        draw=none,
        fill=white,
        fill opacity=0.85,
        text opacity=1,
        font=\footnotesize,
        /tikz/every even column/.append style={column sep=0.4cm},
        cells={anchor=west}
    },
    clip=false,
    unbounded coords=jump,
]

\addplot+[mark=none, thick]
table[col sep=comma, x=Depth, y={MOD-5}] {Depth_distribution.csv};
\addlegendentry{MoD-5\%}

\addplot+[mark=none, thick]
table[col sep=comma, x=Depth, y={MOD-10}] {Depth_distribution.csv};
\addlegendentry{MoD-10\%}

\addplot+[mark=none, thick]
table[col sep=comma, x=Depth, y={MOD-20}] {Depth_distribution.csv};
\addlegendentry{MoD-20\%}

\addplot+[mark=none, thick]
table[col sep=comma, x=Depth, y={MOD-30}] {Depth_distribution.csv};
\addlegendentry{MoD-30\%}

\addplot+[mark=none, thick]
table[col sep=comma, x=Depth, y={MOD-40}] {Depth_distribution.csv};
\addlegendentry{MoD-40\%}

\end{axis}
\end{tikzpicture}

\caption{Distribution of model depths seen by token across different target MoD percentages. We observe that the tokens form a distribution of depths that they see, implying that some tokens receive substantially more compute than others, indicating the model is learning to use its depth adaptively. We also observe that the distribution over depth appears approximately Gaussian for high skipping rates. The reason for this distribution and how to understand its variance is left to future work.}
\label{fig:depth-distribution}
\end{figure*}

The average number of skipped tokens is kept fixed across MoE layers. For attention layers, however, we apply a global threshold, as described above, which results in different MoD percentages across depth. In Figure~\ref{fig:mod-rates-attention} of the appendix, we report the resulting compression rates for all attention layers, where skipped tokens retain their corresponding KV elements. Here, in Figure~\ref{fig:depth-distribution}, we show the distribution of depths seen by each token for all evaluated MoD percentages. These distributions show that some tokens pass through only a few layers, whereas others propagate through the full model, supporting the interpretation that tokens are allocated different amounts of computation depending on their complexity -- i.e., that MoD implements an effective adaptive compute scheme.

\subsection{ISO-FLOP Ablations}
\label{subsec:iso_flop_ablations}

While Mixture-of-Depths (MoD) has been studied previously, it has not yet been widely adopted in open-source production models. A key reason is that layer skipping typically comes at the cost of model quality: under most nonzero skipping rates, the resulting model is generally less capable than its full-depth counterpart. As a result, MoD is usually viewed as a post-training mechanism for trading accuracy for reduced compute, rather than as a natural component of pretraining. This makes it difficult to justify using MoD during pretraining unless there is a clear compensating benefit.

A more compelling perspective is to ask whether, for the same KV-cache budget and the same number of parameters, one can keep the total training compute fixed while reallocating compute from the forward pass toward training on more tokens. If this tradeoff produces a model with comparable capability under an iso-compute training budget, then MoD becomes an attractive choice. In that setting, one obtains an equally capable model after training while also gaining a cheaper forward pass per token, and hence more efficient inference. Moreover, in most MoD settings, the effective number of active parameters used during the forward pass is reduced as well. Because NAG substantially reduces the loss degradation associated with MoD, we believe our norm-agnostic architecture is a particularly strong candidate for testing whether this tradeoff can make MoD viable during pretraining.

We designed a new set of experiments to test this hypothesis. Since different MoD percentages change the per-token training cost, iso-compute training runs process different numbers of tokens. To make this comparison cleaner, we replaced the cosine learning-rate schedule with a constant learning rate of \(7.5 \times 10^{-4}\). For each MoD setting, we computed the number of iterations and training tokens required to match the same total training FLOPs, and then measured the training loss at that compute-matched point. The results are reported in Table~\ref{tab:mod-isocompute-results}. In addition to the achieved loss, we report the loss difference relative to the \(0\%\) MoD baseline and the effective number of executed parameters used in the forward pass, which decreases as the MoD percentage increases.

The results suggest that moderate MoD rates are already justifiable during pretraining under an iso-compute budget. In particular, the \(20\%\) MoD run essentially matches the baseline without MoD, with a loss difference of only \(0.00055\), while using fewer active forward-pass parameters. This suggests that MoD rates around \(20\%\), and likely up to roughly \(25\%\), may provide a favorable tradeoff, since loss differences on the order of a few thousandths are close to the noise level of these runs. Even at \(40\%\) MoD, the loss increases by only \(0.01854\), while the active forward-pass parameter count drops from \(156.00\)M to \(105.54\)M. Thus, although aggressive skipping begins to incur a measurable loss penalty, it may still be attractive in settings where lower per-token compute is valuable. Overall, these experiments provide evidence that NAG makes MoD viable as a pretraining-time scaling strategy, with \(20\)--\(25\%\) MoD emerging as a practical regime for iso-compute training.

\begin{table*}[t]
\centering
\small
\begin{tabular*}{0.75\textwidth}{@{\extracolsep{\fill}}lcccccc}
\toprule
& \multicolumn{6}{c}{MoD \%} \\
\cmidrule(lr){2-7}
& 0 & 5 & 10 & 20 & 30 & 40 \\
\midrule
Training iterations & 16000 & 16749 & 17572 & 19486 & 21869 & 24916 \\
Training tokens (B) & 50 & 52.34 & 54.91 & 60.90 & 68.34 & 77.86 \\
Achieved loss       & 2.38783 & 2.38583 & 2.38659 & 2.38838 & 2.39965 & 2.40637 \\
Loss diff.          & 0.0 & -0.002 & -0.00124 & 0.00055 & 0.01182 & 0.01854 \\
Active forward-pass params (M) & 156.00 & 149.70 & 143.39 & 130.78 & 118.16 & 105.54 \\
\bottomrule
\end{tabular*}
\caption{Iso-compute training results across different MoD percentages. The \(0\%\) MoD column corresponds to the baseline with no MoD, trained for 16K iterations, or 50B tokens. For each nonzero MoD percentage, the number of iterations and training tokens is increased to match the same total training compute budget. We report the achieved training loss, the loss difference relative to the \(0\%\) MoD run, and the effective number of active parameters used in the forward pass. Negative or near-zero loss differences indicate that the MoD run matches or improves over the \(0\%\) MoD baseline under the same training-compute budget, while reducing the number of executed forward-pass parameters.}
\label{tab:mod-isocompute-results}
\end{table*}

\section{Discussion}
\label{sec:discussion}

In this work, we identified residual-stream norm growth as a key limitation in deep residual architectures. As the residual norm grows, later layer updates become small relative to the current representation, limiting their ability to meaningfully modify the residual stream. Since each parameter in the model has the same compute and memory cost, regardless of its position, we argue that it should have an equal effect on the residual stream representation. More broadly, we speculate that a maximally efficient architecture should satisfy this effective parameter equality property; namely, that the marginal impact of removing a parameter should be roughly equal, on average, for all comparable parameters. Since large residual streams cause poor utilization of the deeper layers, we therefore argue that this represents a subtle but important loss of model capacity relative to its potential. 

To solve this, we introduced the Norm-AGnostic (NAG) model, a residual architecture that separates magnitude from direction in the residual stream, and constrains each layer to act through a controlled angular rotation, while tracking and preserving the norm in a separate stream. This allows deeper layers to remain effective without depending on the current residual-stream norm.

Empirically, we show that NAG improves training loss over matched baselines, with the largest gains in deeper models. Our analyses show that NAG reduces uncontrolled norm accumulation, preserves meaningful representation rotation across depth, and improves weight-distribution behavior by producing more zero-centered expert weights without increasing worst-case normalized weight magnitudes. Because each layer output is normalized, orthogonalized, and modulated before being added to the residual stream, NAG also provides a natural mechanism for controlling layer-output activation norms, making it a strong candidate for reducing activation outliers that complicate quantization and thus performant inference. 

More speculatively, the fact that our NAG approach appears to remove residual stream outliers and substantially ameliorates attention sink implies that these phenomena are not intrinsically useful themselves but are instead symptoms of a deeper pathology in the norms of standard residual transformers. In the case of residual stream outliers, it is plausible that these emerge as a direct consequence of the repeated addition of correlated inputs to the residual stream, effectively corresponding to a kind of power iteration of the weights through depth. For attention sink, we believe that NAG provides an alternative way for the network to address its need for null attention — instead of assigning attention mass to the sink token, the model can instead use the norm-modulator to effectively skip the attention layer entirely by outputting a tiny rotation into the residual stream. NAG thus implicitly serves as a kind of input-dependent gating on attention, which has been shown to similarly reduce attention sink \citep{qiu2026gated}.

Our NAG architecture further provides a principled and interpretable basis for Mixture-of-Depths. The norm modulator predicts a layer's maximum relative contribution before the expensive layer computation is performed, enabling token-wise skipping based on residual-stream geometry rather than an opaque learned router. This makes MoD decisions easier to interpret: the routing variable has a direct geometric meaning, measuring how much actual rotation the layer is expected to contribute to the residual stream compared to its maximum potential. As a result, MoD becomes not only a compute-saving mechanism, but also a window into how computation is allocated across tokens and layers.

Most importantly, our iso-FLOP experiments provide, to our knowledge, the first strong justification for using MoD during pretraining. Under a fixed training-compute budget, skipped computation can be reinvested into additional tokens while preserving parameter count and KV-cache size. Moderate MoD rates match the \(0\%\) MoD baseline loss while reducing executed forward-pass parameters and per-token cost. It should be noted that this is the most stringent test of MoD since we are keeping parameter counts fixed and the marginal impact of training on more tokens declines over the course of training. If we were to compare iso-flop and iso-token while increasing the number of parameters, as is typically done in studies of MoE vs dense models, our MoD depth scaling would achieve a similarly decisive advantage. This means that our MoD method is not merely an inference-time tradeoff, but rather makes viable a novel depth-scaling axis for pretraining. 

One interesting feature of NAG is that it removes the final normalization layer, commonly applied to the residual stream before decoding. We observe that this normalization discards vital information about the \emph{gain}, or effective prediction temperature, which is encoded in the norm of the residual stream. By contrast, NAG preserves this information and gives the residual-stream norm a functional interpretation.

In NAG, confidence in the next-token prediction is reflected by an increase in the norm of the residual stream. This increase occurs only when the layers make a strong contribution—that is, when they encounter embeddings aligned with the preferred directions specified by the norm modulator. If the token moving through the forward pass has no clear connection to the features the layers are specialized to detect, the resulting embedding will have a smaller norm at decoding time. This corresponds to a higher effective temperature, reflecting greater uncertainty in the prediction.

Taken together, these results suggest that our norm-agnostic architecture is a promising direction for scaling language models in depth. By controlling residual-stream geometry, stabilizing layer contributions, reducing norm-related pathologies, enabling interpretable adaptive-depth behavior, and supporting pretraining-compatible MoD, NAG provides a unified architectural framework for building deeper, more compute-efficient, and potentially more quantization-friendly models. 

Looking forward more broadly, we believe our NAG residual stream architecture unlocks scaling sparsity in depth. When combined with MoE, this gives us the option to be simultaneously sparse in both width and depth. Determining the optimal ratio of these two forms of sparsities for training capability and final inference performance remains an open question. Moreover, we hypothesize that our NAG architecture allows us to revisit and develop ideas around repeatedly iterating through a fixed ‘instruction set’ of layers \citep{csordas2024moeut,geiping2026scaling} indefinitely. Such an architecture could support unbounded scaling of inference compute at a fixed parameter budget, enabling vector ``programs'' of indefinite and dynamic length to be executed directly in the residual stream without relying on chain of thought reasoning, which must pass through a discrete token bottleneck. 


\bibliographystyle{tmlr}
\bibliography{main}

@article{ba2016layer,
  title={Layer Normalization},
  author={Ba, Jimmy Lei and Kiros, Jamie Ryan and Hinton, Geoffrey E.},
  journal={arXiv preprint arXiv:1607.06450},
  year={2016}
}

@inproceedings{vaswani2017attention,
  title={Attention Is All You Need},
  author={Vaswani, Ashish and Shazeer, Noam and Parmar, Niki and Uszkoreit, Jakob and Jones, Llion and Gomez, Aidan N. and Kaiser, Lukasz and Polosukhin, Illia},
  booktitle={Advances in Neural Information Processing Systems},
  year={2017}
}

@article{elhage2021mathematical,
  title={A mathematical framework for transformer circuits},
  author={Elhage, Nelson and Nanda, Neel and Olsson, Catherine and Henighan, Tom and Joseph, Nicholas and Mann, Ben and Askell, Amanda and Bai, Yuntao and Chen, Anna and Conerly, Tom and others},
  journal={Transformer Circuits Thread},
  volume={1},
  number={1},
  pages={12},
  year={2021}
}

@article{tokpanov2024zyda,
  title={Zyda: A 1.3 t dataset for open language modeling},
  author={Tokpanov, Yury and Millidge, Beren and Glorioso, Paolo and Pilault, Jonathan and Ibrahim, Adam and Whittington, James and Anthony, Quentin},
  journal={arXiv preprint arXiv:2406.01981},
  year={2024}
}

@article{tokpanov2024zyda2,
  title={Zyda-2: a 5 trillion token high-quality dataset},
  author={Tokpanov, Yury and Glorioso, Paolo and Anthony, Quentin and Millidge, Beren},
  journal={arXiv preprint arXiv:2411.06068},
  year={2024}
}

@article{anthony2025training,
  title={Training Foundation Models on a Full-Stack AMD Platform: Compute, Networking, and System Design},
  author={Anthony, Quentin and Tokpanov, Yury and Szot, Skyler and Rajagopal, Srivatsan and Medepalli, Praneeth and Golubeva, Anna and Shyam, Vasu and Washbourne, Robert and Iyer, Rishi and Chaurasia, Ansh and others},
  journal={arXiv preprint arXiv:2511.17127},
  year={2025}
}

@inproceedings{xiong2020layernorm,
  title={On Layer Normalization in the Transformer Architecture},
  author={Xiong, Ruibin and Yang, Yunchang and He, Di and Zheng, Kai and Zheng, Shuxin and Xing, Chen and Zhang, Huishuai and Lan, Yanyan and Wang, Liwei and Liu, Tie-Yan},
  booktitle={International Conference on Machine Learning},
  year={2020}
}

@article{wang2022deepnet,
  title={DeepNet: Scaling Transformers to 1,000 Layers},
  author={Wang, Hongyu and Ma, Shuming and Dong, Li and Huang, Shaohan and Zhang, Dongdong and Wei, Furu},
  journal={arXiv preprint arXiv:2203.00555},
  year={2022}
}

@article{sun2025curse,
  title={The Curse of Depth in Large Language Models},
  author={Sun, Wenfang and Song, Xinyuan and Li, Pengxiang and Yin, Lu and Zheng, Yefeng and Liu, Shiwei},
  journal={arXiv preprint arXiv:2502.05795},
  year={2025}
}

@article{dehghani2018universal,
  title={Universal Transformers},
  author={Dehghani, Mostafa and Gouws, Stephan and Vinyals, Oriol and Uszkoreit, Jakob and Kaiser, Lukasz},
  journal={arXiv preprint arXiv:1807.03819},
  year={2018}
}

@article{raposo2024mixture,
  title={Mixture-of-Depths: Dynamically Allocating Compute in Transformer-Based Language Models},
  author={Raposo, David and Ritter, Sam and Richards, Blake and Lillicrap, Timothy and Humphreys, Peter Conway and Santoro, Adam},
  journal={arXiv preprint arXiv:2404.02258},
  year={2024}
}

@article{bae2025mixture,
  title={Mixture-of-Recursions: Learning Dynamic Recursive Depths for Adaptive Token-Level Computation},
  author={Bae, Sangmin and Kim, Yujin and Bayat, Reza and Kim, Sungnyun and Ha, Jiyoun and Schuster, Tal and Fisch, Adam and Harutyunyan, Hrayr and Ji, Ziwei and Courville, Aaron and Yun, Se-Young},
  journal={arXiv preprint arXiv:2507.10524},
  year={2025}
}

@article{srivastava2015training,
  title={Training very deep networks},
  author={Srivastava, Rupesh K and Greff, Klaus and Schmidhuber, J{\"u}rgen},
  journal={Advances in neural information processing systems},
  volume={28},
  year={2015}
}

@article{greff2016highway,
  title={Highway and residual networks learn unrolled iterative estimation},
  author={Greff, Klaus and Srivastava, Rupesh K and Schmidhuber, J{\"u}rgen},
  journal={arXiv preprint arXiv:1612.07771},
  year={2016}
}

@article{jordan2024muon,
  title={Muon: An optimizer for hidden layers in neural networks, 2024},
  author={Jordan, Keller and Jin, Yuchen and Boza, Vlado and Jiacheng, You and Cesista, Franz and Newhouse, Laker and Bernstein, Jeremy},
  journal={URL https://kellerjordan. github. io/posts/muon},
  volume={6},
  number={3},
  pages={4},
  year={2024}
}

@article{brown2020language,
  title={Language models are few-shot learners},
  author={Brown, Tom and Mann, Benjamin and Ryder, Nick and Subbiah, Melanie and Kaplan, Jared D and Dhariwal, Prafulla and Neelakantan, Arvind and Shyam, Pranav and Sastry, Girish and Askell, Amanda and others},
  journal={Advances in neural information processing systems},
  volume={33},
  pages={1877--1901},
  year={2020}
}

@article{hoffmann2022training,
  title={Training compute-optimal large language models},
  author={Hoffmann, Jordan and Borgeaud, Sebastian and Mensch, Arthur and Buchatskaya, Elena and Cai, Trevor and Rutherford, Eliza and Casas, DDL and Hendricks, Lisa Anne and Welbl, Johannes and Clark, Aidan and others},
  journal={arXiv preprint arXiv:2203.15556},
  volume={10},
  year={2022}
}

@article{achiam2023gpt,
  title={Gpt-4 technical report},
  author={Achiam, Josh and Adler, Steven and Agarwal, Sandhini and Ahmad, Lama and Akkaya, Ilge and Aleman, Florencia Leoni and Almeida, Diogo and Altenschmidt, Janko and Altman, Sam and Anadkat, Shyamal and others},
  journal={arXiv preprint arXiv:2303.08774},
  year={2023}
}

@article{liu2024deepseek,
  title={Deepseek-v3 technical report},
  author={Liu, Aixin and Feng, Bei and Xue, Bing and Wang, Bingxuan and Wu, Bochao and Lu, Chengda and Zhao, Chenggang and Deng, Chengqi and Zhang, Chenyu and Ruan, Chong and others},
  journal={arXiv preprint arXiv:2412.19437},
  year={2024}
}

@article{figliolia2025compressed,
  title={Compressed Convolutional Attention: Efficient Attention in a Compressed Latent Space},
  author={Figliolia, Tomas and Alonso, Nicholas and Iyer, Rishi and Anthony, Quentin and Millidge, Beren},
  journal={arXiv preprint arXiv:2510.04476},
  year={2025}
}

@article{washbourne2026zaya1,
  title={ZAYA1-8B Technical Report},
  author={Washbourne, Robert and Iyer, Rishi and Figliolia, Tomas and Zheng, Henry and Lorig-Roach, Ryan and Yang, Sungyeon and Yuvraj, Pritish and Anthony, Quentin and Tokpanov, Yury and Yang, Xiao and others},
  journal={arXiv preprint arXiv:2605.05365},
  year={2026}
}

@article{jastrzkebski2017residual,
  title={Residual connections encourage iterative inference},
  author={Jastrz{\k{e}}bski, Stanis{\l}aw and Arpit, Devansh and Ballas, Nicolas and Verma, Vikas and Che, Tong and Bengio, Yoshua},
  journal={arXiv preprint arXiv:1710.04773},
  year={2017}
}

@inproceedings{de2020batch,
  title = {Batch Normalization Biases Residual Blocks Towards the Identity Function in Deep Networks},
  author = {De, Soham and Smith, Samuel L.},
  booktitle = {Advances in Neural Information Processing Systems},
  volume = {33},
  year = {2020},
  url = {https://proceedings.neurips.cc/paper/2020/hash/e6b738eca0e6792ba8a9cbcba6c1881d-Abstract.html}
}

@misc{csordas2025depth,
  title = {Do Language Models Use Their Depth Efficiently?},
  author = {Csord{\'a}s, R{\'o}bert and Manning, Christopher D. and Potts, Christopher},
  year = {2025},
  eprint = {2505.13898},
  archivePrefix = {arXiv},
  primaryClass = {cs.CL},
  url = {https://arxiv.org/abs/2505.13898}
}

@inproceedings{gromov2025unreasonable,
  title = {The Unreasonable Ineffectiveness of the Deeper Layers},
  author = {Gromov, Andrey and Tirumala, Kushal and Shapourian, Hassan and Glorioso, Paolo and Roberts, Daniel A.},
  booktitle = {International Conference on Learning Representations},
  year = {2025},
  url = {https://openreview.net/forum?id=ngmEcEer8a}
}

@inproceedings{jiang2025tracing,
  title = {Tracing Representation Progression: Analyzing and Enhancing Layer-Wise Similarity},
  author = {Jiang, Jiachen and Zhou, Jinxin and Zhu, Zhihui},
  booktitle = {International Conference on Learning Representations},
  year = {2025},
  url = {https://openreview.net/forum?id=vVxeFSR4fU}
}

@inproceedings{men2025shortgpt,
  title = {{S}hort{GPT}: Layers in Large Language Models are More Redundant Than You Expect},
  author = {Men, Xin and Xu, Mingyu and Zhang, Qingyu and Yuan, Qianhao and Wang, Bingning and Lin, Hongyu and Lu, Yaojie and Han, Xianpei and Chen, Weipeng},
  booktitle = {Findings of the Association for Computational Linguistics: ACL 2025},
  month = jul,
  year = {2025},
  address = {Vienna, Austria},
  publisher = {Association for Computational Linguistics},
  pages = {20192--20204},
  doi = {10.18653/v1/2025.findings-acl.1035},
  url = {https://aclanthology.org/2025.findings-acl.1035/}
}

@inproceedings{veit2016residual,
  title = {Residual Networks Behave Like Ensembles of Relatively Shallow Networks},
  author = {Veit, Andreas and Wilber, Michael J. and Belongie, Serge J.},
  booktitle = {Advances in Neural Information Processing Systems},
  volume = {29},
  pages = {550--558},
  year = {2016},
  url = {https://arxiv.org/abs/1605.06431}
}

@article{sajjad2022effect,
  title = {On the Effect of Dropping Layers of Pre-trained Transformer Models},
  author = {Sajjad, Hassan and Dalvi, Fahim and Durrani, Nadir and Nakov, Preslav},
  journal = {Computer Speech \& Language},
  volume = {77},
  pages = {101429},
  month = jan,
  year = {2023},
  doi = {10.1016/j.csl.2022.101429},
  url = {https://www.sciencedirect.com/science/article/pii/S0885230822000596}
}

@inproceedings{fan2020reducing,
  title = {Reducing Transformer Depth on Demand with Structured Dropout},
  author = {Fan, Angela and Grave, Edouard and Joulin, Armand},
  booktitle = {International Conference on Learning Representations},
  year = {2020},
  url = {https://openreview.net/forum?id=SylO2yStDr}
}

@article{kim2024shortened,
  title = {Shortened {LLaMA}: Depth Pruning for Large Language Models with Comparison of Retraining Methods},
  author = {Kim, Bo-Kyeong and Kim, Geonmin and Kim, Tae-Ho and Castells, Thibault and Choi, Shinkook and Shin, Junho and Song, Hyoung-Kyu},
  journal = {arXiv preprint arXiv:2402.02834},
  year = {2024},
  url = {https://arxiv.org/abs/2402.02834}
}

@inproceedings{yang2024laco,
  title = {{L}a{C}o: Large Language Model Pruning via Layer Collapse},
  author = {Yang, Yifei and Cao, Zouying and Zhao, Hai},
  booktitle = {Findings of the Association for Computational Linguistics: EMNLP 2024},
  month = nov,
  year = {2024},
  address = {Miami, Florida, USA},
  publisher = {Association for Computational Linguistics},
  pages = {6401--6417},
  doi = {10.18653/v1/2024.findings-emnlp.372},
  url = {https://aclanthology.org/2024.findings-emnlp.372/}
}

@inproceedings{ding2021cogview,
  title = {{CogView}: Mastering Text-to-Image Generation via Transformers},
  author = {Ding, Ming and Yang, Zhuoyi and Hong, Wenyi and Zheng, Wendi and Zhou, Chang and Yin, Da and Lin, Junyang and Zou, Xu and Shao, Zhou and Yang, Hongxia and Tang, Jie},
  booktitle = {Advances in Neural Information Processing Systems},
  volume = {34},
  pages = {19822--19835},
  year = {2021},
  url = {https://proceedings.neurips.cc/paper/2021/hash/a4d92e2cd541fca87e4620aba658316d-Abstract.html}
}

@article{team2024gemma,
  title={Gemma 2: Improving open language models at a practical size},
  author={Team, Gemma and Riviere, Morgane and Pathak, Shreya and Sessa, Pier Giuseppe and Hardin, Cassidy and Bhupatiraju, Surya and Hussenot, L{\'e}onard and Mesnard, Thomas and Shahriari, Bobak and Ram{\'e}, Alexandre and others},
  journal={arXiv preprint arXiv:2408.00118},
  year={2024}
}

@inproceedings{xiong2020layer,
  title = {On Layer Normalization in the Transformer Architecture},
  author = {Xiong, Ruibin and Yang, Yunchang and He, Di and Zheng, Kai and Zheng, Shuxin and Xing, Chen and Zhang, Huishuai and Lan, Yanyan and Wang, Liwei and Liu, Tie-Yan},
  booktitle = {Proceedings of the 37th International Conference on Machine Learning},
  series = {Proceedings of Machine Learning Research},
  volume = {119},
  pages = {10524--10533},
  year = {2020},
  publisher = {PMLR},
  url = {https://proceedings.mlr.press/v119/xiong20b.html}
}

@article{csordas2024moeut,
  title={Moeut: Mixture-of-experts universal transformers},
  author={Csord{\'a}s, R{\'o}bert and Irie, Kazuki and Schmidhuber, J{\"u}rgen and Potts, Christopher and Manning, Christopher D},
  journal={Advances in Neural Information Processing Systems},
  volume={37},
  pages={28589--28614},
  year={2024}
}

@article{geiping2026scaling,
  title={Scaling up test-time compute with latent reasoning: A recurrent depth approach},
  author={Geiping, Jonas and McLeish, Sean and Jain, Neel and Kirchenbauer, John and Singh, Siddharth and Bartoldson, Brian and Kailkhura, Bhavya and Bhatele, Abhinav and Goldstein, Tom},
  journal={Advances in Neural Information Processing Systems},
  volume={38},
  pages={41340--41391},
  year={2026}
}

@article{qiu2026gated,
  title={Gated attention for large language models: Non-linearity, sparsity, and attention-sink-free},
  author={Qiu, Zihan and Wang, Zekun and Zheng, Bo and Huang, Zeyu and Wen, Kaiyue and Yang, Songlin and Men, Rui and Yu, Le and Huang, Fei and Huang, Suozhi and others},
  journal={Advances in Neural Information Processing Systems},
  volume={38},
  pages={100092--100118},
  year={2026}
}

@inproceedings{devlin2019bert,
title = {{BERT}: Pre-training of Deep Bidirectional Transformers for Language Understanding},
author = {Devlin, Jacob and Chang, Ming-Wei and Lee, Kenton and Toutanova, Kristina},
booktitle = {Proceedings of the 2019 Conference of the North American Chapter of the Association for Computational Linguistics: Human Language Technologies, Volume 1 (Long and Short Papers)},
pages = {4171--4186},
year = {2019},
address = {Minneapolis, Minnesota},
publisher = {Association for Computational Linguistics},
doi = {10.18653/v1/N19-1423},
url = {https://aclanthology.org/N19-1423}
}

@article{elango2026latentmoe,
  title={LatentMoE: Toward Optimal Accuracy per FLOP and Parameter in Mixture of Experts},
  author={Elango, Venmugil and Bhatia, Nidhi and Waleffe, Roger and Shafipour, Rasoul and Asida, Tomer and Khattar, Abhinav and Assaf, Nave and Golub, Maximilian and Guman, Joey and Mitra, Tiyasa and others},
  journal={arXiv preprint arXiv:2601.18089},
  year={2026}
}

@misc{zhai2026exclusiveselfattention,
      title={Exclusive Self Attention}, 
      author={Shuangfei Zhai},
      year={2026},
      eprint={2603.09078},
      archivePrefix={arXiv},
      primaryClass={cs.LG},
      url={https://arxiv.org/abs/2603.09078}, 
}

@misc{sun2026cursedepthlargelanguage,
      title={The Curse of Depth in Large Language Models}, 
      author={Wenfang Sun and Xinyuan Song and Pengxiang Li and Lu Yin and Yefeng Zheng and Shiwei Liu},
      year={2026},
      eprint={2502.05795},
      archivePrefix={arXiv},
      primaryClass={cs.LG},
      url={https://arxiv.org/abs/2502.05795}, 
}

@misc{chen2026postlayernormbackstableexpressive,
      title={Post-LayerNorm Is Back: Stable, ExpressivE, and Deep}, 
      author={Chen Chen and Lai Wei},
      year={2026},
      eprint={2601.19895},
      archivePrefix={arXiv},
      primaryClass={cs.LG},
      url={https://arxiv.org/abs/2601.19895}, 
}

@misc{zhang2019improvingdeeptransformerdepthscaled,
      title={Improving Deep Transformer with Depth-Scaled Initialization and Merged Attention}, 
      author={Biao Zhang and Ivan Titov and Rico Sennrich},
      year={2019},
      eprint={1908.11365},
      archivePrefix={arXiv},
      primaryClass={cs.CL},
      url={https://arxiv.org/abs/1908.11365}, 
}

@inproceedings{he2016deep,
  title     = {Deep Residual Learning for Image Recognition},
  author    = {He, Kaiming and Zhang, Xiangyu and Ren, Shaoqing and Sun, Jian},
  booktitle = {Proceedings of the IEEE Conference on Computer Vision and Pattern Recognition (CVPR)},
  pages     = {770--778},
  month     = jun,
  year      = {2016}
}

@inproceedings{geva2022transformerfeedforward,
  title     = {Transformer Feed-Forward Layers Build Predictions by Promoting Concepts in the Vocabulary Space},
  author    = {Geva, Mor and Caciularu, Avi and Wang, Kevin and Goldberg, Yoav},
  booktitle = {Proceedings of the 2022 Conference on Empirical Methods in Natural Language Processing},
  pages     = {30--45},
  address   = {Abu Dhabi, United Arab Emirates},
  publisher = {Association for Computational Linguistics},
  month     = dec,
  year      = {2022},
  doi       = {10.18653/v1/2022.emnlp-main.3},
  url       = {https://aclanthology.org/2022.emnlp-main.3/}
}

@article{belrose2023eliciting,
  title         = {Eliciting Latent Predictions from Transformers with the Tuned Lens},
  author        = {Belrose, Nora and Ostrovsky, Igor and McKinney, Lev and Furman, Zach and Smith, Logan and Halawi, Danny and Biderman, Stella and Steinhardt, Jacob},
  journal       = {arXiv preprint arXiv:2303.08112},
  year          = {2023},
  eprint        = {2303.08112},
  archivePrefix = {arXiv},
  primaryClass  = {cs.LG},
  doi           = {10.48550/arXiv.2303.08112},
  url           = {https://arxiv.org/abs/2303.08112}
}

@article{xie2023residual,
  title         = {ResiDual: Transformer with Dual Residual Connections},
  author        = {Xie, Shufang and Zhang, Huishuai and Guo, Junliang and Tan, Xu and Bian, Jiang and Awadalla, Hany Hassan and Menezes, Arul and Qin, Tao and Yan, Rui},
  journal       = {arXiv preprint arXiv:2304.14802},
  year          = {2023},
  eprint        = {2304.14802},
  archivePrefix = {arXiv},
  primaryClass  = {cs.CL},
  doi           = {10.48550/arXiv.2304.14802},
  url           = {https://arxiv.org/abs/2304.14802}
}

@inproceedings{bachlechner2021rezero,
  title     = {ReZero is All You Need: Fast Convergence at Large Depth},
  author    = {Bachlechner, Thomas and Majumder, Bodhisattwa Prasad and Mao, Henry and Cottrell, Gary and McAuley, Julian},
  booktitle = {Proceedings of the Thirty-Seventh Conference on Uncertainty in Artificial Intelligence},
  pages     = {1352--1361},
  year      = {2021},
  volume    = {161},
  series    = {Proceedings of Machine Learning Research},
  publisher = {PMLR},
  url       = {https://proceedings.mlr.press/v161/bachlechner21a.html}
}

@inproceedings{touvron2021going,
  title     = {Going Deeper With Image Transformers},
  author    = {Touvron, Hugo and Cord, Matthieu and Sablayrolles, Alexandre and Synnaeve, Gabriel and J{\'e}gou, Herv{\'e}},
  booktitle = {Proceedings of the IEEE/CVF International Conference on Computer Vision (ICCV)},
  pages     = {32--42},
  month     = oct,
  year      = {2021}
}

@inproceedings{brody2023expressivityrole,
  title     = {On the Expressivity Role of {L}ayer{N}orm in Transformers' Attention},
  author    = {Brody, Shaked and Alon, Uri and Yahav, Eran},
  booktitle = {Findings of the Association for Computational Linguistics: ACL 2023},
  pages     = {14211--14221},
  address   = {Toronto, Canada},
  publisher = {Association for Computational Linguistics},
  month     = jul,
  year      = {2023},
  doi       = {10.18653/v1/2023.findings-acl.895},
  url       = {https://aclanthology.org/2023.findings-acl.895/}
}

@article{graves2016adaptive,
  title         = {Adaptive Computation Time for Recurrent Neural Networks},
  author        = {Graves, Alex},
  journal       = {arXiv preprint arXiv:1603.08983},
  year          = {2016},
  eprint        = {1603.08983},
  archivePrefix = {arXiv},
  primaryClass  = {cs.NE},
  doi           = {10.48550/arXiv.1603.08983},
  url           = {https://arxiv.org/abs/1603.08983}
}

@inproceedings{dehghani2019universal,
  title     = {Universal Transformers},
  author    = {Dehghani, Mostafa and Gouws, Stephan and Vinyals, Oriol and Uszkoreit, Jakob and Kaiser, {\L}ukasz},
  booktitle = {International Conference on Learning Representations},
  year      = {2019},
  url       = {https://arxiv.org/abs/1807.03819}
}

@inproceedings{elbayad2020depthadaptive,
  title     = {Depth-Adaptive Transformer},
  author    = {Elbayad, Maha and Gu, Jiatao and Grave, Edouard and Auli, Michael},
  booktitle = {International Conference on Learning Representations},
  year      = {2020},
  url       = {https://openreview.net/forum?id=SJg7KhVKPH}
}

@article{raposo2024mixturedepths,
  title         = {Mixture-of-Depths: Dynamically Allocating Compute in Transformer-Based Language Models},
  author        = {Raposo, David and Ritter, Sam and Richards, Blake and Lillicrap, Timothy and Humphreys, Peter Conway and Santoro, Adam},
  journal       = {arXiv preprint arXiv:2404.02258},
  year          = {2024},
  eprint        = {2404.02258},
  archivePrefix = {arXiv},
  primaryClass  = {cs.LG},
  doi           = {10.48550/arXiv.2404.02258},
  url           = {https://arxiv.org/abs/2404.02258}
}

@inproceedings{xiao2024streamingllm,
  title={Efficient Streaming Language Models with Attention Sinks},
  author={Xiao, Guangxuan and Tian, Yuandong and Chen, Beidi and Han, Song and Lewis, Mike},
  booktitle={International Conference on Learning Representations},
  year={2024}
}

@inproceedings{gu2025attention_sink,
  title={When Attention Sink Emerges in Language Models: An Empirical View},
  author={Gu, Xiangming and Pang, Tianyu and Du, Chao and Liu, Qian and Zhang, Fengzhuo and Du, Cunxiao and Wang, Ye and Lin, Min},
  booktitle={International Conference on Learning Representations},
  year={2025}
}

@inproceedings{bondarenko2023quantizable,
  title     = {Quantizable Transformers: Removing Outliers by Helping Attention Heads Do Nothing},
  author    = {Bondarenko, Yelysei and Nagel, Markus and Blankevoort, Tijmen},
  booktitle = {Advances in Neural Information Processing Systems},
  year      = {2023},
  url       = {https://openreview.net/forum?id=sbusw6LD41}
}

@inproceedings{sun2024massive,
  title     = {Massive Activations in Large Language Models},
  author    = {Sun, Mingjie and Chen, Xinlei and Kolter, J. Zico and Liu, Zhuang},
  booktitle = {Conference on Language Modeling},
  year      = {2024},
  url       = {https://openreview.net/forum?id=F7aAhfitX6}
}

@inproceedings{dettmers2022llmint8,
  title = {{LLM.int8()}: 8-bit Matrix Multiplication for Transformers at Scale},
  author = {Dettmers, Tim and Lewis, Mike and Belkada, Younes and Zettlemoyer, Luke},
  booktitle = {Advances in Neural Information Processing Systems},
  volume = {35},
  pages = {30318--30332},
  year = {2022}
}

@inproceedings{xiao2023smoothquant,
  title = {{SmoothQuant}: Accurate and Efficient Post-Training Quantization for Large Language Models},
  author = {Xiao, Guangxuan and Lin, Ji and Seznec, Mickael and Wu, Hao and Demouth, Julien and Han, Song},
  booktitle = {Proceedings of the 40th International Conference on Machine Learning},
  pages = {38087--38099},
  year = {2023},
  editor = {Krause, Andreas and Brunskill, Emma and Cho, Kyunghyun and Engelhardt, Barbara and Sabato, Sivan and Scarlett, Jonathan},
  volume = {202},
  series = {Proceedings of Machine Learning Research},
  publisher = {PMLR}
}

@inproceedings{wei2023outliersuppression,
  title = {Outlier Suppression+: Accurate Quantization of Large Language Models by Equivalent and Effective Shifting and Scaling},
  author = {Wei, Xiuying and Zhang, Yunchen and Li, Yuhang and Zhang, Xiangguo and Gong, Ruihao and Guo, Jinyang and Liu, Xianglong},
  booktitle = {Proceedings of the 2023 Conference on Empirical Methods in Natural Language Processing},
  pages = {1648--1665},
  year = {2023},
  address = {Singapore},
  publisher = {Association for Computational Linguistics},
  doi = {10.18653/v1/2023.emnlp-main.102},
  url = {https://aclanthology.org/2023.emnlp-main.102/}
}

@inproceedings{gu2025attentionsink,
  title = {When Attention Sink Emerges in Language Models: An Empirical View},
  author = {Gu, Xiangming and Pang, Tianyu and Du, Chao and Liu, Qian and Zhang, Fengzhuo and Du, Cunxiao and Wang, Ye and Lin, Min},
  booktitle = {International Conference on Learning Representations},
  year = {2025},
  url = {https://openreview.net/forum?id=78Nn4QJTEN}
}

@inproceedings{barbero2025firsttoken,
  title = {Why do {LLM}s attend to the first token?},
  author = {Barbero, Federico and Arroyo, Alvaro and Gu, Xiangming and Perivolaropoulos, Christos and Veli{\v{c}}kovi{\'c}, Petar and Pascanu, Razvan and Bronstein, Michael M.},
  booktitle = {Conference on Language Modeling},
  year = {2025},
  url = {https://openreview.net/forum?id=tu4dFUsW5z}
}

@inproceedings{sun2024massiveactivations,
  title = {Massive Activations in Large Language Models},
  author = {Sun, Mingjie and Chen, Xinlei and Kolter, J. Zico and Liu, Zhuang},
  booktitle = {Conference on Language Modeling},
  year = {2024},
  url = {https://openreview.net/forum?id=F7aAhfitX6}
}

@inproceedings{son2024prefixing,
  title = {Prefixing Attention Sinks can Mitigate Activation Outliers for Large Language Model Quantization},
  author = {Son, Seungwoo and Park, Wonpyo and Han, Woohyun and Kim, Kyuyeun and Lee, Jaeho},
  editor = {Al-Onaizan, Yaser and Bansal, Mohit and Chen, Yun-Nung},
  booktitle = {Proceedings of the 2024 Conference on Empirical Methods in Natural Language Processing},
  month = nov,
  year = {2024},
  address = {Miami, Florida, USA},
  publisher = {Association for Computational Linguistics},
  url = {https://aclanthology.org/2024.emnlp-main.134/},
  doi = {10.18653/v1/2024.emnlp-main.134},
  pages = {2242--2252}
}

@inproceedings{kaul2025fromattention,
  title = {From Attention to Activation: Unraveling the Enigmas of Large Language Models},
  author = {Kaul, Prannay and Ma, Chengcheng and Elezi, Ismail and Deng, Jiankang},
  booktitle = {International Conference on Learning Representations},
  year = {2025},
  url = {https://openreview.net/forum?id=IjduZQK8gM}
}
\clearpage

\onecolumn
\appendix
\subsection{MoD Percentages Across Attention Layers}
\label{subsec:mod_percentages_across_attention_layers}

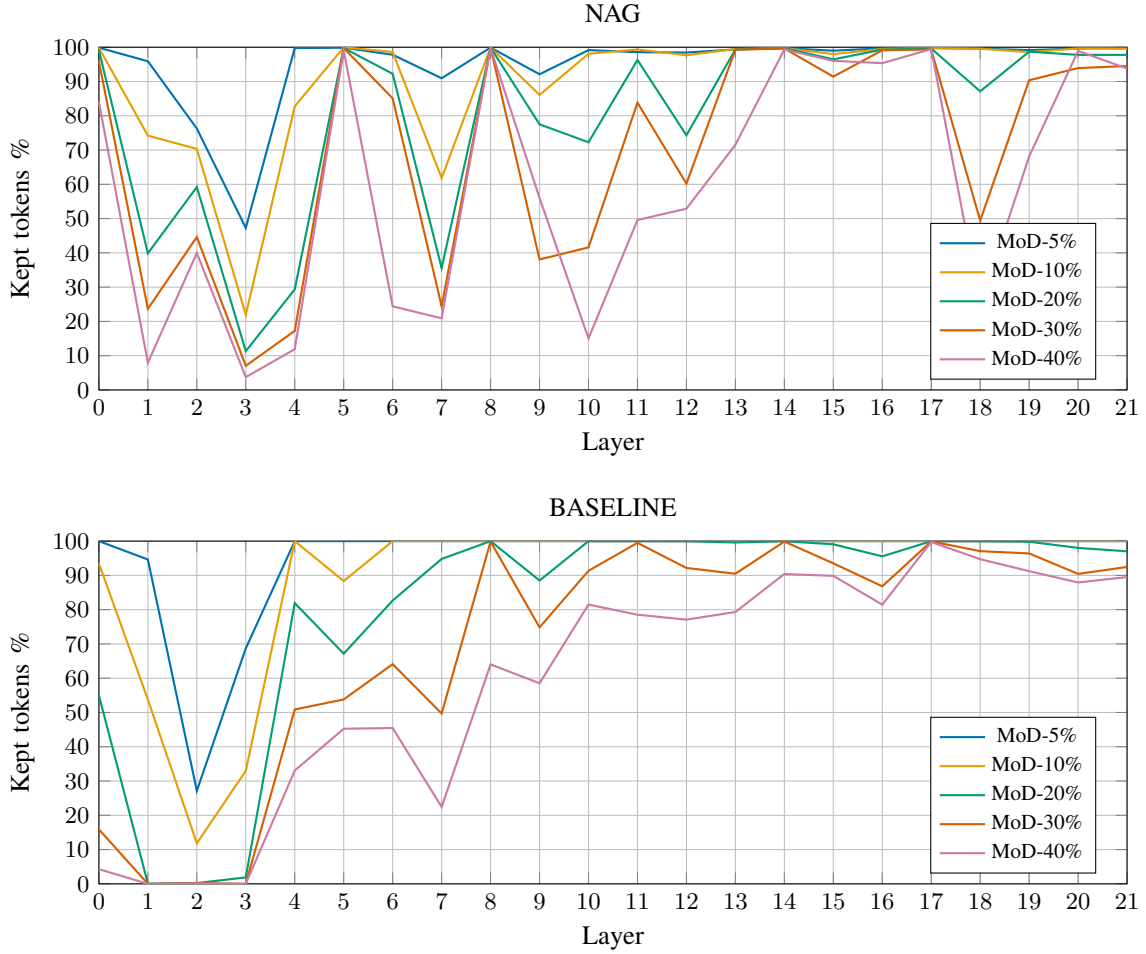
\begin{figure*}[!]
\centering

\begin{tikzpicture}
\begin{groupplot}[
    group style={
        group size=1 by 2,
        vertical sep=2.0cm
    },
    scale only axis=true,
    width=0.75\textwidth,
    height=0.25\textwidth,
    xlabel={Layer},
    ylabel={Kept tokens \%},
    ymin=0,
    ymax=100,
    grid=both,
    legend pos=south east,
    legend style={font=\footnotesize},
    tick label style={font=\small},
    cycle list name=multicolor,
    ytick distance=10,
    xtick distance=1,
    xmin=0,
    xmax=21,
]

\nextgroupplot[
    title={NAG},
]

\addplot table[col sep=comma, x=Layer, y={5}] {attention_skip_rates/MOD_keepkv_nag.csv};
\addlegendentry{MoD-5\%}

\addplot table[col sep=comma, x=Layer, y={10}] {attention_skip_rates/MOD_keepkv_nag.csv};
\addlegendentry{MoD-10\%}

\addplot table[col sep=comma, x=Layer, y={20}] {attention_skip_rates/MOD_keepkv_nag.csv};
\addlegendentry{MoD-20\%}

\addplot table[col sep=comma, x=Layer, y={30}] {attention_skip_rates/MOD_keepkv_nag.csv};
\addlegendentry{MoD-30\%}

\addplot table[col sep=comma, x=Layer, y={40}] {attention_skip_rates/MOD_keepkv_nag.csv};
\addlegendentry{MoD-40\%}

\nextgroupplot[
    title={BASELINE},
]

\addplot table[col sep=comma, x=Layer, y={5}] {attention_skip_rates/MOD_keepkv_baseline.csv};
\addlegendentry{MoD-5\%}

\addplot table[col sep=comma, x=Layer, y={10}] {attention_skip_rates/MOD_keepkv_baseline.csv};
\addlegendentry{MoD-10\%}

\addplot table[col sep=comma, x=Layer, y={20}] {attention_skip_rates/MOD_keepkv_baseline.csv};
\addlegendentry{MoD-20\%}

\addplot table[col sep=comma, x=Layer, y={30}] {attention_skip_rates/MOD_keepkv_baseline.csv};
\addlegendentry{MoD-30\%}

\addplot table[col sep=comma, x=Layer, y={40}] {attention_skip_rates/MOD_keepkv_baseline.csv};
\addlegendentry{MoD-40\%}

\end{groupplot}
\end{tikzpicture}

\caption{MoD rates per attention layer for NAG and BASELINE under different target MoD percentages.}

\label{fig:mod-rates-attention}
\end{figure*}

In Section~\ref{sec:mod_results}, we noted that attention layers are skipped using a global threshold, allowing computation to be allocated unevenly across depth. Figure~\ref{fig:mod-rates-attention} shows the percentage of tokens retained at each attention layer for both NAG and the baseline across different target MoD percentages. At higher MoD rates, several layers retain very few tokens, suggesting that these layers may be candidates for further trimming or removal.

\subsection{Attention Sink in NAG vs. Baseline}
\label{subsec:attention_sink_in_nag_vs_baseline}

Here we present the full set of average attention matrices across all twenty layers for the NAG model and the baseline for the model with ratio 60. The NAG layers exhibit a clear causal attention pattern without a pronounced attention sink. In contrast, many baseline layers concentrate a large fraction of attention mass on the initial sink token, suggesting that these layers rely more heavily on a fixed low-information position rather than distributing attention across contextually relevant tokens. This pattern is consistent with reduced effective use of attention computation in the baseline.

\begin{figure*}[t]
    \centering
    \setlength{\tabcolsep}{1.5pt}
    \renewcommand{\arraystretch}{0.85}

    \newcommand{\attimg}[1]{\includegraphics[width=0.155\textwidth]{#1}}
    \newcommand{\attlbl}[2]{\scriptsize #1 Layer #2}

    \begin{tabular}{@{}cc@{\hspace{0.5em}}cc@{\hspace{0.5em}}cc@{}}

    \attlbl{Baseline}{0} & \attlbl{NAG}{0} &
    \attlbl{Baseline}{1} & \attlbl{NAG}{1} &
    \attlbl{Baseline}{2} & \attlbl{NAG}{2} \\

    \attimg{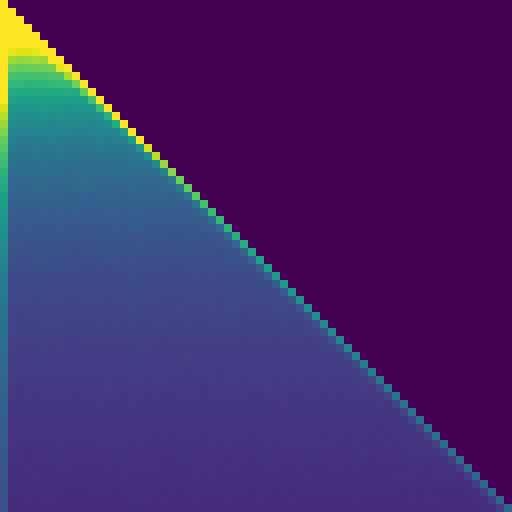} &
    \attimg{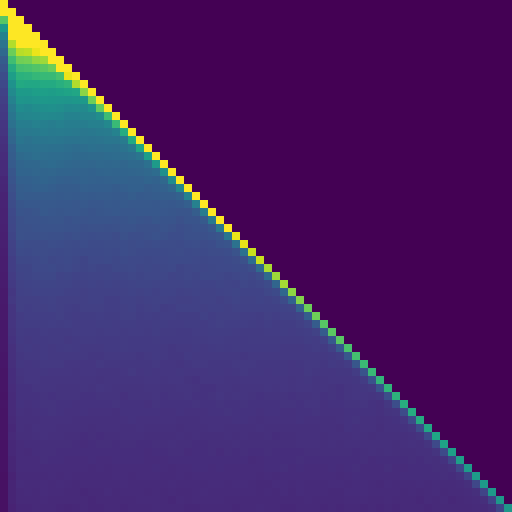} &
    \attimg{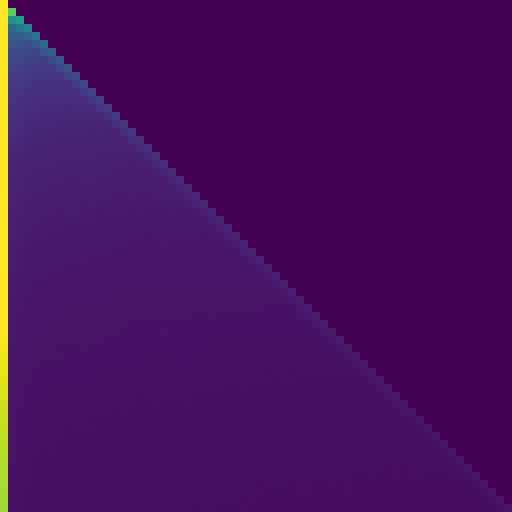} &
    \attimg{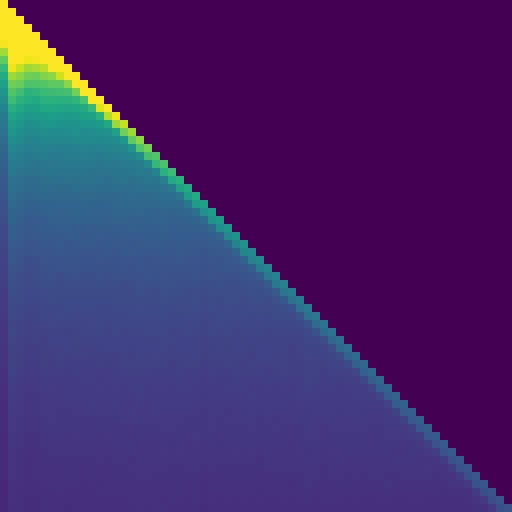} &
    \attimg{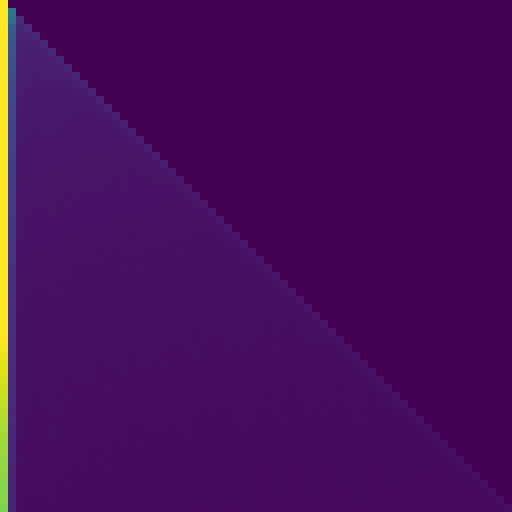} &
    \attimg{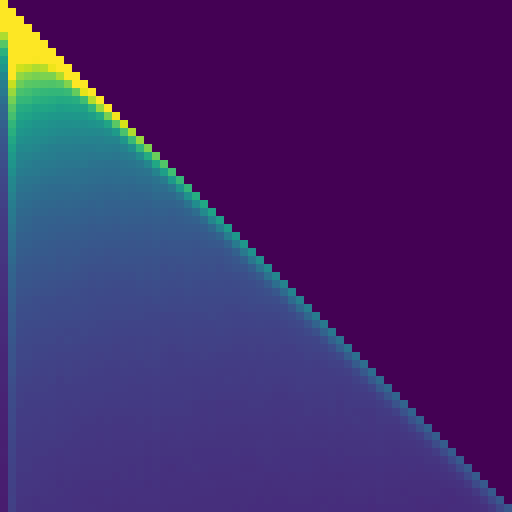} \\[-0.33em]

    \attlbl{Baseline}{3} & \attlbl{NAG}{3} &
    \attlbl{Baseline}{4} & \attlbl{NAG}{4} &
    \attlbl{Baseline}{5} & \attlbl{NAG}{5} \\

    \attimg{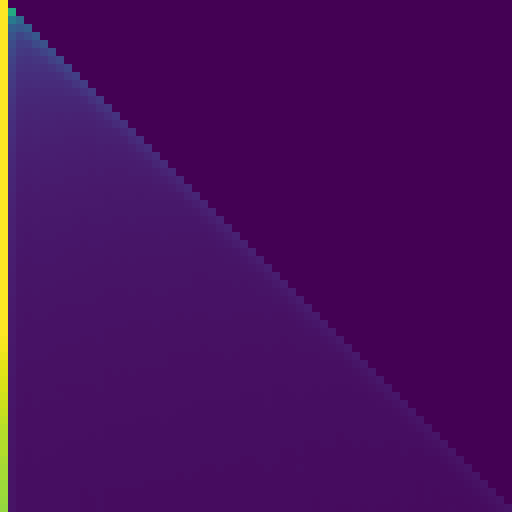} &
    \attimg{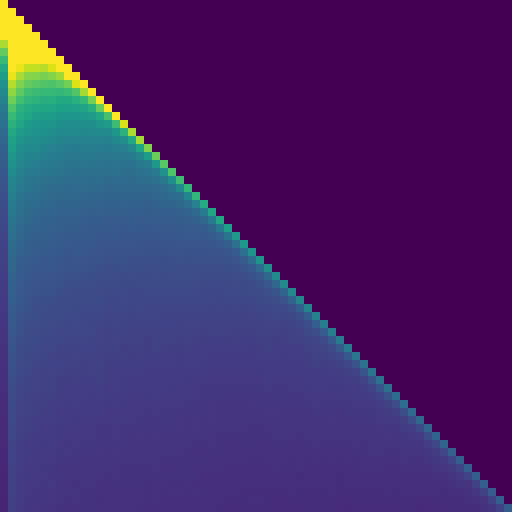} &
    \attimg{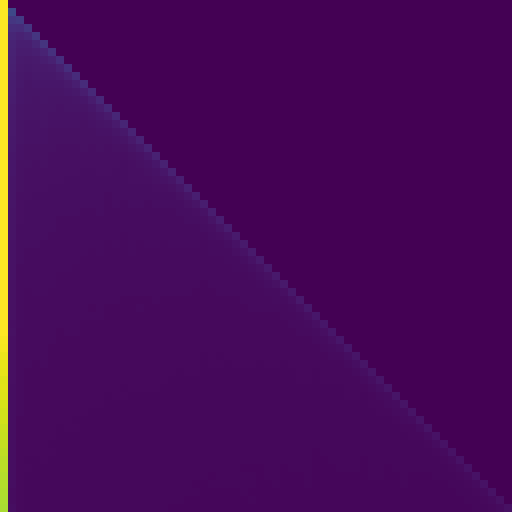} &
    \attimg{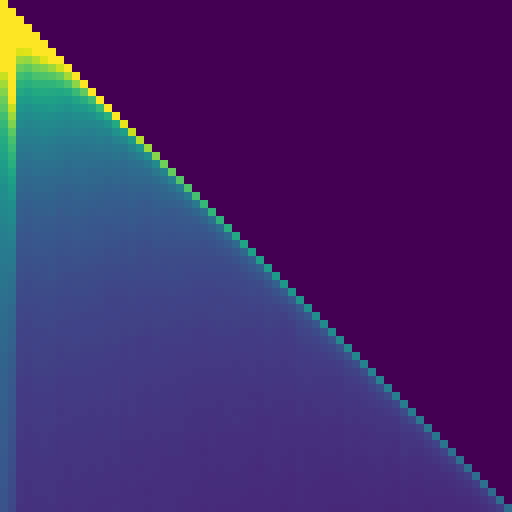} &
    \attimg{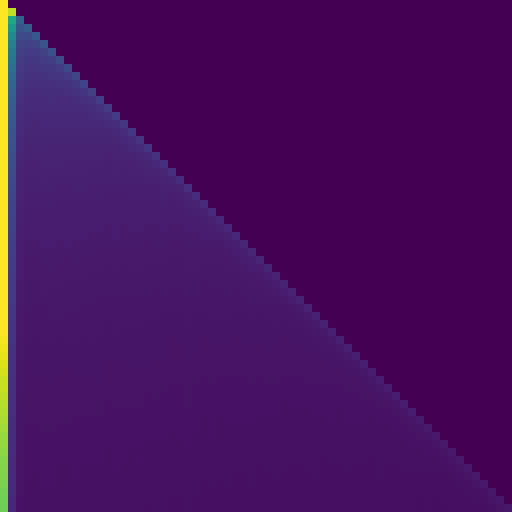} &
    \attimg{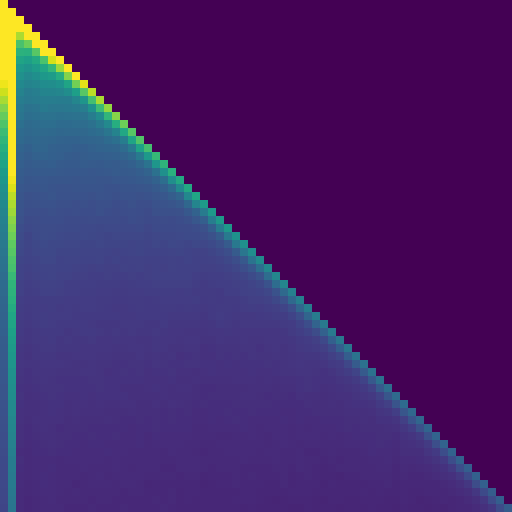} \\[-0.33em]

    \attlbl{Baseline}{6} & \attlbl{NAG}{6} &
    \attlbl{Baseline}{7} & \attlbl{NAG}{7} &
    \attlbl{Baseline}{8} & \attlbl{NAG}{8} \\

    \attimg{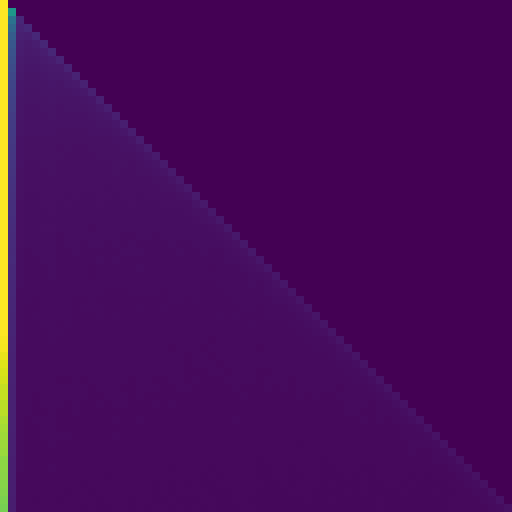} &
    \attimg{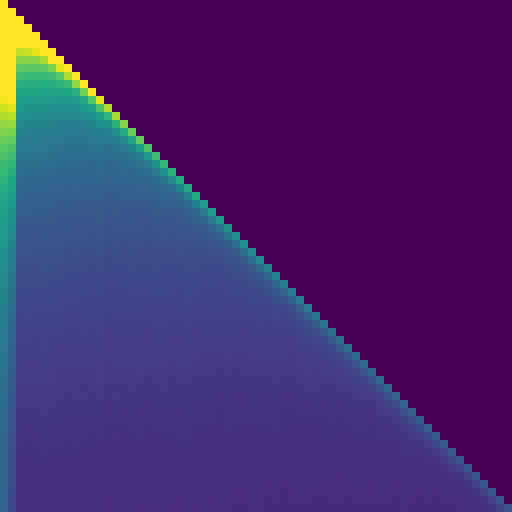} &
    \attimg{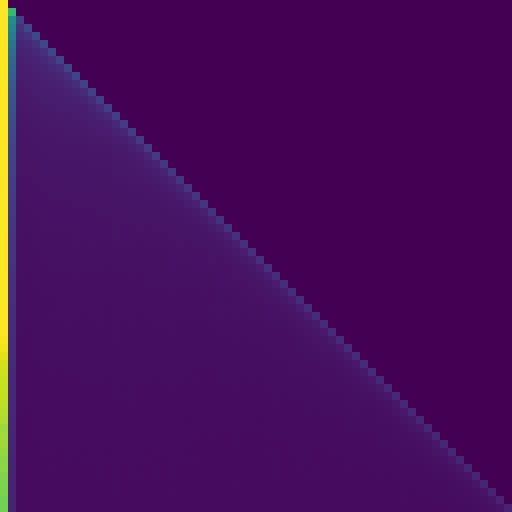} &
    \attimg{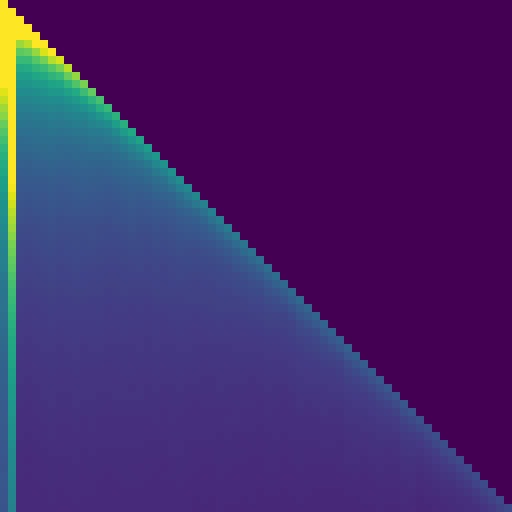} &
    \attimg{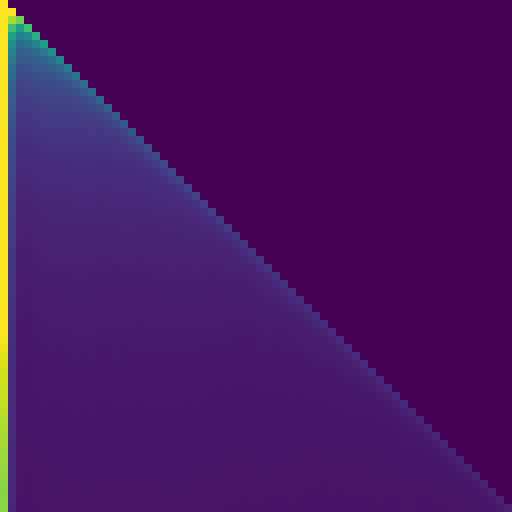} &
    \attimg{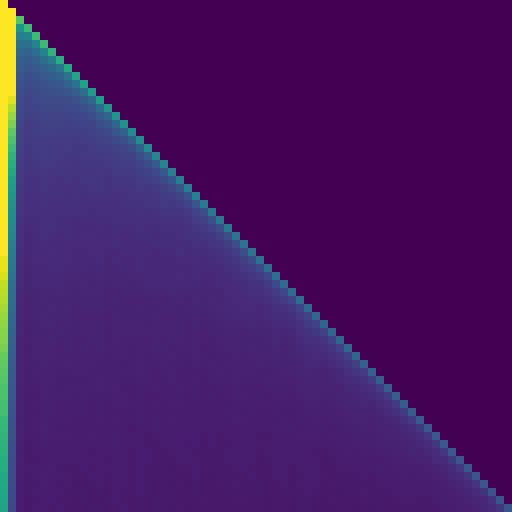} \\[-0.33em]

    \attlbl{Baseline}{9} & \attlbl{NAG}{9} &
    \attlbl{Baseline}{10} & \attlbl{NAG}{10} &
    \attlbl{Baseline}{11} & \attlbl{NAG}{11} \\

    \attimg{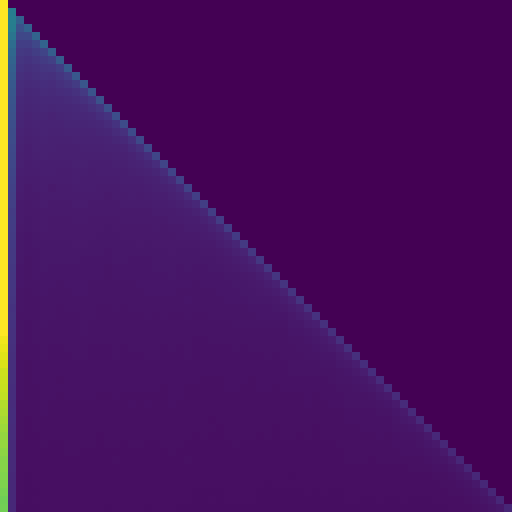} &
    \attimg{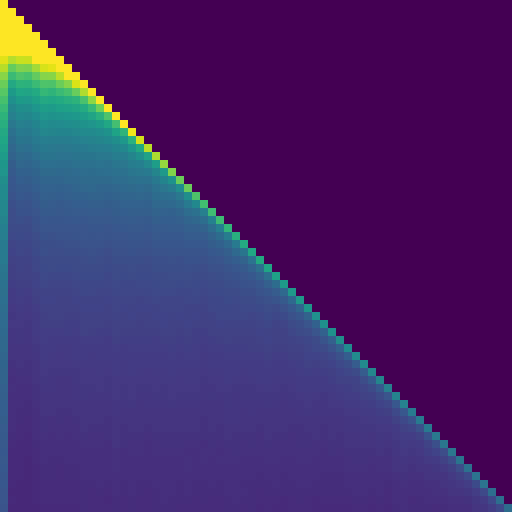} &
    \attimg{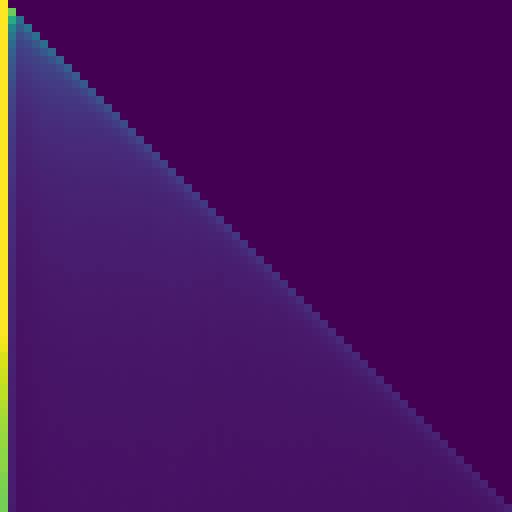} &
    \attimg{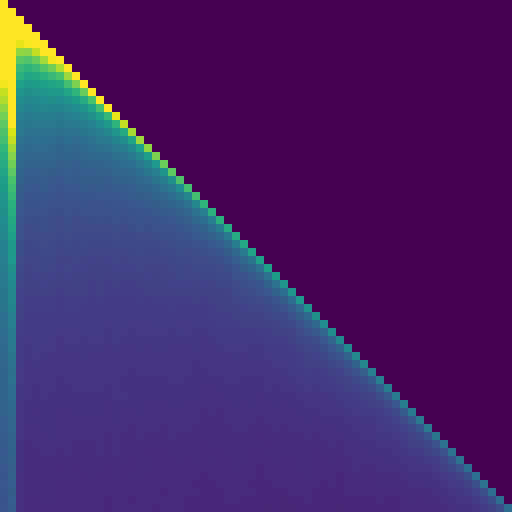} &
    \attimg{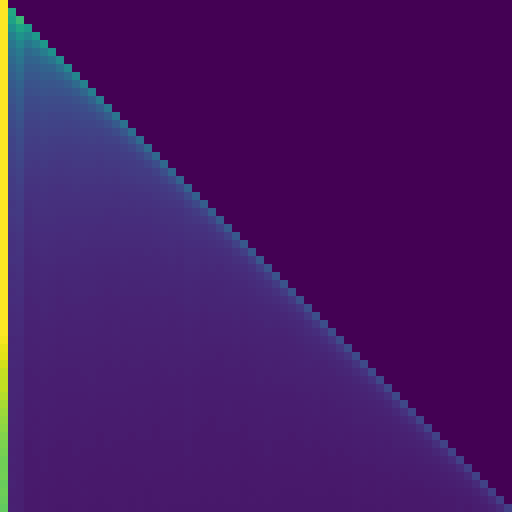} &
    \attimg{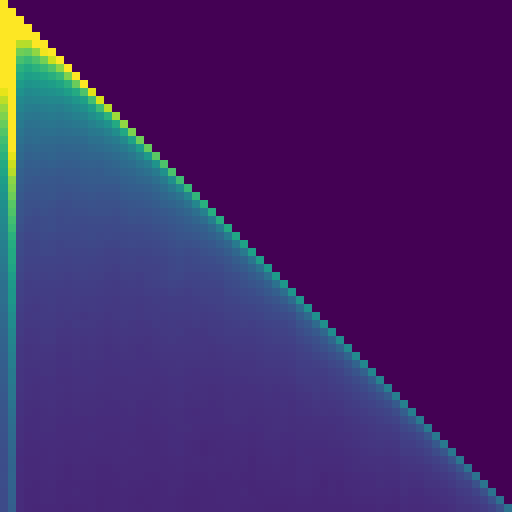} \\[-0.33em]

    \attlbl{Baseline}{12} & \attlbl{NAG}{12} &
    \attlbl{Baseline}{13} & \attlbl{NAG}{13} &
    \attlbl{Baseline}{14} & \attlbl{NAG}{14} \\

    \attimg{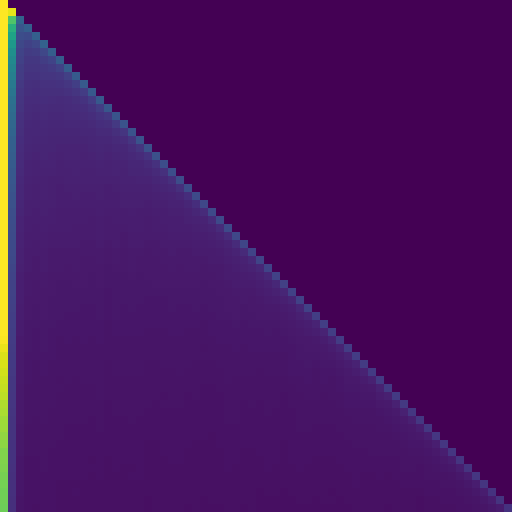} &
    \attimg{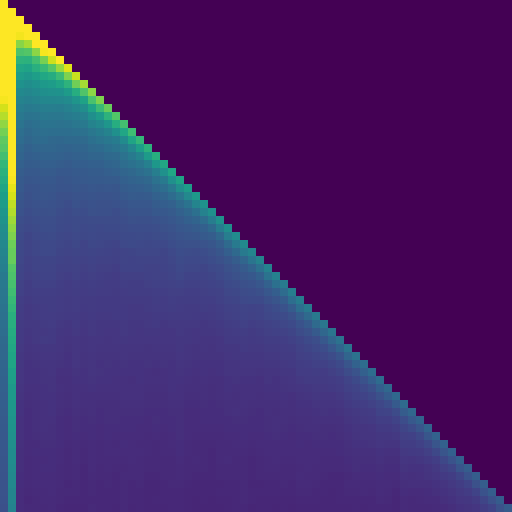} &
    \attimg{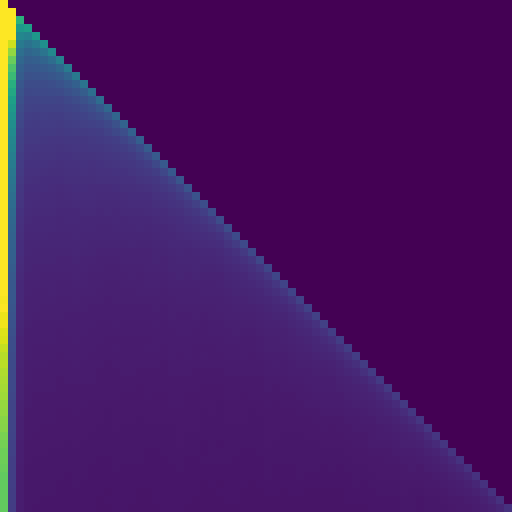} &
    \attimg{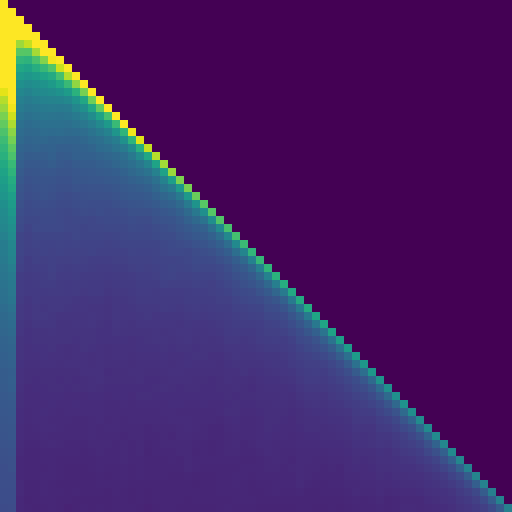} &
    \attimg{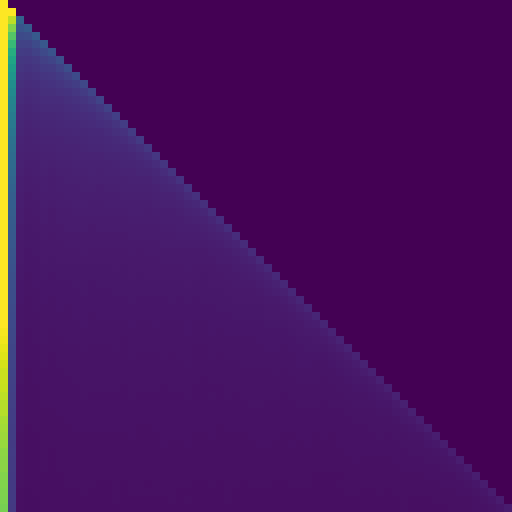} &
    \attimg{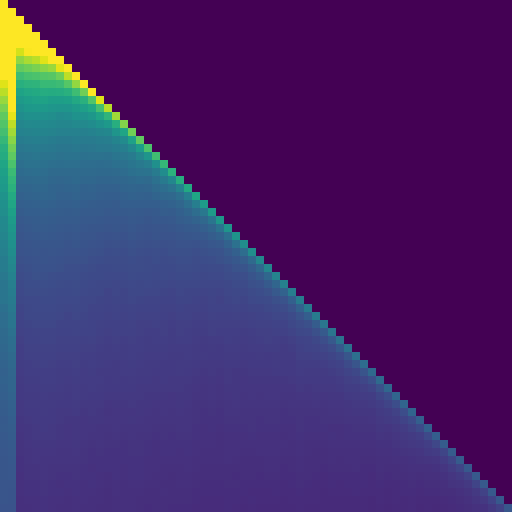} \\[-0.33em]

    \attlbl{Baseline}{15} & \attlbl{NAG}{15} &
    \attlbl{Baseline}{16} & \attlbl{NAG}{16} &
    \attlbl{Baseline}{17} & \attlbl{NAG}{17} \\

    \attimg{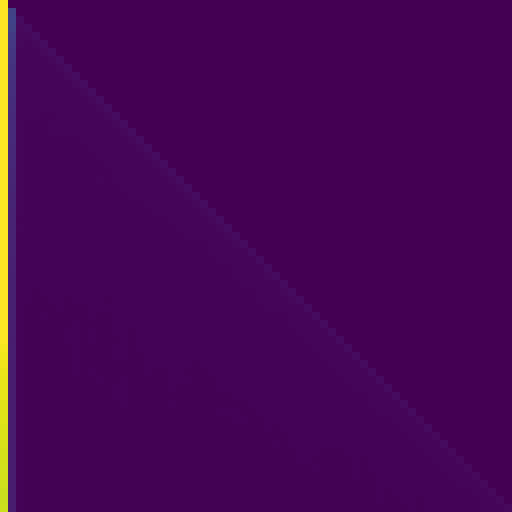} &
    \attimg{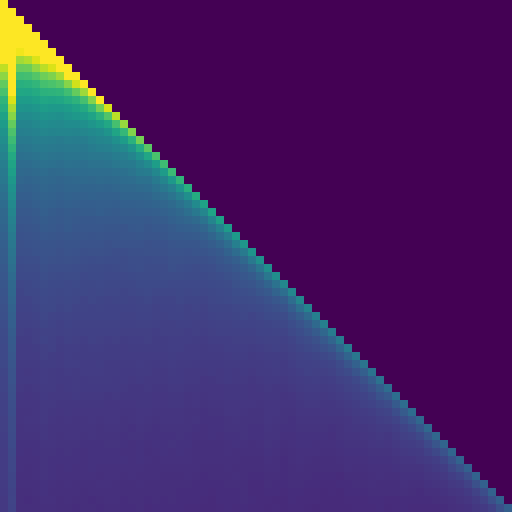} &
    \attimg{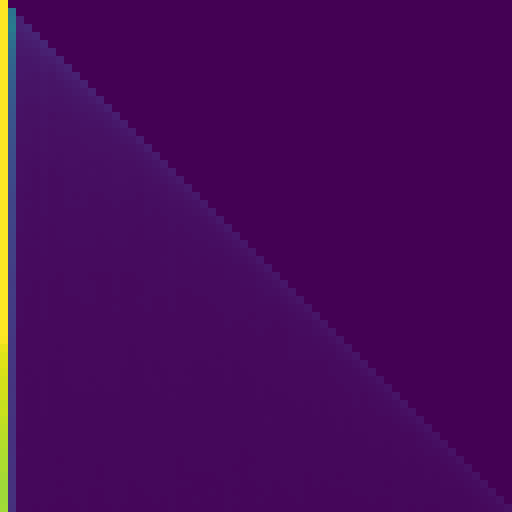} &
    \attimg{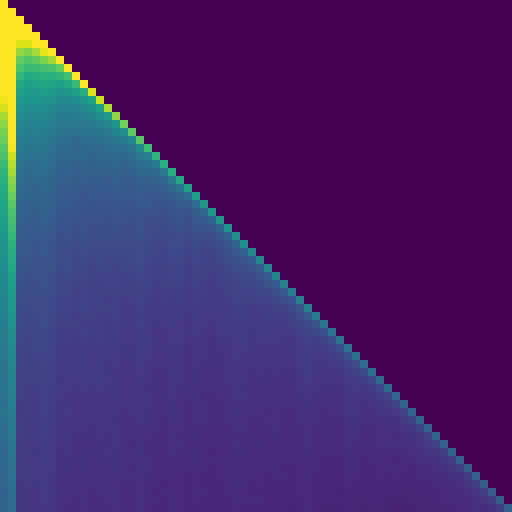} &
    \attimg{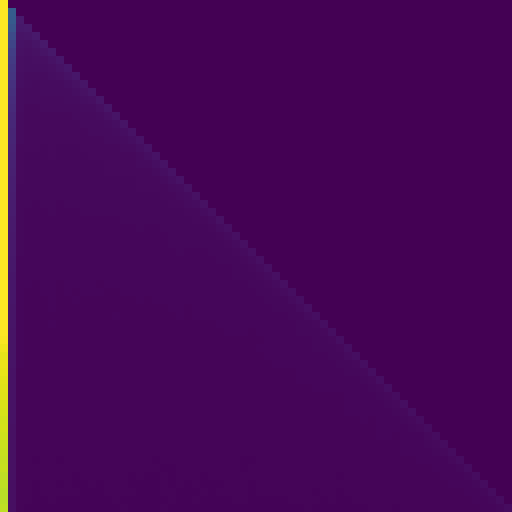} &
    \attimg{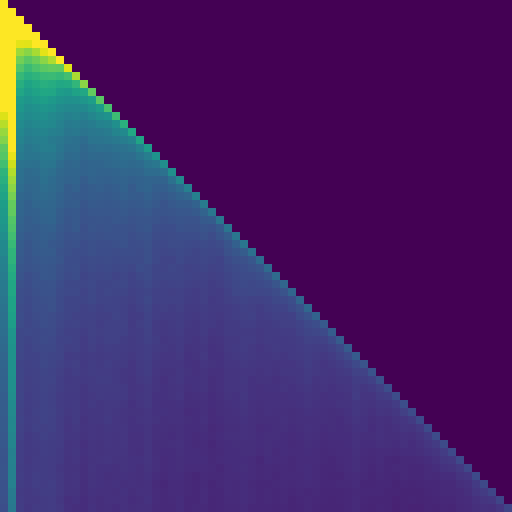} \\[-0.33em]

    \attlbl{Baseline}{18} & \attlbl{NAG}{18} &
    \attlbl{Baseline}{19} & \attlbl{NAG}{19} &
    \attlbl{Baseline}{20} & \attlbl{NAG}{20} \\

    \attimg{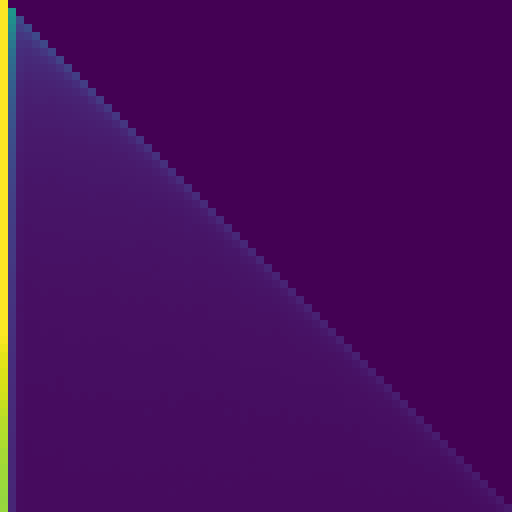} &
    \attimg{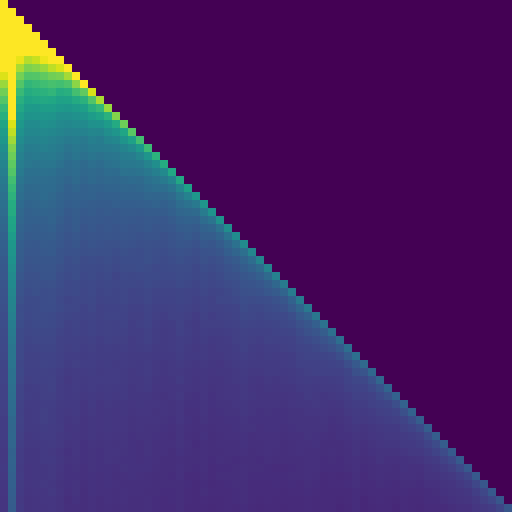} &
    \attimg{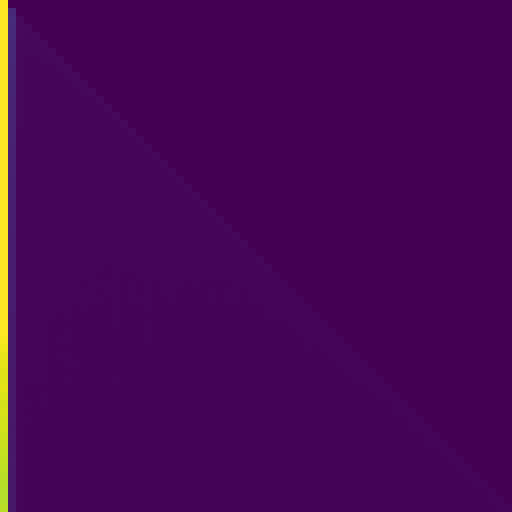} &
    \attimg{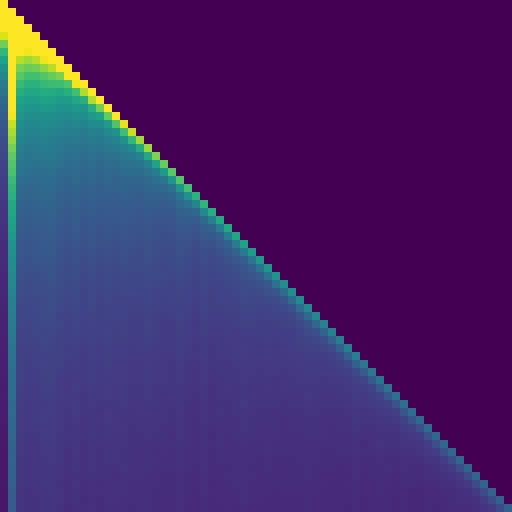} &
    \attimg{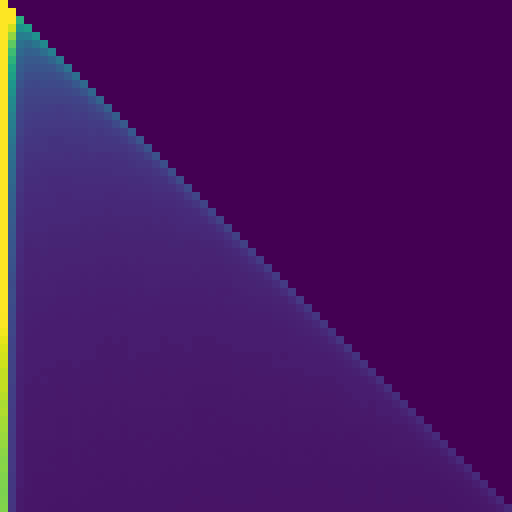} &
    \attimg{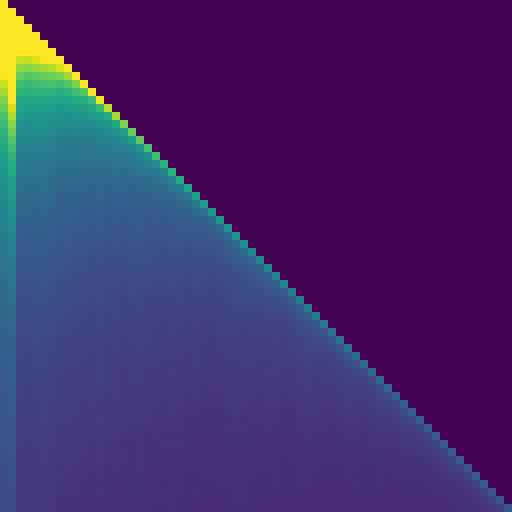}

    \end{tabular}

    \caption{Layer-wise comparison of average post-softmax attention patterns over the first 64 token positions. Rows correspond to query positions and columns to key positions. Each pair compares the baseline model and NAG at the same attention layer. The baseline exhibits pronounced vertical bands in early key-token positions across many layers, consistent with an attention-sink pattern, whereas these bands are substantially attenuated in NAG. This suggests that NAG reduces the prominence of attention sinks under our evaluation protocol.}
    
    \label{fig:attention_sink_layerwise_comparison}
\end{figure*}

\end{document}